\documentclass[pdflatex,sn-mathphys-num]{sn-jnl}

\usepackage{graphicx}%
\usepackage{multirow}%
\usepackage{amsmath,amssymb,amsfonts}%
\usepackage{amsthm}%
\usepackage{mathrsfs}%
\usepackage[title]{appendix}%
\usepackage{textcomp}%
\usepackage{manyfoot}%
\usepackage{booktabs}%
\usepackage{algorithm}%
\usepackage{algorithmicx}%
\usepackage{algpseudocode}%
\usepackage{listings}%
\usepackage{pdflscape}
\usepackage{adjustbox}
\usepackage{array}
\usepackage{mathtools}
\usepackage{makecell}
\usepackage{longtable}
\usepackage[table]{xcolor}
\usepackage{graphicx}
\usepackage{geometry}
\newcommand{\revise}[1]{\textcolor{black}{#1}}
\definecolor{lightred}{HTML}{FFCCCC} 
\usepackage{bm}
\usepackage[edges]{forest}

\usepackage{subfiles} 
\usepackage{booktabs,amsmath}

\tikzset{%
    parent/.style =          {align=center,text width=2cm,rounded corners=3pt, line width=0.3mm, fill=gray!10,draw=gray!80},
    child/.style =           {align=center,text width=2.3cm,rounded corners=3pt, fill=blue!10,draw=blue!80,line width=0.3mm},
    grandchild/.style =      {align=center,text width=2cm,rounded corners=3pt},
    greatgrandchild/.style = {align=center,text width=1.5cm,rounded corners=3pt},
    greatgrandchild2/.style = {align=center,text width=1.5cm,rounded corners=3pt},    
    referenceblock/.style =  {align=center,text width=1.5cm,rounded corners=2pt},
    unitask/.style =           {align=center,text width=2.2cm,rounded corners=3pt, fill=blue!10,draw=blue!80,line width=0.3mm},   
    pretrain_work/.style =           {align=center, text width=3cm,rounded corners=3pt, fill=blue!10,draw=blue!0,line width=0.3mm},  
    multitask/.style =           {align=center,text width=2.2cm,rounded corners=3pt, fill=red!10,draw=red!80,line width=0.3mm},   
    template_work/.style =           {align=center,text width=3cm,rounded corners=3pt, fill=red!10,draw=red!0,line width=0.3mm},    
    answer/.style =           {align=center,text width=1.8cm,rounded corners=3pt, fill= cyan!10,draw= cyan!80,line width=0.3mm},   
    answer_work/.style =           {align=center,text width=5cm,rounded corners=3pt, fill= cyan!10,draw= cyan!0,line width=0.3mm},      
    multiple/.style =           {align=center,text width=1.8cm,rounded corners=3pt, fill= orange!10,draw= orange!80,line width=0.3mm},   
    multiple_work/.style =           {align=center,text width=5cm,rounded corners=3pt, fill= orange!10,draw= orange!0,line width=0.3mm},        
    tuning/.style =           {align=center,text width=1.8cm,rounded corners=3pt, fill= magenta!10,draw= magenta!80,line width=0.3mm},   
    tuning_work/.style =           {align=center,text width=5cm,rounded corners=3pt, fill= magenta!10,draw= magenta!0,line width=0.3mm},          
}

\theoremstyle{thmstyleone}%
\theoremstyle{thmstyletwo}%

\theoremstyle{thmstylethree}%

\begin{document}

\title[Article Title]{Deep Learning based Visually Rich Document Content Understanding: A Survey}


\author[1,2]{\fnm{Yihao} \sur{Ding}}

\author*[1,2]{\fnm{Soyeon Caren} \sur{Han}}\email{caren.han@unimelb.edu.au}

\author[1]{\fnm{Jean} \sur{Lee}}

\author[1]{\fnm{Eduard} \sur{Hovy}}

\affil*[1]{\orgdiv{School of Computing and Information Systems}, \orgname{The University of Melbourne}, \orgaddress{\city{Melbourne}, \state{VIC}, \postcode{3010}, \country{Australia}}}

\affil[2]{\orgdiv{School of Computer Science}, \orgname{The University of Sydney}, \orgaddress{\city{Sydney}, \state{NSW}, \postcode{2006}, \country{Australia}}}


\abstract{\revise{Visually Rich Documents (VRDs) play a vital role in domains such as academia, finance, healthcare, and marketing, as they convey information through a combination of text, layout, and visual elements. Traditional approaches to extracting information from VRDs rely heavily on expert knowledge and manual annotation, making them labor-intensive and inefficient. Recent advances in deep learning have transformed this landscape by enabling multimodal models that integrate vision, language, and layout features through pretraining, significantly improving information extraction performance.
This survey presents a comprehensive overview of deep learning-based frameworks for VRD Content Understanding (VRD-CU). We categorize existing methods based on their modeling strategies and downstream tasks, and provide a comparative analysis of key components, including feature representation, fusion techniques, model architectures, and pretraining objectives. Additionally, we highlight the strengths and limitations of each approach and discuss their suitability for different applications. The paper concludes with a discussion of current challenges and emerging trends, offering guidance for future research and practical deployment in real-world scenarios.}}




\maketitle

\section{Introduction}
\label{sec:introduction}
\subsection{Backgrounds}
\revise{Visually Rich Documents (VRDs) are widely used across various domains, including finance, healthcare, law, education, and research. Both structured and unstructured visual-textual information must be interpreted in combination to derive meaningful insights. Unlike plain-text documents, VRDs contain heterogeneous components such as paragraphs, tables, charts, diagrams, and images, which are intricately arranged in formats like PDFs \cite{mmvqa,pdfvqa}, Word documents \cite{docbank}, or scanned images \cite{vrdiu}. These visual and textual elements, referred to as document \textbf{semantic entities}, play a critical role in summarizing, illustrating, or substantiating key content, often through complex layouts and visual cues.}

\revise{Understanding VRDs presents significant challenges due to their semi-structured or unstructured nature, which limits the effectiveness of traditional rule-based or linear text processing methods. This complexity has given rise to Visually Rich Document Content Understanding (VRD-CU), a specialized field focused on automatically extracting, interpreting, and reasoning over such documents to enable machine-readability and scalable knowledge extraction.}

\revise{Historically, early VRD-CU approaches were heuristic-driven, relying on layout analysis and domain-specific templates \cite{watanabe1995layout, seki2007information}, later expanding to include statistical models \cite{rusinol2013field, oliveira2017fast}. However, these methods required manually crafted rules, lacked flexibility, and often failed to generalize beyond fixed document structures or domains.
The emergence of deep learning marked a turning point in VRD-CU. Pioneering models such as Chargrid \cite{chargrid} and BERTgrid \cite{bertgrid} applied convolutional and recurrent neural networks to treat document layouts as two-dimensional text grids. This led to the development of feature-based models such as PICK \cite{pick}, TRIE \cite{trie}, and Ephoie \cite{ephoie}, which integrated layout features with textual signals for improved extraction.}

\revise{To acquire comprehensive VRD representations and improve generalization across VRD tasks, BERT-style pretrained frameworks have been proposed to achieve comprehensive document representation in VRD. They align multimodal information through various pretraining tasks. However, their focus on fine-grained word sequences limits the ability to capture higher-level semantic relationships. To address these limitations, coarse-grained models are introduced\cite{udoc, selfdoc} that focus on entity-level or semantic block-level understanding, offering stronger abstraction capabilities while sacrificing some fine-grained detail. Striking a balance between detail and abstraction, joint-grained frameworks such as StructExt \cite{structext}, StructExtV2 \cite{structextv2}, and WuKong \cite{wukong} have emerged, combining both granular and holistic representations. More recently, Large Language Model (LLM)-based architectures—including LayoutLLM \cite{layoutllm} and HRVDA \cite{hrvda}—have been proposed to support cross-modal reasoning and complex semantic understanding through instruction tuning and prompt-driven paradigms.}

In this survey, we provide a comprehensive review of deep learning-based VRD-CU methodologies, tracing their evolution from early neural encoders to current multimodal transformers and LLM-based frameworks. We categorize existing approaches based on task formulation, model architecture, feature integration strategies, and pretraining objectives, comparing their effectiveness, generalisability, and scalability across benchmark datasets. We further highlight emerging trends, challenges, and future directions, aiming to support both researchers and practitioners in navigating this dynamic and impactful research area.

\subsection{Scope}

This survey aims to provide a comprehensive and critical review of recent advances in \textbf{deep learning-based Visually Rich Document Content Understanding (VRD-CU)}. We focus on approaches that leverage neural architectures to understand and reason over the content of complex VRDs. The scope of this survey is defined by the following criteria:

\begin{itemize}
    \item \textbf{Focus on Content-Level Understanding Tasks:} We include works that address \textit{document content understanding} tasks, specifically Key Information Extraction (KIE), Visually Rich Document Question Answering (VRD-QA), Entity Linking (EL). We \textbf{exclude} studies that are limited to \textit{document structure analysis} (e.g., layout segmentation or reading order detection) without directly contributing to semantic content understanding.

    \item \textbf{Deep Learning-based Models:} We primarily review papers that propose novel \textit{deep learning methods} tailored for VRD-CU tasks. While Section~\ref{sec:traditional_approach} briefly summarises early heuristic and traditional machine learning approaches for historical context, the core focus remains on modern neural frameworks.

    \item \textbf{Dataset Coverage:} We compile and analyse publicly available \textit{benchmark datasets} for VRD-CU tasks, particularly those introduced or widely adopted in \textit{top-tier conferences and journals since 2019}. These datasets support content-level reasoning in multimodal documents.

    \item \textbf{Emphasis on Multimodal and Multi-page Understanding:} In contrast to previous surveys that focus narrowly on subtasks or single-page settings, this paper highlights models capable of \textit{multimodal reasoning} and handling \textit{entire, multi-page documents holistically}, better reflecting real-world document comprehension needs.
\end{itemize}

This survey is designed to support researchers and practitioners in understanding the current landscape of deep learning models for VRD-CU, identifying emerging trends, and navigating the key resources, including models and datasets, which define this growing field.
It should be noted that this article only investigates models and datasets designed for the entire document or the document page rather than the specific document components such as \textit{table}, \textit{chart}. Therefore, models and datasets proposed for table detection, table structure recognition, and chart or plot question answering will not be summarized in the main body of this survey. 

\subsection{Related Surveys} \label{sec:related_survey}
Several surveys offer a comprehensive overview of general document understanding tasks. These surveys primarily focus on document layout analysis \cite{subramani2020survey}, table extraction \cite{liu2023tabular}, named entity recognition \cite{ehrmann2023named}, and document image analysis \cite{lombardi2020deep} across diverse document types, including invoices \cite{saout2024overview} and historical document \cite{ehrmann2023named, lombardi2020deep}. In particular, computer vision-based research focuses on scanned document analysis and structure understanding. While these studies have advanced document image analysis, they often focus on fragmented subtasks and fall short of providing a holistic understanding of an entire document on multiple pages. 

Recent advancements in deep learning have fueled the emergence of VRD-CU tasks, which demand complex document content understanding capabilities. These tasks include key information extraction, question answering, and document entity linking. However, existing surveys have not adequately addressed the unique challenges and opportunities presented by VRD-CU, with a specific emphasis on deep learning-based multimodal approaches \cite{cui2021document}. To bridge this gap, this survey aims to provide a comprehensive overview of VRD-CU frameworks and datasets, including multimodal feature extractions and fusions in both mono and multi-task VRD models.

\subsection{Contributions}
The main contributions of this paper can be summarized as follows:
\begin{itemize}
    \item The paper provides a detailed review and systematic categorization of VRD-CU frameworks and benchmark datasets, organized based on adopted strategies and downstream tasks.
    \item It critically examines and compares different techniques used in VRD-CU models, focusing on feature representation and fusion, model architecture, and pretraining methods, highlighting their strengths, limitations, and appropriate scenarios.
    \item The paper identifies emerging trends and challenges in visually rich document content understanding, offering insights into future research directions and practical applications.
\end{itemize}

\subsection{Survey Structure}
Section~\ref{sec:introduction} introduces the background of VRD-CU, outlining the aim, scope, and key contributions of this survey. Section~\ref{sec:background} provides essential background knowledge, including definitions of the VRD-CU tasks, the evolution of document understanding techniques and terminology definition. Sections~\ref{sec:mono_task} and~\ref{sec:multi_task} review notable mono-task and multi-task frameworks, respectively. Section~\ref{sec:dataset_metric} presents benchmark datasets across three VRD-CU subtasks with evaluation metrics description. Section~\ref{sec:discussion} offers a critical analysis of the reviewed models, highlighting their strengths and limitations through quantitative evaluation. Section~\ref{sec:challenges} identifies emerging trends and key challenges, with a focus on practical applications. Finally, Section~\ref{sec:conclusion} summarizes the survey findings.

\section{Background}
\label{sec:background}

\subsection{Task Definition}
\begin{figure}[t]
  \centering
  \includegraphics[width=\linewidth]{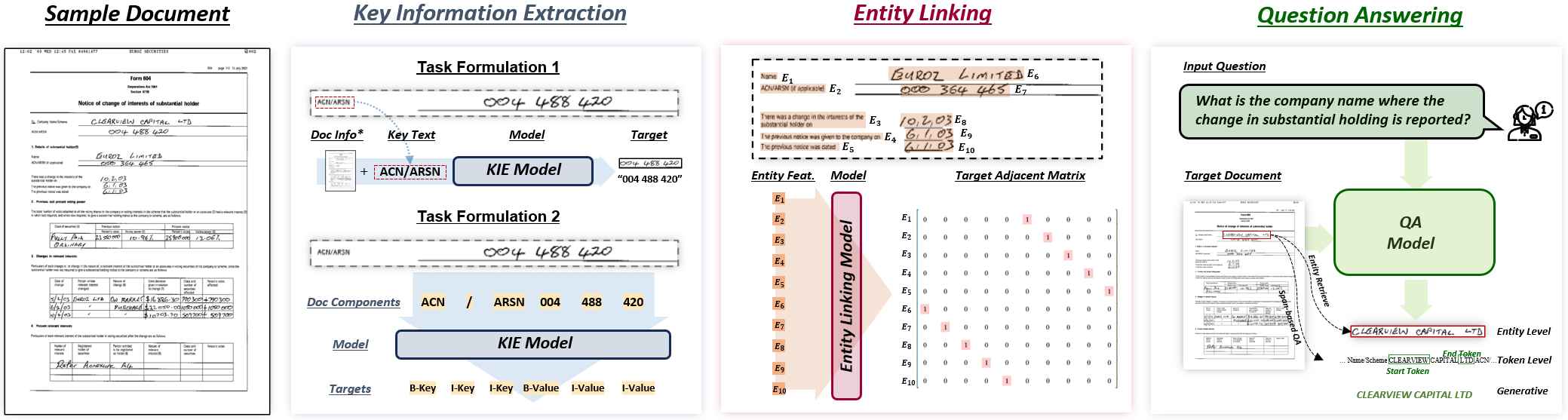}
  \caption{Visually rich document content understanding task clarifications.}
  \label{fig:vrdu_task_definition}
\end{figure}

Based on their purpose and application scenarios, visually rich document content understanding (VRD-CU) tasks can be categorized into three main types: \textbf{Key Information Extraction}, \textbf{Question Answering}, and \textbf{Entity Linking}, as illustrated in Figure~\ref{fig:vrdu_task_definition}.

\begin{itemize}
    \item \textbf{Key Information Extraction}: refers to identifying and extracting the relevant information based on the given text queries. Distinct pre-defined queries can be defined based on the domain of targeting documents and practical demands. For example, the crucial information of scanned receipts contains "\textit{Store Name}", "\textit{Address}", "\textit{Item}" and "\textit{Price}", while for the financial reports, "\textit{Company Name}", "\textit{Share Holder Name}", "\textit{Number of Interests}" may be the critical information need to be extracted. 
    \item \textbf{Entity Linking}: refers to identifying the semantic relations between document semantic entities to construct the logical structure of the input document image.
   
    \item \textbf{Question Answering}: is a task of answering questions about a VRD by using natural languages. Based on the answer types, it can be divided into extractive QA and generative QA. The answer from extractive QA is directly extracted from the target document, while generative QA requires generating answers based on comprehensively understanding questions and related VRDs. 
   
\end{itemize}

\subsection{Development of Document Understanding}

\subsubsection{Traditional Approaches} \label{sec:traditional_approach}
Rule-based methods, as highlighted in several studies \cite{watanabe1995layout,o1993document,seki2007information,rusinol2013field}, have demonstrated high precision in domain-specific applications. However, these methods have several drawbacks: they are manually intensive, costly, and require expert intervention for tailored customization. Additionally, they are inflexible, often necessitating frequent manual updates, even for minor modifications.
In response to these limitations, machine learning-based approaches have been proposed for document understanding. For instance, SVM-based methods have been utilized for layout understanding \cite{oliveira2017fast}, and TF-IDF techniques combined with hand-crafted features have been applied for extracting information from invoices. Furthermore, rule-based and statistical models have been used for entity extraction. Despite these advancements, machine learning methods rely heavily on human intervention and domain-specific expertise. They are time-consuming and often deliver suboptimal performance. Moreover, the majority of these methods typically depend on single-modality data, restricting them to either layout, text, or visual inputs.

\subsubsection{Single Modality-based Approaches} With the advancement of deep learning, deeper model architectures such as CNNs \cite{chargrid,yang2017learning} have emerged, and pretrained language \cite{bert,roberta} and vision models \cite{fasterrcnn,maskrcnn} are now commonly used as robust baselines for understanding VRD content and structure. Due to the multimodal nature of VRDs, which involves the integration of text, vision, and layout, researchers are increasingly focusing on leveraging this combined information to achieve significant improvements in various downstream VRD-CU tasks.

\subsubsection{Cross Modality based Approaches} Considering the multimodal nature of VRDs, many frameworks propose various methods to encode multimodal information, including text, vision, and layout, and fuse them effectively. Text and vision information is normally encoded by pretrained backbones such as BERT \cite{bert} or RoBERTa \cite{roberta} for textual features, Faster-RCNN \cite{fasterrcnn} or Mask-RCNN \cite{maskrcnn} for visual features. For layout information, different encoding methods are introduced, including linear projection \cite{docstruct}, 2D positional encoding \cite{layoutlm}, and attention bias to allow the proposed models to be layout-sensitive. Different feature fusion methods are introduced including summing up \cite{pick}, concatenation \cite{formnet}, attention-based contextual learning \cite{majumder2020representation}, and prompting \cite{icld3ie}. However, most of those frameworks leverage implicit knowledge from pretrained backbones with a task-orientated shadow adapter for specific VRD-CU downstream tasks such as KIE \cite{formnetv2,laser,chen2023task,genkie} or EL \cite{zhang2021entity,kvpformer,carbonell2021named}. Those frameworks tend to achieve delicate performance on specific tasks or document formats instead of acquiring a generalised model to represent documents comprehensively. 

\subsubsection{Multimodal Pre-training Approches} Inspired by the success of BERT-style models \cite{bert,roberta} in acquiring knowledge through self-supervised learning, pretrained document understanding models have emerged to harness self-supervised or supervised pretraining tasks from extensive document collections. LayoutLM \cite{layoutlm}, the first encoder-only VRD-CU model, utilizes self-supervised tasks, such as masked vision-language modelling, with text and layout information. Subsequent models have expanded on this by integrating layout information \cite{lilt,layoutmask,structext} and visual cues \cite{layoutlmv2,layoutlmv3} through multimodal transformers. While encoder-only models have shown significant improvements on various benchmark datasets \cite{funsd,cord,sroie,docvqa,rvlcdip}, they often require detailed annotations and are limited by fixed input lengths. To address these limitations, encoder-decoder frameworks \cite{udop,donut,dessurt} and prompt-based methods for LLMs/MLLMs \cite{layoutllm,hrvda} have been developed, enhancing layout awareness and performance in VRD-CU tasks. However, a significant gap remains in effectively applying these models in real-world scenarios with zero shot.

\subsection{Terminology Definition}
After reviewing the evolution of VRD-CU, we introduce standardized terminology to categorize existing frameworks along three key dimensions: \textbf{Downstream Task Focus}, \textbf{Model Architecture}, \textbf{Textual Granularity} and \textbf{Visual Information}.

\begin{itemize}
    \item \textbf{Downstream Task Focus:} Frameworks are categorized into \textit{mono-task} and \textit{multi-task} types. Mono-task frameworks (see Section~\ref{sec:mono_task}) are tailored for a single VRD-CU task (e.g., KIE or VQA) with task-specific designs, while multi-task frameworks (see Section~\ref{sec:multi_task}) are designed to handle multiple tasks simultaneously, typically leveraging pretraining or shared components to improve generalizability.
    \item \textbf{Model Architecture:} Frameworks can be grouped into \textit{encoder-only}, \textit{encoder-decoder}, and \textit{decoder-only} categories. Encoder-only models focus on representation learning; encoder-decoder models generate task-specific outputs from document inputs; decoder-only models use autoregressive decoding for end-to-end generation.
    \item \textbf{Textual Granularity:} Depending on how document structure is represented, frameworks process inputs at different levels: \textbf{fine-grained} (word-level), \textbf{coarse-grained} (e.g., entity-level inputs such as paragraphs, tables, or section headers), or \textbf{joint-grained}, which combines both. Fine-grained inputs capture detailed spatial and lexical information through individual word tokens, while coarse-grained representations emphasize higher-level semantic structures and contextual dependencies.

    \item \textbf{Visual Information:} For frameworks that incorporate visual information, we categorize them based on the visual feature extraction methods: \textbf{RoI-based} and \textbf{patch-based}. RoI-based frameworks often use object detectors or document parsers to identify Regions of Interest (RoIs) corresponding to semantic entities, followed by a CNN (e.g., ResNet) to extract visual features via RoI-Align—commonly referred to as \textit{RoI-based} frameworks. In contrast, Transformer-based frameworks (e.g., ViT \cite{vit}, Swin-Transformer \cite{liu2021swin}) typically divide the document image into uniform patches and encode them sequentially, forming the basis of \textit{patch-based} frameworks.

\end{itemize}
These terms provide a unified lens to understand and compare the design choices across VRD-CU frameworks in Section~\ref{sec:mono_task}, Section~\ref{sec:multi_task} and Section~\ref{sec:discussion}.

\section{Mono-Task Document Understanding Frameworks}
\label{sec:mono_task}
VRD-CU  comprises several distinct downstream tasks, each designed to meet specific application needs and user requirements. This section introduces methods tailored to individual tasks. As illustrated in Figure~\ref{fig:mono_task_framework}, we focus on three core VRD-CU tasks—\textbf{Key Information Extraction (KIE)}, \textbf{Entity Linking (EL)}, and \textbf{Question Answering (QA)}—and summarize corresponding models along with key trends and insights in their development.

\begin{figure*}[htb]
\footnotesize
\begin{forest}
    for tree={
        forked edges,
        grow'=0,
        draw,
        rounded corners,
        node options={align=center,},
        text width=2.8cm,
        s sep=6pt,
        calign=child edge, 
        calign child=(n_children()+1)/2
    }
    [Mono-task based models (Content Understanding), for tree={fill=lime!60}
        [Key Information Extraction, for tree={fill=cyan!40}
            [\textbf{Feature-driven} Models, for tree={fill=cyan!25}
                [Chargrid\cite{chargrid};
                CUTIE\cite{cutie};
                BERTgrid\cite{bertgrid};
                ACP\cite{palm2019attend};
                XYLayoutLM\cite{xylayoutlm}, for tree={fill=cyan!10}]
            ]
            [\textbf{Joint-learning} Frameworks, for tree={fill=cyan!25}
                [TRIE\cite{trie};
                VIES\cite{ephoie}, for tree={fill=cyan!10}]
            ]
            [\textbf{Relation-aware} Models, for tree={fill=cyan!25}
                [Majumder et al. \cite{majumder2020representation};
                Liu et al. \cite{liu2019graph};
                PICK\cite{pick};
                FormNet\cite{formnet};
                FormNetv2\cite{formnetv2}, for tree={fill=cyan!10}]
            ]
            [\textbf{Few/Zero-shot Learning} Models, for tree={fill=cyan!25}
                [LASER\cite{laser};
                Chen et al.\cite{chen2023task};
                Cheng et al.\cite{cheng2020one};
                QueryForm\cite{queryform}, for tree={fill=cyan!10}]
            ]
            [\textbf{Prompt-based} Frameworks, for tree={fill=cyan!25}
                [GenKIE\cite{genkie};
                ICL-D3IE\cite{icld3ie};
                LMDX\cite{lmdx}, for tree={fill=cyan!10}]
            ]
        ]
        [Entity Linking, for tree={fill=blue!40}
            [\textbf{Entity-level} Linking, for tree={fill=blue!20}
                [DocStruct\cite{docstruct};
                Zhang et al.\cite{zhang2021entity};
                KVPFormer.\cite{kvpformer}, for tree={fill=blue!10}]
            ]
            [\textbf{Word-level} Linking, for tree={fill=blue!20}
                [Carbonell et al. \cite{carbonell2021named};
                SPADE \cite{hwang2021spatial};DocTR\cite {doctr}, for tree={fill=blue!10}]
            ]
        ]
        [Question Answering, for tree={fill=teal!40}
            [\textbf{Single Page} Frameworks, for tree={fill=teal!25}
                [Please refer to \textbf{Section}~\ref{sec:multi_task}. , for tree={fill=teal!10}]
            ]
            [\textbf{Multi-Page} Frameworks, for tree={fill=teal!25}
                [Hi-VT5\cite{mpdocvqa};
                GRAM\cite{gram};
                Kang et al. \cite{kang2024multi}, for tree={fill=teal!10}]
            ]
        ]
    ]
\end{forest}

\caption{Mono-task visually rich document understanding models}
\label{fig:mono_task_framework}
\end{figure*}

\subsection{Key Information Extraction}
Key Information Extraction (KIE), a typical natural language processing task, refers to the task of identifying and extracting crucial pieces of information from textual data. Unlike typical name entity recognition methods for addressing plain text, VRDs contain visually rich entities like \textit{tables} and \textit{charts}, as well as spatial and logical layout arrangements to enhance the challenge of extracting crucial information. Although plain-text pretrained language models, such as BERT \cite{bert}, RoBERTa \cite{roberta}, and ALBERT \cite{albert}, are widely used as solid baselines on many benchmark datasets, more recent works introduce layout-aware pretrained models such as LayoutLM families \cite{layoutlm,layoutlmv2,layoutlmv3}, LiLT \cite{lilt}, Bros \cite{bros} to enhance the document representation by leveraging visual and layout information and achieving SoTA performance on several downstream tasks (see Section~\ref{sec:multi_task}). This section will mainly focus on models specifically proposed for the document KIE models or only evaluated on KIE benchmark datasets including FUNSD \cite{funsd}, CORD \cite{cord}, SROIE \cite{sroie}, XFUND \cite{layoutxlm}, etc. Based on innovative aspects, we categorise KIE frameworks into five types: \textit{Feature-driven models} use multimodal cues for rich feature representation. \textit{Joint Learning frameworks} integrate auxiliary tasks to enhance document representation. \textit{Relation-aware models} leverage spatial or logical relations via graphs or masked attention mechanisms. \textit{Few/Zero-shot learning frameworks} explore methods for extracting key information with minimal labelled data, often using transfer learning. \textit{Prompt-based frameworks} use structured prompts to guide specific information extraction from pretrained models or LLM/MLLMs. These categories represent diverse approaches to improving KIE by leveraging specific model designs and learning strategies tailored to document understanding challenges.

\subsubsection{Feature-driven Models}

In the initial phase, certain Recurrent Neural Network (RNN)-based models \cite{lample2016neural} were introduced, primarily focusing on key information extraction tasks from plain text. However, these approaches overlook the importance of visual cues and layout information. Hence, several multimodal frameworks with feature-driven designs have been proposed to generate more representative document representations.

\textbf{\textit{Chargrid}} \cite{chargrid} first mentioned the significance of 2D structure for document KIE and designed a character-box-based chargrid to convert the textual and 2D layout structure into coloured \textbf{visual cues} feed into CNN. Instead of using fine-grained character information, \textbf{\textit{CUTIE}} \cite{cutie} and \textit{\textbf{BERTgrid}} \cite{bertgrid} utilise various word embedding methods with bounding box coordinates to persevere the layout structure. \textbf{\textit{ACP}} \cite{palm2019attend} leverages attention mechanism and multi-aspect features, including visual, semantic (both character and word) and spatial, into dilated CNN for capturing both short and long-term dependencies of each word piece.


Joint learning KIE frameworks are proposed to leverage multi-level features to mitigate the information gap between various focused tasks. \textbf{\textit{TRIE}} \cite{trie} firstly offers an end-to-end framework for simultaneously conducting Optical Character Recognition (OCR) and KIE. OCR module will generate multi-aspect features, including positional, visual and textual aspects. Adaptively trainable weighting mechanisms are adopted to generate the fused embedding, followed by a Bi-LSTM-based Entity Extraction Module to conduct the final prediction. \textbf{\textit{VIES}} \cite{ephoie} is introduced to use the multi-level cues of vision, position and text to generate more comprehensive representations. Both token and segment (entity) level positional and visual features are acquired from text detection modules, and dual-level textual features are gathered by text recognition brunch, fused by a self-attention-based fusion module for sequence labelling.

\subsubsection{Relation-aware Models}

Beyond leveraging multimodal features, modeling the spatial and logical relationships between document components can yield more robust and context-aware representations. Relation-aware models explicitly encode these inter-element dependencies—often using graph structures—to support complex information extraction tasks. \textbf{\textit{Majumder et al.}}~\cite{majumder2020representation} propose a two-stage approach: field-specific detectors first generate candidates, which are then scored by a neural model that encodes both textual and spatial features. This enables the model to adapt across varying document layouts by focusing on field relevance. 

\subsubsection{Relation-aware Models}

Beyond leveraging multimodal signals, modeling the spatial and logical relationships between document components can yield more robust and context-aware representations. Relation-aware models explicitly encode these inter-element dependencies—often using graph structures—to support complex information extraction tasks.
\textbf{\textit{Majumder et al.}}~\cite{majumder2020representation} propose a two-stage approach: field-specific detectors first generate candidates, which are then scored by a neural model that encodes both textual and spatial features. This enables the model to adapt across varying document layouts by focusing on field relevance.

Graph-based frameworks have gained traction for encoding spatial/logical relationships. \textbf{\textit{Liu et al.}}~\cite{liu2019graph} pioneer this by constructing a fully connected graph where each node is a textual entity $E_i \in \mathbb{E}$ and edges capture relative spatial features:
\[
e_{ij} = \left[x_{ij},\ y_{ij},\ \frac{w_i}{h_i},\ \frac{h_j}{h_i},\ \frac{w_j}{h_i}\right]
\]
Graph convolution is then applied over node-edge-node triplets using self-attention to propagate contextual information.

\textbf{\textit{PICK}}~\cite{pick} extends this design with multimodal node features, combining transformer-based text embeddings and CNN-based visual features. The edge representation is also enriched with sentence length information. 
To enhance tagging performance, PICK applies a soft adjacency matrix for graph learning, yielding task-specific node embeddings.

\textbf{\textit{FormNet}}~\cite{formnet} introduces a novel attention mechanism—\textit{Rich Attention}—that incorporates positional order and pixel distances (along $x/y$ axes) into self-attention scores. This allows transformers to be aware of layout order and spatial proximity. Additionally, a graph encoder integrates neighborhood context prior to serialization, mitigating order ambiguity. 

\textbf{\textit{FormNetv2}}~\cite{formnetv2} further enhances this by introducing visual cues as edge features and applying contrastive learning. Two types of graph corruption—topological (edge dropping) and feature-level (modality masking)—are used to generate augmented views. The model then learns robust multimodal embeddings using a normalized temperature-scaled cross-entropy loss~\cite{sohn2016improved}.

In summary, these relation-aware models highlight the shift from flat sequence modeling to structure-aware document representation. While their specific graph construction and fusion techniques vary, they all aim to capture the rich spatial and semantic dependencies crucial for key information extraction and related tasks.

\subsubsection{Few-shot Learning Frameworks} Extracting key information from VRDs using deep learning typically requires extensive manual annotation. However, in many real-world scenarios, obtaining large-scale, high-quality labels is costly and time-consuming. To address this challenge, few-shot and one-shot learning frameworks have been developed to enable key information extraction with minimal annotated data.

\textbf{\textit{LASER}}~\cite{laser} builds upon the architecture of \textbf{\textit{LayoutReader}}~\cite{layoutreader}, reformulating entity recognition from a sequence labeling task into a generative one. By embedding entity type information—referred to as label surface names—into the target sequence, LASER enables the model to become label-semantic aware. It uses a \textit{Partially Triangular }attention mask, allowing a single encoder to encode both the source text and the generated sequence. The input sequence $\{t_1, ..., t_n\}$, comprising the sum of word, spatial, and positional embeddings, is fed into the encoder $\mathcal{E}$ with full self-attention. The decoder attends only to previously generated tokens. The generative target format is defined as:
\begin{gather}
t_{i-1},\ [B],\ t_i, ..., t_j,\ [E],\ e_1, ..., e_k,\ [T],\ t_{j+1}
\end{gather}
where $[B]$ and $[E]$ denote the entity span, $e_i$ represent the label surface name, and $[T]$ marks the end of the label name. Since label tokens and special symbols do not appear in the source, learnable embeddings are assigned to them. A binary classifier predicts whether the next token should be copied from the source, effectively guiding the generative process.

\textbf{\textit{Chen et al.}}~\cite{chen2023task} propose an entity-level \textit{N-way soft-K-shot} learning framework for key information extraction under few-shot settings. Targeting rare or unseen entity types, their approach employs meta-learning~\cite{snell2017prototypical, oreshkin2018tadam} with a hierarchical decoder and contrastive learning module (\textbf{\textit{ContrastProtoNet}}) for task personalization and improved adaptation. They also introduce \textbf{\textit{FewVEX}}, a benchmark dataset designed for entity-level few-shot VRD extraction, and demonstrate that their method significantly outperforms existing meta-learning baselines in robustness and performance.

In the context of \textbf{one-shot} learning, \textbf{\textit{Cheng et al.}}~\cite{cheng2020one} propose a graph-based method that transfers spatial relationships from support to query documents. Their approach leverages attention mechanisms to model interactions between static regions (keys/landmarks) and dynamic regions (values/fields), enabling accurate label probability distribution inference. A self-attention module captures dependencies among field entities, while a pairwise Conditional Random Field (CRF) with belief propagation completes the inference, resulting in an end-to-end trainable pipeline.

\subsubsection{Prompt-learning Frameworks} 
Prompt learning is a technique in natural language processing that guides models using specific prompts to elicit targeted responses. With the emergence of large-scale models, it has become an effective way to leverage contextual representations and the implicit knowledge embedded within pretrained models. Since most LLMs~\cite{chatgpt,llama} and MLLMs~\cite{llava} are primarily trained on plain text or natural images, layout-aware prompting and in-context learning methods~\cite{gpt3} have been proposed to adapt these models for VRD-CU.

\textbf{\textit{QueryForm}}~\cite{queryform} introduces a query-based framework for zero-shot document key information extraction. It employs a dual prompting mechanism—\textit{entity prompts} (E-prompts) and \textit{schema prompts} (S-prompts)—to transfer knowledge from large-scale, weakly annotated pretraining data (e.g., webpages) to target domains. The design ensures that the pretraining and fine-tuning objectives are well aligned, enabling consistent query-conditional predictions across both stages.
During pretraining, HTML tags are used to construct the E-prompt $\bm{e_p}$, while the S-prompt $\bm{\tilde{s}_p}$ is derived from webpage domain information. In the fine-tuning stage, $\bm{e_p}$ becomes predefined, and $\bm{s_p}$ is treated as a set of learnable vectors. Let $\bm{x}$ denote the serialized text of the input document. The prediction targets during pretraining and fine-tuning are defined as:
\begin{align}
\bm{\hat{y}} &= \mathcal{F}([\bm{s_p};\ \mathcal{E}[\bm{e_p}, \bm{x}]]), \\
\bm{y} &= \mathcal{F}([\mathcal{E}[\bm{\tilde{s}_p};\ \bm{e_p}, \bm{x}]])
\end{align}
where $\mathcal{E}$ is the feature encoder and $\mathcal{F}$ represents the remaining layers of the language model. The training objective is to minimize the cross-entropy loss between $\bm{\hat{y}}$ and $\bm{y}$.

\textbf{\textit{GenKIE}} \cite{genkie} proposes an encoder-decoder-based multimodal KIE framework to leverage prompt to adapt various datasets and better leverage multimodal information. Following \cite{layoutlm,layoutlmv2}, different encoding methods are adopted to acquire textual, layout and visual embeddings. Byte Pair Encoding is used as language pretrained backbones, and the OCR extracted document content is concatenated with predefined prompts split by the "[SEP]" token between OCR tokens and each prompt. The 2D-positional encoding introduced by \cite{layoutlm} is used to acquire the layout embedding of each OCR extracted token, and ResNet \cite{resnet} is used to extract the visual representations following \cite{wang2022ofa}. The multimodal representations are fed into an encoder to learn interactively between modalities. Prompts, inserted at the end of the encoder's textual inputs, are either template-style or question-style. For the entity extraction task, the prompt specifies the target entity type, and the decoder generates the entity value (e.g., for "Company is?", the decoder outputs the company name). For the entity-labeling task, the prompt includes the value, and the decoder provides the entity type (e.g., for "Es Kopi Rupa is [SEP]", the decoder identifies the entity type).

\textbf{\textit{ICL-D3IE}} \cite{icld3ie} is the first framework to employ LLMs with in-context learning to extract key information from VRDs using the iterative updated diverse demonstrates. Before designing the initial diverse demonstrations, the most similar $n$ training documents to the $n$ test samples need to be selected by calculating the cosine-similarity of document representations encoded by Sentence-BERT \cite{sentencebert}. Then, different types of demonstrations are introduced to integrate multiple-view context information into LLMs. Hard Demonstrations highlight the most challenging cases, are initially designed based on the incorrectly predicted cases from GPT-3 \cite{gpt3} predictions and are updated based on prediction results during the training process. Layout-aware demonstrations are created by selecting the adjacent hard segments to understand the positional relations. Formatting demonstrations are designed to guide LLMs in formatting the outputs for easy post-processing. 

\textit{\textbf{LMDX}} \cite{lmdx} designs a pipeline to use arbitrary LLMs to extract singular, repeated and hierarchical entities from VRDs. The document images are first fed into off-the-shelf OCR tools and are divided into smaller document chunks to be processed by the LLMs with accessible input length. Then, prompts are generated under XML-like tags to control the LLM's responses and mitigate hallucination. Document Representation is a prompt contains the chunk content with the coordinates of OCR lines to bring layout modality to LLMs. After that, the task description and scheme representation prompts are designed to explain the task to accomplish and determine the output format. During inference, $N$ prompts with $K$ LLMs completions are generated to sample the correct answers. 

\subsubsection{Summary of Key Information Extraction}

Several models like Chargrid \cite{chargrid} and ACP \cite{palm2019attend} have enhanced VRD-CU by integrating visual and textual information. Additionally, auxiliary tasks such as OCR, utilized by \cite{trie,ephoie}, aid in improving multimodal feature representations through joint training. However, these frameworks often rely on smaller, randomly initialized models, which produce less representative features compared to those generated by large-scale pre-trained models like LayoutLM \cite{layoutlm} and SelfDoc \cite{selfdoc}.
Documents typically exhibit specific layouts and logical structures, which has prompted many models \cite{pick,formnet,formnetv2} to adopt graph-based approaches. These methods capture spatial and logical correlations among document elements, such as key-value pairs, leading to a more comprehensive document representation.
While these frameworks have achieved improvements in document representation, their effectiveness hinges on having sufficient well-annotated training samples, which are time-consuming to acquire. This limitation has escalated the demand for few-shot \cite{laser} and zero-shot \cite{queryform} frameworks, which leverage contrastive learning and innovative attention mechanisms. Additionally, prompt learning has been applied to distill implicit knowledge from large-scale layout-aware pre-trained models \cite{bros,layoutmask} and large language models (LLMs/MLLMs) \cite{icld3ie,lmdx}.
Despite these advances, a performance gap remains between well-fine-tuned models and few/zero-shot frameworks, highlighting the ongoing challenges in VRD-CU optimization.

\subsection{Document Entity Linking} 
Documents are normally structured hierarchically, where parent-child relations always exist in various documents, such as key-value pairs in forms and section paragraphs in reports or papers. Unlike most VRD key information extraction models, which focus on recognising the semantic entity categories in a sequence tagging task ignoring the relation between entities, linking the logical associations between document semantic entity pairs has recently been of greater interest.

\subsubsection{Entity-Level Entity Linking}
Document entity linking aims to identify the relation between document entities. Some frameworks use the known entity bounding boxes, ignoring the entity recognition step and mainly focusing on exploring the relation between input document entities. 

\textbf{\textit{DocStruct}} \cite{docstruct} is the first entity linking framework that predicts hierarchical relationships between document semantic entities using multimodal features. It extracts the [CLS] token from a BERT-like encoder for textual representation ($T_e$) and uses an RNN to encode sequential RoI visual features ($V_e$) from ResNet-50. The layout feature ($P_e$), a linear projection of the bounding box coordinates $[x_1,y_1,\dots,x_4,y_4]$, is concatenated with $T_e$. The final entity representation is computed as:
\begin{gather}
\alpha = \text{Sigmoid}(W[T_e, P_e, V_e] + b), \
E = [T_e; P_e] + \alpha V_e,
\end{gather}
where $\alpha$ gates the influence of visual features, following \cite{wang2019words}. For each entity pair $(E_i, E_j)$, the probability of a parent-child link is $P_{i \rightarrow j} = E_i M E_j$, where $M$ is an asymmetric parameter matrix. Negative sampling \cite{mikolov2013distributed} is used during training to address sparsity and class imbalance.

\textbf{\textit{SERA}} (Semantic Entity Relation extraction As dependency parsing) \cite{zhang2021entity} frames entity linking as a dependency parsing task. It uses \textit{LayoutLM} \cite{layoutlm} to extract entity-level textual features $T_e$, concatenated with projected label embeddings $L_e$ to form entity representations $E = [T_e, L_e]$. Contextual encoders (e.g., Transformer, BiLSTM, Graph) process $E$, and a Biaffine parser computes relation scores:
\begin{gather}
h_i^{key} = \sigma(W^{key}E_i + b^{key}), \
h_j^{value} = \sigma(W^{value}E_j + b^{value}), \
p_b = h_i^{key} W_{b1} h_j^{value} + h_i^{key} W_{b2}.
\end{gather}
To incorporate layout, a 2D layout feature $l_{ij}$ (min width/height distance) produces a layout score $p_l = W_l l_{ij} + b_l$. The final relation score is $p = p_b + p_l$, used for binary or multi-label classification.

\textbf{\textit{KVPFormer}} \cite{kvpformer} reformulates entity linking as a QA task. It uses a Transformer-based encoder-decoder to model joint-grained representations. Each entity’s embedding $E = [T; l]$ is formed by averaging token-level features $T$ and concatenating with its label embedding $l$. A Transformer encoder with spatial attention bias models entity relations. A binary classifier first detects key entities (questions), which are passed to a DETR-style decoder \cite{carion2020end} to predict related entities (answers). For each key entity, top-K answer candidates are ranked by a sigmoid score and refined via a softmax-based coarse-to-fine strategy.

\subsubsection{Word-level Entity Linking} As acquiring entity information needs prior knowledge from manual annotation or layout analysing models, some works utilise serialised OCR-extracted sequence of words as inputs to extract the structured relations. However, as the logical relation links semantic entities, word-level frameworks must group word tokens into entities before exploring their association. 

\textbf{\textit{Carbonell et al.}} \cite{carbonell2021named} introduces a framework comprising three modules for sequentially conducting token grouping, entity labelling and relation prediction. Firstly, each text token $t$ is represented by $[L_t; T_t]$ where $L_t = [x,y,w,h]$ is the coco format bounding box coordinates of $t$ and $T_t$ is the work/representations. All tokens are fed into a token grouping GNN, $\mathcal{G}_{group}$, as a node where the edge between nodes is determined by k-NN to avoid high consumption of fully connected GNN. The $\mathcal{G}_{group}$ is trained on a link prediction task to predict the edge score between two nodes to group words, of which scores larger than a predefined threshold $\rho$. Then, the grouped words are fed into a Graph Attention Network (GAT) to use multi-head attention to aggregate words into entities and follow an MLP to conduct node classification to predict the category of each document entity. At last, another link prediction GNN, $\mathcal{G}_{link}$, is trained on edge classification based on aggregated entities. 

\textit{\textbf{SPADE}} \cite{hwang2021spatial} formulate the word-level entity linking as a spatial dependency parsing task for serializing (ordering and grouping) tokens and predicting inter-group relation between grouped tokens. Firstly, a spatial text encoder is designed to make spatial-aware attention by introducing a relative spatial vector considering relative, physical and angle aspects. During this task, two binary matrices must be predicted $M_g$ for token grouping and inter-group linking $M_l$. The vertices comprised by a entity types $\mathbb{V}$ and sequence of tokens $\mathbb{T}$ the encoded entity category and token are represented as $c$ and $t$, respectively, and the relation score between vertices $v_i \rightarrow v_j$ can be calculated by:
\begin{gather}
h_i =
\begin{cases}
c_{v_i}, & \text{for } v_i \in \mathbb{V}, \\
W_h t_{v_i}, & \text{otherwise}
\end{cases}, \\
d = W_d v_j, \\
s_0 = h_i^T W_0 d, \ s_1 = h_i^T W_1 d .
\end{gather}

The probability is acquired by $p_{ij} = \frac{\exp(s_{0,j})}{\exp(s_{0,j}) + \exp(s_{1,j})}$. An adjustable threshold is set to construct the $M_g$ or $M_l$. 

\textit{\textbf{DocTR}} \cite{doctr} formulate the entity linking as an anchor word-based entity detection and association problem. Each document entity is represented by the anchor word to convert the entity extraction and linking to word-level tasks. It contains a Deformable DETR-based vision encoder to extract multi-scale visual feature extraction. A \textit{\textbf{LayoutLM}} based language encoder is applied to encode word-level textual representations. The outputs from vision/language encoders are fed into the vision-language decoder with the language-conditional queries to conduct entity extraction and linking. The decoder queries are one-to-one mapping with language encoder inputs. The entity extraction task aims to predict whether the query underlying token level input is an anchor word and corresponding categories, while entity linking is acquired by 

\subsubsection{Summary of Entity Linking Models}

Several models like Chargrid \cite{chargrid} and ACP \cite{palm2019attend} have enhanced VRD-CU by integrating visual and textual information. Additionally, auxiliary tasks such as OCR, utilised by \cite{trie,ephoie}, aid in improving multimodal feature representations through joint training. However, these frameworks often rely on smaller, randomly initialised models, which produce less representative features compared to those generated by large-scale pretrained models like LayoutLM \cite{layoutlm} and SelfDoc \cite{selfdoc}.
Documents typically exhibit specific layouts and logical structures, which has prompted many models \cite{pick,formnet,formnetv2} to adopt graph-based approaches. These methods capture spatial and logical correlations among document elements, such as key-value pairs, leading to a more comprehensive document representation.
While these frameworks have achieved improvements in document representation, their effectiveness hinges on having sufficient well-annotated training samples, which are time-consuming to acquire. This limitation has escalated the demand for few-shot \cite{laser} and zero-shot \cite{queryform} frameworks, which leverage contrastive learning and innovative attention mechanisms. Furthermore, prompt learning has been applied to distil implicit knowledge from large-scale layout-aware pretrained models \cite{bros,layoutmask} and large language models (LLMs/MLLMs) \cite{icld3ie,lmdx}.
Despite these advances, a performance gap remains between well-fine-tuned models and few/zero-shot frameworks, highlighting the ongoing challenges in VRD-CU optimization. 

\subsection{VRD Question Answering}

Unlike key information extraction, which targets specific details within document images, answering natural language questions involves interpreting more complex intentions and requires models to facilitate interactive understanding between the queries and document representations \cite{vdoc}. The introduction of DocVQA \cite{docvqa} marked a significant shift in focus from natural scene images to text-dense, layout-aware single-page document images, establishing a benchmark in the field. As advancements have continued, demands have recently emerged for models capable of addressing more complex, multi-page scenarios \cite{mpdocvqa,pdfvqa,mmvqa}. These emerging requirements highlight the need for models to process multimodal inputs and navigate through extensive documents, reflecting user inquiries' evolving complexity and naturalness in document-based question-answering systems. This section will briefly review the SoTAs in single-page document VQA models and introduce some recently proposed multi-page document understanding solutions \cite{mpdocvqa,gram}. 

\subsubsection{Single-page VRD-QA}
Similar to key information extraction, single-page question answering (QA) on VRDs often begins with classical pretrained language models such as BERT~\cite{bert} and RoBERTa~\cite{roberta}, which perform span-based QA by extracting relevant text token sequences. In addition, general-domain vision-language models like VisualBERT~\cite{visualbert}, LXMERT~\cite{lxmert}, and ViLT~\cite{vilt} have been adopted to identify semantic entities in documents~\cite{pdfvqa}.
Beyond these plain-text or general-domain visual-language models, numerous layout-aware architectures—specifically pretrained on document-centric data (see Section~\ref{sec:multi_task})—have achieved state-of-the-art performance on single-page document QA tasks. These approaches highlight the critical role of incorporating layout information during preprocessing to improve model performance on visually structured documents.

\subsubsection{Multi-page VRD-QA}

With the growing need to retrieve answers from multi-page documents~\cite{mpdocvqa,mmvqa}, existing single-page models~\cite{layoutlm,lilt,bros} face limitations due to their 512-token input constraints. Recent advancements address this by introducing long-sequence transformers, hierarchical encoders, and page-locating modules tailored for multi-page VRD-QA.

\textbf{\textit{Hi-VT5}}~\cite{mpdocvqa} proposes a hierarchical encoder-decoder framework based on T5 \cite{t5} for multi-page generative QA. Each page is encoded independently using a T5-based multimodal encoder, incorporating layout-aware question and OCR tokens~\cite{biten2022latr}, visual patch features from DIT~\cite{dit}, and learnable page tokens. The enhanced page token representations ($P'$) are aggregated and passed to the decoder to generate answers. As Hi-VT5 uses T5 (capable of handling up to 20,480 tokens), it significantly expands the input capacity beyond traditional VRD models. To compensate for T5's lack of inherent layout awareness, masked language modeling is employed during pretraining to inject visual and spatial signals.

\textbf{\textit{GRAM}}~\cite{gram} builds on a pretrained single-page model (DocFormerV2~\cite{docformerv2}) and introduces a lightweight global encoder between layers to enable inter-page interaction. Learnable page tokens attend across pages via sparse self-attention, capturing document-level context. To align with the pretrained backbone, ALiBi~\cite{press2021train} attention bias is applied to preserve attention to these new tokens. Unlike Hi-VT5, which only decodes from page tokens, GRAM incorporates all fine-grained token embeddings during decoding. It also employs C-Former~\cite{t5} to compress cross-page information, reducing computational overhead.

\textbf{\textit{Kang et al.}}~\cite{kang2024multi} propose a two-stage framework using Pix2Struct~\cite{pix2struct} as a single-page encoder fine-tuned on DocVQA~\cite{docvqa}. A self-attention-based scoring module is trained on top of the frozen encoder to select the most relevant page based on question-page matching. The selected page is then fed into the decoder to generate the answer, ensuring efficient retrieval by focusing only on the most informative content.

Overall, these multi-page VRD-QA frameworks advance beyond single-page limitations by leveraging hierarchical encoding, inter-page attention, and selective page routing to support more complex, large-scale document understanding tasks.

\subsubsection{Summary of VRD-QA Models}
VRD-QA is a relatively new research area, introduced by DocVQA \cite{docvqa}, which focuses on answering natural language questions based on document images. Unlike key information extraction—which targets predefined key-value pairs—VRD-QA requires a deeper understanding of the entire document and its relevance to the question, demanding more comprehensive semantic and structural representations.
Pretrained VRD-CU models such as LayoutLMv2 and LayoutLMv3 \cite{layoutlmv2,layoutlmv3} perform well on single-page document QA tasks. However, they face limitations when applied to multi-page documents due to input length constraints. To address this, recent works \cite{mpdocvqa,gram} have proposed identifying the most relevant page(s) and then applying single-page QA techniques to extract answers.
Despite these advances, real-world applications often involve more complex scenarios—such as long-range dependencies and cross-page relationships—which remain open challenges in the VRD-QA field and warrant further investigation.

\subsection{Summary of Mono-Task Models}
\label{sec:mono_summary}
To achieve better performance on specific VRD-CU tasks, mono-task frameworks often adopt task-specific designs—for example, graph-based spatial encoding for KIE \cite{formnet,formnetv2} or page-locating modules for multi-page VRD-QA \cite{mpdocvqa}. While optimized for individual objectives, these specialized components often form the foundation for multi-task learning. For instance, relation-aware modeling in KIE can support entity linking, and layout-aware encoders used in QA can be adapted for structured extraction. Recognizing these shared components is essential for understanding cross-task generalization and designing unified VRD-CU systems.
Early frameworks relied heavily on hand-crafted features and rule-based heuristics, such as bounding box grouping or document-specific layout parsing. However, with the emergence of pretrained models like LayoutLM, BROS, and LiLT \cite{layoutlm,bros,lilt}, these manual designs have been largely replaced by shared multimodal encoders trained on large-scale document corpora. This shift has enabled the development of more general, instruction-tuned frameworks capable of handling multiple tasks with minimal adaptation.
Section~\ref{sec:multi_task} builds on this trend, introducing multi-task models that unify these components through shared architectures, input granularity settings, and cross-task generalization strategies.

\section{Multi-Task VRD Understanding Models}
\begin{figure*}[htb]
\footnotesize
\begin{forest}
    for tree={
        forked edges,
        grow'=0,
        draw,
        rounded corners,
        node options={align=center,},
        text width=2.7cm,
        s sep=6pt,
        calign=child edge, 
        calign child=(n_children()+1)/2
    }
    [\textbf{Multi-task based models}, for tree={fill=brown!45}
        [\textbf{Fine-grained} \textbf{Pretrained} Models  (\textbf{Encoder-only}), for tree={fill=violet!40}
            [\textbf{Layout-aware} Pretrained Models, for tree={fill=violet!20}
                [LayoutLM \cite{layoutlm};
                BROS\cite{bros};
                LiLT \cite{lilt};
                XDoc \cite{xdoc};
                LayoutMask \cite{layoutmask};
                StructuralLM \cite{layoutlm}, for tree={fill=violet!10}]
            ]
            [\textbf{Visual-integrated} \textbf{Pretrained} Models, for tree={fill=violet!20}
                [LayoutLMv2\cite{layoutlmv2};
                LayoutXLM\cite{layoutxlm};
                DocFormer\cite{docformer};
                LayoutLMv3\cite{layoutlmv3}, for tree={fill=violet!10}]
            ]
        ]
        [\textbf{Coarse and Joint-grained} \textbf{Pretrained} Models (Encoder-only), for tree={fill=pink!40}
            [\textbf{Coarse-grained} Models, for tree={fill=pink!20}
                [SelfDoc\cite{selfdoc};
                UniDoc\cite{udoc}], for tree={fill=pink!10}
            ]
            [\textbf{Joint-grained} Models, for tree={fill=pink!20}
                [StrucText\cite{structext};
                Fast-StrucText\cite{faststructext};
                MGDoc \cite{mgdoc};
                WUKONG-READER \cite{wukong};
                GeoLayoutLM \cite{geolayoutlm}, for tree={fill=pink!10}]
            ]
        ]
        [\textbf{Encoder-Decoder} \textbf{Pretrained} Frameworks, for tree={fill=red!40}
            [\textbf{OCR-dependent} Encoder-Decoder Frameworks, for tree={fill=red!20}
                [TiLT\cite{selfdoc};
                UDOP\cite{udop};
                DocFormerv2\cite{docformerv2};
                ViTLP\cite{vitlp}, for tree={fill=red!10}]
            ]
            [\textbf{OCR-free} Frameworks, for tree={fill=red!20}
                [Donut\cite{donut};
                Dessurt\cite{dessurt};
                SeRum \cite{serum};
                StructTextV2 \cite{structextv2}, for tree={fill=red!10}]
            ]
            [\textbf{LLM-based} Frameworks, for tree={fill=red!20}
            [HRVDA\cite{hrvda};
            LayoutLLM\cite{layoutllm}, for tree={fill=red!10}]
            ]
        ]
        [\textbf{Non-Pretrained} Frameworks, for tree={fill=yellow!40}
            [CALM\cite{calm};
                LayoutGCN\cite{layoutgcn},for tree={fill=yellow!20}
            ]
        ]
    ]
\end{forest}

\caption{Multi-task visually rich document understanding frameworks.}
\label{fig:multi_task_framework}
\end{figure*}

\label{sec:multi_task}
As illustrated in Figure~\ref{fig:multi_task_framework}, recent frameworks for VRD-CU increasingly adopt multi-task learning designs. These frameworks can be broadly categorized based on their \textbf{model architecture} and \textbf{input granularity}, which are key to determining their capability and generalizability.
In terms of \textbf{architecture}, models are typically classified as \textbf{Encoder-only}, \textbf{Encoder-Decoder}. Encoder-only models utilize transformer-based encoders to capture multimodal document features and are often self-supervised pretrained on large-scale datasets. Encoder-Decoder models, usually generation-oriented, consist of an encoder that processes visual or textual input and a decoder that produces structured output. These can be further divided into \textbf{OCR-dependent} models, which require external text input, and \textbf{OCR-free} models that operate directly on document images. Non-pretrained models, in contrast, are task-specific and avoid extensive pretraining, often using lightweight or graph-based architectures.
\textbf{Input granularity} distinguishes how models represent document information. \textbf{Fine-grained} models encode detailed features at the token or region level, preserving spatial and textual fidelity. \textbf{Coarse-grained} models abstract over larger structural units, such as blocks or paragraphs, to capture high-level semantics. \textbf{Joint-grained} approaches integrate both levels to support comprehensive document understanding.

\subsection{Fine-grained Pretrained Models}
Inspired by BERT-style pretrained models, many researchers have proposed effective methods to integrate layout and visual information into models, aiming to enhance the comprehensiveness of textual token representations.

\subsubsection{Layout-aware Pretrained Language Modeling} Understanding layout structure and spatial correlations between textual tokens can yield a more comprehensive representation of documents, going beyond what plain-text input offers. To this end, various methods have been proposed to encode layout features. These methods, combined with tailored pretraining tasks, enable models to capture layout-aware information better and effectively fuse textual and layout features.

\textbf{\textit{LayoutLM}} \cite{layoutlm} is the first pretrained document understanding model by leveraging textual and layout information in the pretraining stage. BERT architecture is the backbone and 2-D positional embedding \footnote{Please refer to Section~\ref{sec:positional_feature} for more detailed information on 2-D positional encoding} with textual information is used to pretrain on IIT-CDIP Test Collection 1.0. Two specific pretraining tasks are first introduced, named \textit{Masked Visual-Language Model} (MVLM) and \textit{Multi-label Document Classification}, to generate layout-aware textual representation and more comprehensive document representation, respectively. Like Masked Language Modeling adopted by most pretrained language models, MVLM randomly masks some input tokens but keeps the corresponding 2-D position embeddings to predict masked tokens to ensure the pretrained model is aware of the spatial relations between input tokens. MDC is a supervised pretraining task to predict the input document types (e.g. forms, exam papers, academic papers) that generate more comprehensive document-level representations. The fine-tuned LayoutLM could perform much better than textual-only frameworks on key information extraction \cite{funsd,sroie} and document classification \cite{rvlcdip}.

\textbf{\textit{BROS}} \cite{bros} proposes a pretrained VRDU model that captures the continuous nature of 2D space through a novel positional encoding and a correlation-aware attention mechanism. Given a token's bounding box coordinates $[x_1, y_1, \dots, x_4, y_4]$, the positional embedding is computed as:
\[
\text{pos}_t = W_{p1}p_1 + W_{p2}p_2 + W_{p3}p_3 + W_{p4}p_4,
\]
where each $p_i = \mathcal{F}_{\sin}(x_i) \oplus \mathcal{F}_{\sin}(y_i)$ and $W_{p*} \in \mathbb{R}^{2d_{\text{pos}} \times d}$.

To replace vanilla self-attention, BROS introduces an attention score $\alpha_{ij}$ that captures intra- and inter-modal interactions between textual and positional features:
\begin{align}
    \alpha_{ij} =\ 
    &\underbrace{\left( W^q T_i \right)^\top \left( W^q T_j \right)}_{\text{text-text}} + 
    \underbrace{\left( W^q T_i \circ W^{\text{pos}} \text{pos}_i \right)^\top \left( W^{\text{pos}} \text{pos}_j \right)}_{\text{text-pos}} \notag \\
    &+ \underbrace{\left( W'^{\text{pos}} \text{pos}_i \right)^\top \left( W'^{\text{pos}} \text{pos}_j \right)}_{\text{pos-pos}}.
\end{align}

Inspired by SpanBERT \cite{joshi2020spanbert}, BROS also adopts an area-masked language modeling task, masking tokens within randomly sampled rectangular regions to enhance span-level understanding.

\textbf{\textit{StructuralLM}} \cite{structurallm} is the first VRDU model to leverage image patches (termed "cells") to group tokens and perform patch-level pretraining. Built on BERT, it computes multimodal patch representations $P$ and introduces two pretraining tasks: \textit{Masked Visual-Language Modeling} (MVLM) and \textit{Cell Position Classification} (CPC).
Each patch is defined by a bounding box $(x_0, y_0, x_1, y_1)$ and encoded using the 2-D positional encoding from LayoutLM. Tokens $\{t_1, t_2, \dots, t_n\}$ within a patch share this 2-D position. A token $t_i$ is represented by: $t_i = T_i + \text{pos}^{2D}_{t_i} + \text{pos}^{1D}_i,$ where $T_i$ is the token embedding, $\text{pos}^{2D}_{t_i}$ is the patch-level layout embedding, and $\text{pos}^{1D}_i$ is the token's sequence position.
MVLM masks tokens using patch-level layout embeddings, while CPC predicts the area index of a token among $N$ equally divided regions. Together, these tasks encourage the model to capture spatial dependencies at the patch level.

\textbf{\textit{LiLT}} \cite{lilt} proposes a \textbf{language-independent} layout Transformer for mono- and multilingual document understanding. Text and layout are encoded separately and fused during pretraining, with their representations concatenated for downstream tasks. Each token $t_j$ is represented as:
\[
T_j = T_j + \text{pos}_{t_j}^{1D} + \text{pos}_{t_j}^{2D}.
\]

Unlike LayoutLM, layout features are normalised to $[0,1000]$, and six layout attributes $[x_0, y_0, x_1, y_1, w, h]$ are encoded via:
\[
L = W_L (W_x x_0 \oplus W_y y_0 \oplus W_x x_1 \oplus W_y y_1 \oplus W_w w \oplus W_h h) + \text{pos}_L,
\]
where $W_L \in \mathbb{R}^{6d'_L \times d_L}$ and $W_*$ are learnable projections.
Text and layout embeddings are processed via two sub-models, with a bi-directional attention complementation mechanism (BiACM) enhancing cross-modal interaction. Three pretraining tasks are introduced: MVLM, Key Point Location (predicting token area indices), and a text-layout alignment task (predicting whether text-layout pairs match). LiLT achieves strong results on multilingual document benchmarks \cite{ephoie,layoutxlm}.

\textbf{\textit{XDoc}}~\cite{xdoc} presents a unified architecture that handles diverse input formats—plain text, documents, and web data. It applies different encoding strategies: BERT-style for plain text; LayoutLM-style for documents with adapted 2D box encoding; and XPath-based embedding for web structures. All formats are pretrained with masked language modeling (MLM) and fine-tuned on corresponding benchmarks.

\textbf{\textit{LayoutMask}}~\cite{layoutmask} targets the issue of improper reading order in OCR-dependent systems. It removes visual inputs and replaces global 1D positional encoding with a segment-based local variant, restarting the order within each OCR segment. The model introduces two pretraining objectives: (1) word-level MVLM with higher masking probabilities for segment boundaries, and (2) \textit{Masked Position Modeling}, predicting the 2D coordinates of masked words to reinforce layout understanding.

\subsubsection{Visual Integrated Models} 
\label{sec:vision_integrated}
Integrating visual cues with textual and layout information during pretraining significantly enhances a model’s ability to capture rich and comprehensive document semantics. While earlier frameworks primarily focused on text and layout, recent approaches extend these by introducing visual-text alignment tasks to strengthen cross-modal understanding. This integration allows models to better interpret the intricate relationships among visual elements, textual content, and spatial structures within documents.
Existing vision-integrated models can be broadly categorized based on how visual features are extracted: (1) \textit{RoI-based models}, which employ region-of-interest (RoI) alignment using CNN backbones to generate high-level visual representations; and (2) \textit{pixel-based models}, which directly extract pixel-level features via Vision Transformers.

\textbf{\textit{LayoutLMv2}}~\cite{layoutlmv2} is the first pretrained model to integrate textual, layout, and visual modalities into a unified multimodal Transformer. Building upon LayoutLM, it introduces a trainable ResNeXt-FPN visual encoder whose outputs are projected to the same dimensional space as textual embeddings. Each modality is assigned a segment ID (e.g., $seg_t$, $seg_v$) added to its embeddings. Spatial-aware self-attention is adopted, incorporating 1D and 2D relative positional biases:
\begin{align}
b^{1D} &= W_{b_{1D}}(j - i), \\
b^{2D} &= W_{b^{2D}_x}(x0_i - x0_j) + W_{b^{2D}_y}(y0_i - y0_j)
\end{align}
Three pretraining tasks are used: MVLM, \textit{Text-Image Alignment} (TIA), and \textit{Text-Image Matching} (TIM), enhancing cross-modal understanding.

\textbf{\textit{LayoutXLM}}~\cite{layoutxlm} extends LayoutLMv2 to multilingual settings with the same architecture but trains on 22M multilingual PDF and 8M scanned English documents. The MVLM task is adapted for multilingual contexts.

\textbf{\textit{DocFormer}}~\cite{docformer} proposes a multimodal encoder architecture combining LayoutLM-based text encoding and CNN-based visual encoding via ResNet-50. Textual ($\mathbb{T} \in \mathbb{R}^{d \times N}$) and visual ($\mathbb{V} \in \mathbb{R}^{d \times N}$) features are enhanced with 2D and 1D positional embeddings. A novel self-attention score $\alpha_{ij}^{v}$ includes multiple spatial and semantic biases:
\begin{align}
\alpha_{ij}^{v} &= (W^K_v V_j)^\top (W^Q_v V_i) + (pos_{ij})^\top W^Q_v V_i + (pos_{ij})^\top W^K_v V_j \notag \\
&\quad + (W^Q_s pos_{ij}^{2d})^\top (W^K_s pos_{ij}^{2d})
\end{align}
Two modalities are fused at each encoder layer via addition: $\mathbb{M}_l = \mathbb{T}_l + \mathbb{V}_l$.

\textbf{\textit{LayoutLMv3}}\cite{layoutlmv3} enhances model efficiency by eliminating heavy CNNs and adopting a pure Transformer-based vision encoder. Inspired by ViT, it divides document images into patches and projects them linearly with learnable 1D positional encodings. Layout information is encoded using segment-level bounding boxes with 2D positional embeddings. In addition to masked language modeling (MLM), LayoutLMv3 introduces two new pretraining tasks: \textit{Masked Image Modeling} (MIM), where 40\% of image patches are masked and reconstructed using discrete tokens with a cross-entropy loss\cite{bao2021beit}; and \textit{Word-Patch Alignment} (WPA), a binary classification objective that predicts whether a text token and image patch pair is aligned. Together, these tasks promote fine-grained multimodal alignment between visual and textual features.


\subsection{Coarse and Joint-grained Pretrained Models}
Fine-grained models achieve state-of-the-art performance on many downstream tasks but face challenges with input length limitations and capturing document image layout and logical arrangement. To address these issues, coarse-grained or joint-grained frameworks have been introduced. To mitigate these limitations, these frameworks leverage multimodal information from document semantic entities such as paragraphs, tables, and textlines.
\subsubsection{Coarse-grained Frameworks}
\textbf{\textit{SelfDoc}} \cite{selfdoc} is the first pretrained VRD-CU model leveraging coarse-grained document semantic entity for various understanding tasks. Unlike fine-grained OCR-based models, it uses Faster-RCNN to extract Regions of Interest (RoIs) from semantic entities, reducing input length and improving efficiency on dense, long documents. Visual embeddings are derived from RoIs, while Sentence-BERT \cite{sentencebert} generates textual embeddings from the OCR-extracted text. These are fed into separate BERT-style encoders to model intra-modality context. A cross-modality encoder with cross-attention layers enables inter-modality learning. During pretraining, random masking is applied to either text or visual tokens. A modality-adaptive attention mechanism further adjusts the weight of visual and textual cues dynamically for robust entity-level representations.

\textbf{\textit{UniDoc}} \cite{udoc} is an coarse-grained model with a trainable image encoder using RoI-Align \cite{maskrcnn} to extract visual features and a novel cross-attention mechanism for multimodal fusion. It computes textual embeddings by combining average word embeddings and linearly projected bounding box coordinates. Product quantization \cite{jegou2010product} is used to discretize RoI features into a finite visual codebook. A \textit{Gated Cross-Attention} module is applied after multi-head fusion, where concatenated $[V:T]$ features are passed through a non-linear layer to generate modality-aware attention biases ($\beta_v$, $\beta_t$). UniDoc employs three pretraining tasks: Masked Sentence Modelling, Visual Contrastive Learning, and Vision-Language Alignment (adapted from LayoutLMv2 \cite{layoutlmv2})—the last aligning image-text pairs at the entity level instead of region-level.

\subsubsection{Joint-grained Frameworks}

\textbf{\textit{StrucText}} \cite{structext} is a multimodal pretrained VRD-CU model that integrates fine-grained text and coarse-grained visual features to capture rich geometric and semantic information. Each text token is assigned a layout embedding $L = W_l[x_0, y_0, x_1, y_1, w, h]$ using $W_l \in \mathbb{R}^{6 \times N}$, and visual features are extracted from entities via a pretrained ResNet50-FPN. A segment-ID embedding aligns text and visual representations within the same entity. Three self-supervised tasks guide training: \textit{Masked Visual-Language Modeling} (MVLM), \textit{Segment Length Prediction} (SLP), and \textit{Paired Box Direction} (PBD). SLP predicts entity lengths to encourage multimodal fusion, while PBD captures spatial relationships between entities.

\textbf{\textit{Fast-StructText}} \cite{faststructext} builds upon StrucText to improve efficiency and representation power. It uses a more compact layout encoding $[x_0, y_0, x_1, y_1]$, an hourglass transformer with Merging and Extension blocks for token compression and recovery, and a \textit{Symmetry Cross-Attention} (SCA) mechanism for enhanced modality interaction. Merging blocks apply weighted pooling to downsample token sequences, while extension blocks use repeat up-sampling. Fast-StructText introduces new self-supervised tasks alongside MVLM: \textit{Graph-based Token Relation} (GTR) for learning spatial relations, \textit{Sentence Order Prediction} (SOP) for semantic continuity, and \textit{Text-Image Alignment} to strengthen cross-modal grounding.

\textbf{\textit{MGDoc}} \cite{mgdoc} is the first multimodal, multi-granular pretrained framework designed to enhance inter-granular learning and cross-modal fusion through multi-granular and cross-modal attention mechanisms. It uses pretrained language encoders \cite{sentencebert} and vision backbones \cite{resnet} to extract features across different granularities—from word to page. These features are combined with positional embeddings \cite{faststructext} and modality-type embeddings \cite{structext} to form the final input representations. To capture hierarchical relations among pages, entities, and words, two attention biases are added: a binary hierarchical bias indicating inclusion relations, and a relation bias based on relative bounding box positions. Cross-modal attention is used to integrate information across modalities. Pretraining involves three tasks: Masked Text Modeling (MTM), Masked Vision Modeling (MVM), and Multi-Granularity Modeling. MTM and MVM mask and reconstruct multi-granular representations using mean absolute error, while token-entity linking is modeled via dot-product similarity to promote multi-granular alignment.

\textbf{\textit{WUKONG}} \cite{wukong} utilizes fine-grained inputs (tokens with bounding boxes and document images) but enhances learning through coarse-grained self-supervised tasks. A Mask-RCNN backbone extracts visual features from textlines via RoIHead, while RoBERTa \cite{wei2020robust} encodes token representations in two stages: the first six layers for initial textual features (augmented with layout features \cite{layoutlmv2}), and the remaining layers for multimodal fusion with visual features. Four pretraining tasks are designed: Masked Language Modeling (MLM), Textline-Region Contrastive Learning (TRC), Masked Region Modeling (MRM), and Textline Grid Matching (TCM). TRC enhances entity-level cross-modal alignment via contrastive learning \cite{yao2021filip}. MRM randomly masks 15\% of textlines and predicts their visual embeddings \cite{selfdoc}. TCM divides the image into grids and predicts the grid location for masked textlines, strengthening layout awareness. Each task’s loss is weighted with scaling parameters during training.

\textbf{\textit{GeoLayoutLM}} \cite{geolayoutlm} is a sophisticated multimodal framework that distinctively incorporates geometric information through specialised pretraining tasks and the development of innovative relation heads. Inspired by the dual-stream structure of METER and \textit{\textbf{SelfDoc}} \cite{selfdoc}, GeoLayoutLM features separate vision and text-layout modules coupled with interactive co-attention layers that enhance the integration of visual and textual data. The model introduces two advanced relation heads—the Coarse Relation Prediction (CRP) head and the Relation Feature Enhancement (RFE) head—which refine relation feature representation crucial for both pretraining and fine-tuning phases. The pretraining regimen includes tasks designed to understand geometric relationships, such as GeoPair, GeoMPair, and GeoTriplet, aiding the model in grasping the complex dynamics of document layouts. During fine-tuning, the model utilises pretrained parameters to optimise both semantic entity recognition and relation extraction tasks, employing a novel inference technique that enhances relation pair selection accuracy by focusing on the most probable relationships and minimizing variance among potential options. 

\subsection{Encoder-Decoder Pretrained Frameworks}

In addition to the above encoder-only frameworks, researchers have proposed encoder-decoder pretrained models that often approach tasks like Key Information Extraction (KIE) or Visual Question Answering (VQA) in a generative style. Addressing limitations of OCR-dependent frameworks, such as accumulated OCR errors and incorrect reading orders, OCR-free models have been introduced for end-to-end VRD-CU. 

\subsubsection{OCR-dependent Encoder-Decoder Frameworks}

\textbf{\textit{TILT}} \cite{tilt} is a T5-based transformer encoder-decoder architecture enhanced with a relative spatial bias in the self-attention mechanism to acquire fine-grained token representations. It incorporates encoding methods that apply relative sequence input bias and capture horizontal and vertical distance biases in the attention scores. A U-Net-based framework is applied to extract the fixed-size feature maps fed into the encoder together. They follow T5 pretraining strategies on RVL-CIDP dataset \cite{rvlcdip} but use a salient span masking scheme adopted by \cite{t5,guu2020retrieval}. As the first encoder-decoder framework, \textbf{\textit{TILT}} requires off-the-shelf OCR tools to acquire the textual token sequence. To conduct VRD content understanding end-to-end, some OCR-free frameworks are proposed to solve the limitations of OCR-dependent models.

\textbf{\textit{UDOP}} \cite{udop} introduces an encoder-decoder framework for document understanding, built on a ViT-based model \cite{vit} and inspired by LayoutLMv3 \cite{layoutlmv3}. It enhances textual token embeddings by summing them with aligned image patch embeddings when the token's bounding box center overlaps with a patch. TILT-style positional biases are used, while ID embeddings are omitted. The decoder consists of two cross-attending modules: a bidirectional Transformer for text-layout decoding and an MAE-style vision decoder. Pretraining includes both self-supervised tasks (e.g., masked text-layout and image reconstruction) and supervised tasks using benchmark datasets. Additional objectives include layout moulding (predicting token group positions) and visual text recognition (identifying text at specific locations). Supervised fine-tuning is performed on datasets for classification \cite{rvlcdip}, KIE \cite{nda}, VQA \cite{visualmrc,docvqa}, and layout analysis \cite{publaynet}.

\textbf{\textit{DocFormerv2}} \cite{docformerv2} is an encoder-decoder transformer architecture that uses multimodal (visual, textual and positional) to enhance the multimodal understanding and layout-aware language decoder to predict the predictions. The patched image pixels are fed into the convolutional and linear layers to get down-sampled patch embedding. The textual embeddings are acquired by linear projected token one-hot encoding. Both visual and textual embeddings are summed with the 2D-positional encoding \cite{layoutlm} of Patches and linear project bbox embedding ($[x0,y0,x1,y1]$) \cite{lilt}, respectively. Two encoder-based Token to Line (T2L), Token to Grid (T2G), and one decoder-based self-supervised learning task, MLM, are proposed to enable multimodal feature interactive learning. T2L aims to improve the relative position understanding between tokens by predicting the number of textlines between two randomly selected tokens. For improving the layout and structure, understanding needs to split the image into $m\times n$ grids to predict the located grid number of each OCR token. For the decoder MLM, the spatial feature of each masked text token is masked as well, and the other setup follows T5 \cite{t5}. 

\textbf{\textit{ViTLP}} \cite{vitlp} presents an encoder-decoder framework for OCR and document understanding. It employs a ViT-based vision encoder to extract image patch representations, which are decoded autoregressively into text and layout sequences. A special “[LOC]” token encodes bounding box coordinates $[x_0, y_0, x_1, y_1]$ to reduce layout sequence length. To guide generation, “[BOS]” and “[CONT]” tokens are used—representing a token pair {t1, t2} as “[BOS], t1, [LOC], t2, [LOC]”. The decoder has hierarchical heads: the text head predicts the next token using the full sequence, while the layout head predicts bounding boxes only from “[LOC]” tokens. The “[CONT]” token enables variable-length generation by continuing decoding until “[END]”, based on the prefix token ratio.

\subsubsection{OCR-free Pretrained Frameworks}
\textbf{\textit{Donut}} \cite{donut} is the first OCR-free VRD understanding model to understand and extract key information from input document images. Donut contains a Swin Transformer-based visual encoder to encode the input document image into image patches, which are then fed into a BART-based \cite{bart} decoder pretrained on multi-lingual scenarios. During model training, teacher forcing is applied, and in the test stage, inspired by GPT-3 \cite{gpt3}, prompts with special identify tokens are fed into the model for different downstream tasks. The output token sequence contains the special tokens $<START\_*>$ and $<END\_*>$ to identify the type of tasks and struct predict entities. The wrongly structured entity will be treated as an empty prediction. The model is pretrained on next-token prediction on the IIT-CDIP dataset and a Synthetic Dataset, which can be interpreted as a pseudo-OCR task. Similar to Donut, \textbf{\textit{Dessurt}} \cite{dessurt} also proposes an encoder-decoder architecture but a different decoding process. Instead of using BART, the cross-attention used in Dessurt attends to all visual, query and previously generated textual information and is pretrained on more synthetic datasets with different font sizes and handwritten content. 

\textbf{\textit{SeRum}} \cite{serum} introduces an end-to-end OCR-free framework that decodes text based on local visual cues and spatial focus. A Swin Transformer extracts image patch features, which are refined through a Query Decoder to form vision-enhanced query representations, subsequently used by a Text Decoder for auto-regressive text generation. To better capture relevant regions, a Content-aware Token Merge module selects top-$K$ visual tokens based on their correlation with the query, while unselected background tokens enhance the foreground via attention. Three pretraining tasks are proposed: Query-to-Segmentation (Q2S), which predicts instance masks for text regions; Text-to-Segmentation (T2S), which segments based on the decoder’s output; and Segmentation-to-Text (S2T), which generates text from segmented features. These tasks jointly enhance SeRum’s ability to detect, align, and generate text directly from document images.

\textbf{\textit{StrucTextV2}} \cite{structextv2} is an end-to-end structure that uses image-only input to conduct several downstream tasks. It contains a CNN-based visual extractor with FPN strategies \cite{lin2017feature} and follows ViT \cite{vit} to get linear projected flattened patch-level representations. The patch token embeddings serve as the input to the Transformer encoder to enhance the contextually semantic representations. Then, the lightweight fusion network is applied to generate the final representations and fed into two branches during pretraining: made language Modeling (MLM) and Masked Image Modeling (MIM). Instead of using text inputs when MLM is used by other models \cite{bert}, a portion of the text regions are masked with RGB values $[255,255,255]$ randomly with a 2-layer MLP decoder to predict the masked token. MIM masks the rectangular text regions and predicts the RGB values of the missing pixels to improve the document representations. Except for the global average pooled FPN fused visual representations, the MLM-generated hidden state of each text region is concatenated and fed into a Fully Convolutional New York to get the regressed masked missing pixel values. 

\subsubsection{LLM-based Frameworks} 

\label{sec:llm_based_frameworks}
With the rapid development of LLMs/MLLMs, many frameworks are adopted to tackle document-based tasks. Both open-source \cite{qwen,llava} or close-source frameworks achieved promising performance on many benchmark datasets. This section introduces frameworks specifically proposed to address document related techniques with novel architectures. The trend of applying general domain LLMs/MLLMs in VRD-CU are discussed in Section~\ref{sec:llms}.

\textbf{\textit{\revise{DocLLM}}} \revise{\cite{docllm} proposes to address VRD-CU tasks by designing a disentangle spatial attention mechanism to capture alignments between text and layout modalities instead of integrating heavy visual backbones.  
 The attention score between tokens $i$ and $j$ is computed as: 
\begin{gather}
 A_{i,j} = Q^t_i (K^t_j)^\top + \lambda_{t,s} Q^t_i (K^s_j)^\top + \lambda_{s,t} Q^s_i (K^t_j)^\top + \lambda_{s,s} Q^s_i (K^s_j)^\top
\end{gather}
 where $Q^t$, $K^t$ and $Q^s$, $K^s$ are projections from textual and spatial embeddings, respectively. Pretraining adopts an autoregressive block infilling objective to better handle irregular layouts, where coherent OCR-derived blocks are masked and predicted given both prefix and suffix context. DocLLM is then instruction-tuned on 16 datasets covering four tasks (VQA, KIE, etc) using natural prompts, enabling strong generalization without vision encoders.}

\textbf{\textit{\revise{LapDoc}}} \cite{lapdoc} \revise{enhances LLM-based VRD-CU framework by introducing a Layout-Aware Positional (LAP) embedding mechanism. It combines a learnable layout embedding $e_l$ with the standard sinusoidal token positional encoding $e_p$, forming a unified input representation $x = e_t + e_p + e_l$, where $e_t$ is the text embedding. The LAP encoder maps 2D bounding boxes into embeddings that capture spatial structure while preserving the autoregressive training scheme. Pretraining is conducted using standard language modeling on 240M document pages from the GLaMM corpus, integrating both text and layout features. Instruction tuning is performed on 9 datasets spanning four tasks—VQA, classification, KIE, and NLI—using task-specific prompt templates to guide model behavior, thereby enabling generalization across diverse document understanding tasks.}

\revise{Some positional encoding strategies are proposed to enhance LLM/MLLM-based VRD-CU applications. \textbf{\textit{LayTokenLLM}} introduces a layout token that summarizes the spatial configuration of each text segment using an attention-based layout tokenizer. This layout token is inserted alongside the text tokens and shares the same positional index as the first token in the segment, allowing the model to integrate layout cues without disrupting the sequential language modeling objective. By maintaining a full 100\% T-Ratio (i.e., all positions contribute to text prediction), LayTokenLLM ensures efficient learning and inference, outperforming earlier layout-as-token baselines. In parallel, \textbf{\textit{Group Position Embedding (GPE)}} proposes a complementary approach by assigning different spatial features—such as bounding box coordinates and reading order—to distinct attention head groups within rotary or sinusoidal positional encodings. This CoT-style decomposition enables the model to reason over multiple layout dimensions independently, enhancing spatial awareness without altering the model architecture. While LayTokenLLM focuses on embedding layout into the input sequence, GPE injects layout awareness through grouped attention, and both methods demonstrate strong performance across document benchmarks with minimal architectural overhead.}

\revise{\textit{\textbf{DocLayLLM}} incorporates explicit visual information by encoding image patches of document regions into external memory via a vision encoder (e.g., DiT), enabling stronger grounding in complex layouts. Crucially, it introduces Chain-of-Thought (CoT) prompting by fine-tuning on multi-step reasoning sequences, which guide the model to generate intermediate layout-aware explanations before predicting the final answer. This visual CoT-enhanced framework significantly boosts multi-hop document reasoning and better captures hierarchical and spatial relationships, outperforming prior text-only models on VQA, KIE, and NLI benchmarks.}

\textbf{\textit{LayoutLLM}} \cite{layoutllm} introduces an LLM/MLLM-based approach integrated with a pretrained document understanding model to better fuse multimodal information. The input document's visual, textual, and layout information and any question text are encoded by a pretrained LayoutLMv3 \cite{layoutlmv3} encoder and projected into the same embedding space as the adopted LLM, Vicuna-7B-v1.5 \cite{zheng2024judging}. The method incorporates layout-aware pretraining tasks at three levels: document-level (e.g., document summarization), region-level (e.g., layout analysis), and segment-level (e.g., MVLM). These tasks enable the model to achieve comprehensive document understanding. Additionally, a novel module called LayoutCoT is designed to help LayoutLLM focus on question-relevant regions and generate accurate answers through intermediate steps. GPT-3.5-turbo \cite{chatgpt} is used to prepare the dataset for document summarization training and to construct LayoutCoT training data.

\textbf{\textit{HRVDA}} \cite{hrvda} aims to propose a MLLM accepting high-resolution image inputs to conduct fine-grained information extraction from VRDs. A swin-transformer \cite{liu2021swin} is used to encode document images into image patch tokens. A pluggable content detector then identifies visual tokens that contain relevant document content information. Following this, a content filtering mechanism performs token pruning to remove irrelevant tokens. The remaining encoded visual tokens are processed through an MLP to ensure consistency with the LLM embedding space dimensions. These pruned tokens are then fused with instruction features, allowing further filtering of tokens irrelevant to the instructions. The final streamlined set of visual tokens and instructions is fed into the LLM, which generates the corresponding responses.

\subsection{Non-Pretrained Frameworks}

\textbf{\textit{CALM}} \cite{calm} introduces a common-sense augment document understanding framework to understand the query and extrapolate answers not contained in the context of the input document image. They follow LayoutLMv2 \cite{layoutlmv2} to encode input document multimodal representations. The textual token embeddings are fed into a Document Purifier component to merge the tokens $\{t_1,\dots,t_n\}$ belonging to one entity type $N$ to one Upper Layer token $\hat{c}$ by applying average pooling of $\hat{c} = AvePool(t_1,\dots,t_n)$. Each Upper Layer token is concatenated with the commence augmented on ConceptNet NumberBatch \cite{speer2017conceptnet} entity word vector $c'$ to get the final entity representation $c = concat(\hat{c},c')$. A similar Question-Purifier is applied to use common-sense knowledge to enhance the question representation. Then, with the assistance of ConceptNet, relevant common-sense knowledge is recalled based on the common-sense representation of both documents and queries. By considering the predicted question-answer relationship, a final self-attentive graph convolutional network following \cite{zhu2021mucko} is proposed to address document reasoning tasks more effectively.

 \textbf{\textit{LayoutGCN}} \cite{layoutgcn} proposes a lightweight and effective model which contains a fully connected graph where text blocks are nodes and edges connect every two blocks. The model architecture includes a TextCNN-based \cite{textcnn} encoder to encode N-gram textual embeddings, a linear trainable layout encoder to project the normalised bbox coordinates into hyperspace following other layout-aware models, and a visual encoder (CSP-Darknet \cite{wang2020cspnet} for document image features). These features are integrated using a Graph Convolution Network (GCN) to capture relationships between nodes. The final node representation combines text, layout, and visual information, benefiting various VRD-CU tasks. 

\textbf{\textit{XYLayoutLM}} \cite{xylayoutlm} builds on \textbf{\textit{LayoutXLM}} \cite{layoutxlm} by introducing two key modules: an Augmented XY Cut and a Dilated Conditional Position Encoding (DCPE). The Augmented XY Cut improves traditional XY Cut \cite{nagy1984hierarchical} by incorporating adjustable thresholds ($\lambda_x, \lambda_y$) and a shift factor ($\theta$) to refine token clustering and correct OCR-induced reading order errors through recursive XY Tree-based segmentation. DCPE addresses the limitations of Conditional Position Encoding (CPE) \cite{chu2022conditional} by separately modeling textual and visual modalities. It uses 1D convolutions for textual tokens to capture local sequence structure and dilated convolutions \cite{yu2015multi} to encode long-range dependencies without added complexity, improving multimodal representation learning in document understanding tasks.

\subsection{Summary of Multi-Task Frameworks}
Various models are proposed to enhance document representations for VRD-CU tasks by leveraging pretrained language models to enrich text token sequences with layout information through positional encoding \cite{layoutlm}, attention mechanisms \cite{lilt}, and layout-aware tasks \cite{layoutmask}. However, VRDs contain rich visual details like font, texture, and colour and visually complex entities such as tables, charts, and photos. Many models \cite{layoutlmv2,layoutlmv3,docformer} integrate visual cues to enhance fine-grained document features, but their quadratic time and space complexity pose challenges for handling long sequences in multi-page document understanding \cite{mmvqa}.
Fine-grained models excel but struggle with capturing layout and structural details from document images. Coarse-grained frameworks \cite{selfdoc,udoc} mitigate fine-grained limitations by leveraging entity-level multimodal information, yet compressing diverse entity aspects into a single dense vector risks losing information \cite{3mvrd}. Joint-grained frameworks \cite{structext,faststructext,mgdoc,wukong,geolayoutlm} integrate multi-grained information to produce comprehensive representations. Non-pretrained models leverage external knowledge \cite{calm} or lightweight networks \cite{layoutgcn} to rival large-scale pretrained frameworks' performance. Most document understanding models \cite{layoutlm,layoutlmv3,lilt,selfdoc} rely on off-the-shelf OCR tools for text extraction, which can be susceptible to OCR quality issues and incorrect reading orders. OCR-free frameworks directly process document images to mitigate these limitations; However, these frameworks may exhibit sub-optimal performance compared to methods using established OCR tools with additional resource consumption.

\section{Visually Rich Document Content Understanding Datasets} \label{sec:dataset_metric}

Based on the downstream tasks differences, we summarise the Key Information Extraction with Entity Linking datasets and Visually Rich Document Question Answering dataset in Section~\ref{sec:kie_dataset} and \ref{sec:qa_dataset}. 
\subsection{Key Information Extraction and Entity Linking}
\label{sec:kie_dataset}

\begin{table*}[h!]
\centering
\begin{adjustbox}{max width =\linewidth}
\begin{tabular}{llclccccccc}
    \toprule
    \textbf{Name} & \textbf{Venue} & \textbf{Year} & \textbf{Domain} & \textbf{\# Docs} & \textbf{\# Images} & \textbf{\# Keys} & \textbf{MP.} & \textbf{Language} & \textbf{Metrics} & \textbf{Format} \\
    
    \midrule
    FUNSD  & ICDAR-w & 2019 & Multi-source & N/A & 199 & 4 & N & English & F1 & P./ H. \\
    
    SROIE  & ICDAR-c & 2019 & Scanned Receipts & N/A & 973 & 4 & N & English & F1 & P. \\
    
    CORD & Neurips-w & 2019 & Scanned Receipts & N/A & 1,000 & 54 & N & English & F1 & P. \\
    Payment-Invoice  & ACL & 2020 & Invoice Form & N/A & 14,237+595 & 7 & N & English &F1 & D. \\
    Payment-Receipts & ACL & 2020 & Scanned Receipts & N/A & 478 & 2 & N & English & F1 & P. \\
    Kleister-NDA  & ICDAR & 2021 & Private Agreements & 540 & 3,229 & 4 & Y & English & F1 & D. \\
    Kleister-Charity  & ICDAR & 2021 & AFR & 2,778 & 61,643 & 8 & Y & English & F1 & D./ P. \\
    EPHOIE  & AAAI & 2021 & Exam Paper & N/A & 1,494 & 10 & N & Chinese & F1 & P./ H. \\
    
    
    XFUND  & ACL & 2022 & Synthetic Forms & N/A & 1,393 & 4 & N & Multilingual & F1 & D./ P./ H. \\
    Form-NLU  & SIGIR & 2023 & Financial Form & N/A & 857 & 12 & N & English & F1 & D./ P./ H. \\
    
    VRDU-Regist. Form  & KDD & 2023 & Registration Form & N/A & 1,915 & 6 & N & English & F1 & D. \\
    VRDU-Ad-buy Form  & KDD & 2023 & Political Invoice Form & N/A & 641 & 9+1(5) & N & English & F1 & D./P. \\
    DocILE  & ICDAR & 2023 & Invoice Form & 6,680 & 106,680 & 55 & Y & English & AP, CLEval & D./P. \\
    \bottomrule

\end{tabular}
\end{adjustbox}
\caption{Summary of key information extraction (KIE) datasets for VRDs. \textbf{MP.}: Multi-page; \textbf{F1}: F1-score; \textbf{P.}: Plain-text; \textbf{H.}: Handwritten; \textbf{D.}: Digital.}
\label{tab:kie_datasets}
\end{table*}

\subsubsection{Scanned Receipt Datasets}

\textbf{\textit{SROIE}} \cite{sroie} is a widely used dataset for text localization, OCR, and key information extraction from scanned receipts, introduced in the ICDAR 2019 Challenge on "\textit{Scanned Receipts OCR and Key Information Extraction}" The Key Information Extraction (KIE) task focuses on four key types: \textit{Address}, \textit{Date}, \textit{Company}, and \textit{Total}, with corresponding values provided in the annotation file for each receipt. The F1 score for this task is calculated based on Mean Average Precision (MAP) and recall. Note that entity-level annotations are not provided in the official dataset, requiring the use of external tools to obtain the information for coarse-grained or joint-grained models. 

\textbf{\textit{Payment-Receipts}} \cite{payment} is a subset of SROIE, created by sampling up to 5 documents from each template in the original SROIE dataset. The template of each receipt is decided by the Company annotation.  The target schema focuses on extracting only \textit{Date} and \textit{Total}. This subset is used to evaluate the model's ability to handle unseen templates.

\textbf{\textit{CORD}} \cite{cord} is a widely used dataset for post-OCR receipt understanding, featuring two-level labels annotated by crowdsourcing workers. It includes eight superclasses, such as \textit{Store}, \textit{Payment}, \textit{Menu}, \textit{Subtotal}, and \textit{Total}, each with several subclasses. For example, \textit{Store} contains subclasses like \textit{Name}, \textit{Address}, and \textit{Telephone}. CORD provides both textline-level and word-level annotations for both fine-grained and coarse-grained frameworks, with some sensitive information blurred. All models are evaluated on the released first 1,000 samples.

\subsubsection{Form-style Datasets}

\textit{\textbf{FUNSD}} \cite{funsd} is derived from the RVL-CDIP dataset \cite{rvlcdip} by manually selecting 199 readable and diverse template form images. The dataset is annotated using the GuiZero library to provide both entity and word-level annotations, including manual text recognition. Semantic links indicate relationships between entities, such as Question-Answer or Header-Question pairs. Consequently, FUNSD supports key information extraction, OCR, and entity linking tasks.

\textit{\textbf{XFUND}} \cite{layoutxlm} is the first multilingual dataset following the FUNSD format. It collects form templates in seven languages (Chinese, Japanese, Spanish, French, Italian, German, and Portuguese) from the internet. Human annotators fill these templates with synthetic information by typing or handwriting, ensuring each template is used only once. The filled forms are then scanned into document images, processed with OCR, and annotated with key-value pairs. Each language has 199 annotated forms, supporting multilingual key information extraction and entity linking tasks.

\textbf{\textit{Payment-Invoice }}\cite{payment} contains two corpora of invoices from different sources. The first corpus, Invoice 1, includes 14,273 invoices from various vendors with different template styles, used for training and validation. The second corpus, Invoice 2, comprises 595 documents with distinct templates not found in Invoice 1, serving as the test set. Human annotators extract six required keys from each single-page invoice, such as \textit{Invoice Date}, \textit{Total Amount}, and \textit{Tax Amount}. This dataset is suitable for evaluating generative-style models and fine-grained sequence labeling models. For coarse-grained models, additional text line or entity-level information can be extracted using off-the-shelf tools.

\textbf{\textit{VRDU-Registration Form}} \cite{vrdu} is a dataset of registration forms about foreign agents registering with the US government collected from the Federal Communications Commission. Commercial OCR tools extract the text content of the forms. Annotators draw bounding boxes around six unrepeated entities (each entity appears only once per document) per document: \textit{File Date}, \textit{Foreign Principal Name}, \textit{Registrant Name}, \textit{Registration ID}, \textit{Signer Name}, and \textit{Signer Title}. The dataset provides entity-level annotations, which can be easily preprocessed to acquire word-level annotations, supporting any granularity of Key Information Extraction (KIE) models.

\textbf{\textit{VRDU-Ad-buy Form}} \cite{vrdu} consists of 641 invoices or receipts signed between TV stations and campaign groups for political advertisements. It follows the same annotation procedure as the VRDU-Registration Form but involves a more complex schema. This includes nine unique entities (e.g., \textit{Advertiser}, \textit{Agency}, \textit{Contract ID}), four repeated entities (e.g., \textit{Item Description}, \textit{Sub Prices}), and hierarchical entities (e.g., \textit{Line Item}). Repeated entities may contain different values within a single document, while hierarchical entities comprise several repeated entities as components.

\textit{\textbf{Form-NLU}} \cite{formnlu} is a visual-linguistics dataset designed to support researchers in interpreting specific designer intentions amidst various types of noise from different form carriers, including digital, printed, and handwritten forms. Fine-grained key-value pairs, such as \textit{Company Name}, \textit{Previous Notice Date}, and \textit{Previous Shares}, are manually annotated. The training and validation set comprises 535 digital-born forms, with 76 reserved for validation. Additionally, three test sets are provided, containing 146 digital, 50 printed, and 50 handwritten form images, respectively. Form-NLU can be used to evaluate form layout analysis and Key Information Extraction (KIE) models of any granularity. With proper processing, it can also be used to evaluate entity linking frameworks, thanks to the well-annotated key-value pairs.

\subsubsection{Multi-page Datasets} 

\textbf{Kleister-NDA} \cite{nda} is a dataset collected from the Electronic Data Gathering, Analysis, and Retrieval System (EDGAR) focusing on Non-disclosure Agreements (NDAs). During preprocessing, the collected 540 HTML files are converted into digital multi-page PDF files (totalling 3,229 pages) using the Puppeteer library. Four key items, \textit{Effective Date}, \textit{Party}, \textit{Jurisdiction}, and \textit{Term}, are manually annotated by three annotators to extract the corresponding values. The NDA dataset is widely used by fine-grained level models but may require additional processing for frameworks with limited sequence length due to the multi-page inputs.

\textbf{Kleister-Charity} \cite{nda} contains 2,778 annual financial reports from the Charity Commission, which lack strict formatting rules. The Charity Commission website provides eight key pieces of information, such as Postcode, Charity Name, and Report Date. Annotators manually correct minor errors to ensure accuracy. Compared to Kleister-NDA, the Charity dataset has longer document inputs, totalling 61,643 pages, requiring models to handle long sequence outputs. Both Charity and NDA datasets provide only key-value pair annotations, making them suitable for the generation and fine-grained sequence labelling tasks but requiring additional processing to acquire entity-level annotations.

\textbf{DocILE} \cite{docile} comprises three subsets: an annotated set of 6,680 real business documents, an unlabeled set of 932,000 real business documents for unsupervised pretraining, and a synthetic set of 100,000 documents generated with full task labels. Documents come from public sources like the UCSF Industry Documents Library and Public Inspection Files, with annotations for Key Information Localization and Extraction and Line Item Recognition. Synthetic documents were created using annotated templates and a rule-based synthesizer. DocILE provides entity-level annotations that can be easily post-processed to acquire word-level annotations.

\subsubsection{\revise{Summary of KIE Datasets}}
\revise{KIE datasets for VRDs have progressively expanded to encompass a broader range of domains, formats, and complexities, as shown in Table~\ref{tab:kie_datasets}. Early benchmarks such as FUNSD \cite{funsd} and SROIE \cite{sroie} primarily consist of single-page, printed documents with a limited set of predefined keys, featuring simplified annotation schemas that facilitate model prototyping and benchmarking.
In contrast, recent datasets like Kleister-NDA, Kleister-Charity \cite{nda}, and DocILE \cite{docile} incorporate multi-page documents reflective of real-world enterprise scenarios (e.g., contracts, invoices). These introduce challenges such as long context understanding, cross-page context modeling, hierarchical key structures. Further complexity arises from the inclusion of handwritten text (e.g., FUNSD, EPHOIE) and multilingual content (e.g., XFUND), necessitating models capable of handling OCR noise, variable layouts, and linguistic diversity. These trends reflect a shift toward more realistic KIE challenges, with future directions emphasizing multimodal reasoning, cross-lingual adaptability, and robust processing of complex, noisy document structures.}

\subsection{Visually Rich Document Question Answering}
\label{sec:qa_dataset}
\begin{table}[ht]
\centering
\resizebox{\textwidth}{!}{%
\begin{tabular}{llllllllllll}
\toprule
Name & Venue & Year & Domain & \# Doc. & \# Img & \# Q & Answer Type & MP & Format & Metrics & Anno. \\
\midrule
DocVQA & WACV & 2021 & Industrial Reports & N/A & 12,767 & 50,000 & Text & N & D./P./H. & ANLS & Human \\
VisualMRC & AAAI & 2021 & Website & N/A & 10,197 & 30,562 & Text & N & D. & BLUE, etc & Human \\
TAT-DQA & MM & 2022 & Financial Reports & 2,758 & 3,067 & 16,558 & Text/RS-Gen. & Y & D. & EM, F1 & Human \\
RDVQA & MM & 2022 & Data Analysis Report & 8,362 & 8,514 & 41,378 & Text & N & D. & ANLS, ACC & Human \\
CS-DVQA & MM & 2022 & Industry Documents & N/A & 600 & 1,000 & Text and Nodes & N & D./P./H. & ANLS & Human \\
PDFVQA-Task A & PKDD & 2023 & Academic Paper & N/A & 12,337 & 81,085 & Num or Yes/No & N & D. & F1 & Template \\
PDFVQA-Task B & PKDD & 2023 & Academic Paper & N/A & 12,337 & 53,872 & Entity & N & D. & F1 & Template \\
PDFVQA-Task C & PKDD & 2023 & Academic Paper & 1,147 & 12,337 & 5,653 & Entity & Y & D. & EM & Template \\
MPDocVQA & PR & 2023 & Industrial Reports & 6,000 & 48,000 & 46,000 & Text & Y & D./P./H. & ANLS & Human \\
DUDE & ICCV & 2023 & Cross-domain & 5,019 & 28,709 & 41,541 & Text, Yes/No & Y & D. & ANLS & Human \\
MMVQA & IJCAI & 2024 & Academic Paper & 3,146 & 30,239 & 262,928 & Entity & Y & D. & EM, PM, MR & LLM + H \\
\bottomrule
\end{tabular}%
}
\caption{Summary of visually rich document question answering (VRD-QA) datasets. 
\textbf{MP}: Multi-page; \textbf{D./P./H.}: Digital / Printed / Handwritten formats; 
\textbf{RS-Gen.}: Reasoning-based Generation; 
\textbf{Metrics}: ANLS = Average Normalized Levenshtein Similarity, EM = Exact Match, F1 = F1-score, PM = Partial Match, MR = Multilabel Recall, ACC = Accuracy, BLEU = BLEU score; 
\textbf{Anno.}: Annotation source (Human, Template, or LLM + Human).}
\label{tab:vqa_datasets}
\end{table}

\subsubsection{Single Page VRD-QA Datasets}
\textit{\textbf{DocVQA}} \cite{docvqa} is a pioneering dataset in document-based Visual Question Answering (VQA), sourced from the UCSF Industry Document Library. It comprises 50,000 manually generated questions framed on 12,767 document images, encompassing digital, printed, and handwritten formats. The dataset follows an extractive-style QA format similar to benchmarks like SQuAD \cite{rajpurkar2016squad} and VQA \cite{biten2019scene}. Evaluation typically involves fine-grained models such as LayoutLM variants \cite{layoutlm, layoutlmv2, layoutlmv3}, LiLT \cite{lilt}, and generative models \cite{udop, donut}, using metrics like Average Normalized Levenshtein Similarity (ANLS) \cite{stvqa}. However, it requires additional processing for coarse-grained models like SelfDoc \cite{selfdoc}.
\textbf{\textit{CS-DVQA}} \cite{calm} builds upon DocVQA by enhancing QA pairs to better reflect real-world requirements. It extracts 600 images from the DocVQA dataset and generates 1,000 QA pairs under human supervision. During question generation, it incorporates common-sense knowledge from real life, expanding answers beyond extractive in-line text to include question-related nodes (Nodes) sourced from ConceptNet \cite{speer2017conceptnet}. 

\textbf{\textit{VisualMRC}} \cite{visualmrc} is compiled from website screenshots across 35 domains, carefully selected to exclude pages with handwritten content and to prefer pages containing short text (no more than 2 to 3 paragraphs). Unlike other datasets that might only provide question-answer annotations \cite{docvqa} or automatically acquire document semantic entities \cite{pdfvqa}, VisualMRC includes manually annotated layout structures with fine-grained semantic entity types such as \textit{Heading}, \textit{Paragraph}, \textit{Subtitle}, \textit{Picture}, and \textit{Caption}. Question-answer pairs are generated through crowdsourcing. Consequently, VisualMRC is well-suited for evaluating both fine-grained and coarse-grained-based QA frameworks, providing a rich resource for assessing the effectiveness of models in understanding and interpreting detailed document layouts and semantic entities.

\textbf{\textit{PDFVQA-Task A and Task B}} \cite{pdfvqa} form part of the first document VQA dataset from PubMed Central, focusing on content and structural understanding. This dataset includes three tasks: two for single-page documents (Tasks A and B) and one for multi-page documents (Task C). Task A evaluates the structural and spatial relationships within document images, with answers being either counts or Yes/No. Task B focuses on extracting document entities based on their logical and spatial configurations. The PDFVQA dataset provides only coarse-grained, entity-level annotations, necessitating further processing for models that require fine-grained analysis. This setup is ideal for testing models' capabilities in understanding the logical and spatial structures of document images.

\subsubsection{Multi-Page VRD-QA Datasets}

\textbf{TAT-DQA} \cite{tatdqa}, an extension of the TAT-QA \cite{zhu2021tat} dataset, is developed with more complex natural document structures and an expanded set of manually corrected and generated question-answer pairs derived from business financial reports. Unlike other datasets that primarily focus on extractive or simple abstractive answers (such as counting or yes/no), TAT-DQA includes questions requiring arithmetic reasoning, where values must be extracted from tabular data and textual content for discrete calculations. This dataset adopts the evaluation metrics of TAT-QA, including Exact Matching and a numeracy-focused F1 score. These metrics are particularly tailored to assess the accuracy of arithmetic reasoning and data extraction capabilities of the models tested with TAT-DQA. 

\textbf{\textit{RDVQA}} dataset \cite{rdvqa} compiles a large collection of conversational chats and associated images from an E-commerce platform. It employs standard OCR and Named Entity Recognition (NER) techniques to extract text and redact sensitive information, ensuring privacy protection through masking. The dataset includes question-answer pairs within the images, which are manually verified to confirm image clarity and the presence of at least one question-answer pair per image. Although some documents span multiple pages, this dataset is structured such that it can be processed relatively easily by single-page VRD-QA models.

\textbf{\textit{PDFVQA-Task C}} \cite{pdfvqa} is a distinct sub-task within the PDFVQA dataset that expands document VQA to encompass entire long documents, moving beyond the single-page focus of Tasks A and B. In Task C, to answer questions, the model often needs to retrieve information from multiple document entities. Thus, Task C employs Exact Matching for its ground truth annotations. Similar to Tasks A and B, additional processing is required for evaluating models at a fine-grained level. 

\textbf{\textit{MP-DocVQA}} \cite{mpdocvqa} extends the original DocVQA \cite{docvqa} dataset to accommodate multi-page document analysis. This version includes adjacent pages from the same documents, expanding the dataset from 12,767 to 64,057 document images. In adapting to a multi-page format, some questions inappropriate were removed. However, it's important to note that while the dataset allows for questions across multiple pages, the answers remain confined to individual pages; there are no cross-page answers in the MP-DocVQA dataset.

\textbf{\textit{DUDE}} \cite{dude} is the first cross-domain, multi-page document VQA dataset, featuring a diverse collection of documents from various fields such as medical, legal, technical, and financial, and different document types including CVs, reports, and papers. It comprises 5,019 documents, 28,709 document pages, and 41,541 manually annotated questions. Question types vary from extractive in-line text and Yes/No answers to multi-hop reasoning and structural understanding, similar to those in PDFVQA \cite{pdfvqa}. To evaluate model performance, DUDE uses the ANLS metric \cite{docvqa} for assessing answer prediction accuracy. Additionally, it employs two other metrics: Expected Calibration Error \cite{guo2017calibration} and Area-Under-Risk-Coverage-Curve (CURC) \cite{geifman2017selective,jaeger2022call} to gauge the overconfidence and miscalibration in document understanding models. These features make DUDE a comprehensive tool for evaluating cross-domain document understanding models.

\textbf{\textit{MMVQA}} \cite{mmvqa} is a dataset sourced from PubMed Central, designed for the retrieval of multimodal semantic entities from multi-page documents. The questions are generated using ChatGPT \cite{chatgpt} and subsequently verified manually. Unlike other datasets that focus solely on in-line text or text-dense entities, MMVQA also considers entire tables and figures as potential answers to the given questions. This dataset introduces various evaluation metrics to cater to different application scenarios: Exact Matching and Partial Matching Accuracy assess the precision of responses, while Multi-label Recall evaluates how well the model identifies all relevant answers across the document. This diverse set of metrics makes MMVQA suitable for comprehensive performance evaluation in complex, multimodal document understanding tasks.

\subsubsection{\revise{Summary of VRD-QA Datasets}}

\revise{Similar to KIE datasets, in Table~\ref{tab:vqa_datasets}, the development of VRD-QA datasets also demonstrates a marked progression toward practical and challenging scenarios. While earlier datasets \cite{docvqa,visualmrc} concentrated on single-page, clean digital documents with simple textual answers, newer benchmarks increasingly demand complex reasoning over multi-page layouts and multimodal semantic entities (e.g., MPDocVQA \cite{mpdocvqa}, MMVQA \cite{mmvqa}), accommodate diverse and domain-specific content (e.g., financial, academic, cross-domain), and support mixed formats including printed and handwritten inputs. The variety in answer formats—ranging from span-based text to entities, binary decisions, and reasoning-based generation—combined with rich evaluation protocols (e.g., EM, F1, ANLS, BLEU), underscores the rising complexity and real-world applicability of the task, pushing models toward deeper semantic comprehension, robust layout interpretation, and multimodal reasoning under noisy and heterogeneous conditions.}

\subsection{\revise{Evaluation Metrics}}
\textbf{\textit{F1-score}} 
The F1-score is a widely used evaluation metric in KIE, VQA, entity recognition, and document classification. It helps assess how well a model can identify relevant elements while minimizing false positives and false negatives. It provides a balanced measure of a model's precision and recall, calculated as the harmonic mean of precision and recall using the following formula:
\\F1 score = 2 × (Precision × Recall) / (Precision + Recall)

\subsubsection{\revise{Evaluation Metrics in KIE Datasets}}
\textbf{\textit{AP (Average Precision)}} 
AP measures the area under the precision–recall (P–R) curve for ranked retrieval results, calculating the average of precision values at different recall levels. In document understanding and object detection tasks, AP measures how accurately a model can identify and localize different document elements such as tables, figures, and text blocks.

\textbf{\textit{CLEval (Character-Level Evaluation)}} 
CLEval \cite{baek2020cleval} is designed for text detection and recognition tasks, providing a fine-grained assessment through: instance matching process and character scoring process. The instance matching process handles granularity differences between predicted and ground-truth text regions, and the character-level scoring process measures partial correctness. CLEval evaluation metrics provides more nuanced evaluation by considering partially correct results. 

\subsubsection{\revise{Evaluation Metrics in VRD-QA Datasets}}
\textbf{\textit{ANLS (Average Normalized Levenshtein Similarity)}} 
ANLS \cite{biten2019icdar} is a metric designed to evaluate the similarity between predicted and ground truth answers, particularly in tasks such as VQA, where outputs may contain minor errors such as typos. Formally, ANLS can be defined as: For $N$ questions, each with $M$ ground truth answers $a_{ij}$ and a prediction $o_{q_i}$

\begin{equation}
    \mathrm{ANLS} = \frac{1}{N} \sum_{i=1}^{N} \left( \max_{j} s(a_{ij}, o_{q_i}) \right)
\end{equation}

\noindent where $s(a_{ij}, o_{q_i})$ is the thresholded normalized similarity for the $j$-th ground-truth to the $i$-th prediction. ANLS normalizes Levenshtein distance to account for differences in string lengths, producing a score between 0 and 1, where 1 indicates perfect similarity. ANLS gives a high score when predicted text is very similar to the ground truth and gracefully degrades as errors increase.

\textbf{\textit{BLEU (Bilingual Evaluation Understudy)}} 
The BLEU score is a metric originally designed for evaluating machine translation quality, but has been adapted for various text generation tasks, including VQA. BLEU measures the similarity between machine-generated text and reference texts by comparing n-grams. BLEU scores range from 0 to 1, with higher scores indicating greater similarity to reference texts. 

\textbf{\textit{ACC (Accuracy)}} 
Accuracy measures the proportion of correctly predicted instances over the total number of instances. In VRD-QA tasks \cite{docvqa}, this typically refers to the percentage of questions where the predicted answer exactly matches at least one ground-truth answer. A stricter variant, \textit{Exact Match (EM)}, scores an instance as correct only if the predicted entity set is identical to the ground-truth set \cite{mmvqa}. Conversely, \textit{Partial Match (PM)} is a more lenient metric that gives credit for partial correctness—an instance is considered correct if there is any non-empty overlap between the predicted and ground-truth entity sets. While EM emphasizes exact retrieval, PM is more suitable for scenarios where partial information suffices, such as identifying at least one key entity within a document.

\textbf{\textit{MR (Multilabel Recall)}} 
MR is a metric designed for multilabel classification tasks where each instance can belong to multiple categories simultaneously. This metric evaluates the proportion of actual positive labels correctly identified by the model and is particularly important in situations where identifying all relevant positive instances is critical. MR is calculated as: MR = True Positives / (True Positives + False Negatives)

\section{\revise{Technique Overview with Performance Analysis}}
\label{sec:discussion}
Sections~\ref{sec:mono_task} and~\ref{sec:multi_task} present models designed for mono-task and multi-task VRD-CU. This section reviews core components—\textbf{Feature Representation}, \textbf{Cross-Modality Fusion}, \textbf{Model Architecture}, and \textbf{Pre-training}—with a focus on their functional roles, advantages, and limitations across different scenarios. \revise{It further discusses recent progress in integrating \textbf{LLM and MLLM into VRD-CU} pipelines, highlighting emerging trends and paradigm shifts.
Finally, to provide a comprehensive understanding of model performance, we summarize the quantitative results of representative approaches four broadly used VRD-CU benchmarks, offering a comparative view of their empirical effectiveness.}

\subsection{Feature Representation}
\subsubsection{Textual Representation}
\label{sec:textual_feature}
\begin{figure}[t]
  \centering
  \includegraphics[width=0.8\linewidth]{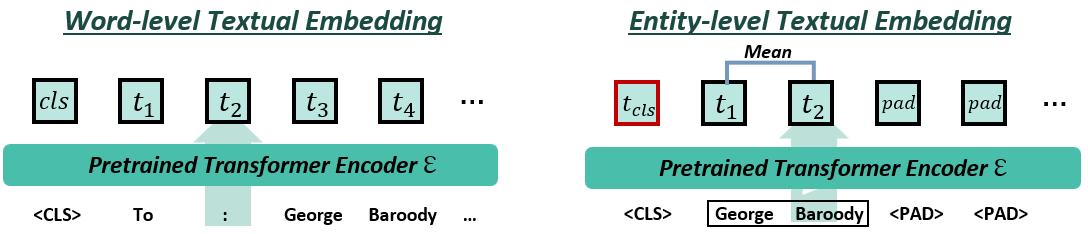}
  \caption{ Fine-grained vs. coarse-grained textual embeddings in VRD-CU. \textbf{Left}: Word-level embedding uses \( \mathcal{E}(\text{text}) = \{t_1, t_2, \dots, t_n\} \). \textbf{Right}: Entity-level embedding aggregates token vectors via mean pooling: \( e = \frac{1}{k} \sum_{i=1}^{k} t_i \), where \( \{t_1, \dots, t_k\} \) are tokens within an entity.
}

  \label{fig:text_representation}
\end{figure}

Text in VRDs provides essential semantic context, crucial for understanding content and conducting various downstream tasks. Depending on the information granularity required by the framework and application scenarios, textual representation methods can generally be categorized at the word or entity level, as illustrated by Figure~\ref{fig:text_representation}.

\textbf{\textit{Word-level Representations}} In VRDs, text sequences extracted by off-the-shelf OCR tools or PDF parsers (e.g., PDFMiner) can be encoded using word embedding methods such as Word2Vec \cite{word2vec} and Glove \cite{glove}. For more comprehensive textual embeddings, various BERT-style bi-directional pretrained transformer models like BERT \cite{bert} and RoBERTa \cite{roberta} are employed to generate context-aware word representations. As the visually rich and structurally complex nature of VRDs, layout-aware and visual-integrated fine-grained models have been developed, such as LayoutLM families \cite{layoutlm,layoutlmv2,layoutlmv3,layoutxlm}, LiLT \cite{lilt}. These models generate word representations that integrate multimodal information, combining text, visual cues, and layout structure to achieve SoTA performance on several downstream tasks.

\textbf{\textit{Entity-level Representations}} To acquire a dense representation of a text sequence for performing entity-level VRD-CU tasks, various approaches are adopted. These include averaging word embeddings specific to an entity or leveraging the [CLS] token to encapsulate the entire sequence, including averaging the word embeddings belonging to an entity or using [CLS] token to represent an entire sequence. SentenceBERT \cite{sentencebert} is also often adopted to encode text sequences within document entities. However, a standardized approach for acquiring textual representations of entities is yet to be established, necessitating preliminary testing and validation. 

\subsubsection{Visual Representation}
\label{sec:visual_feature}
\begin{figure}[t]
  \centering
  \includegraphics[width=0.8\linewidth]{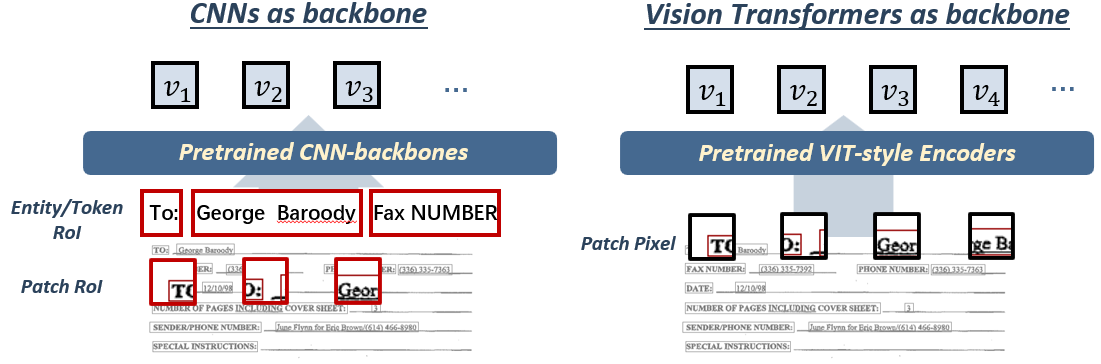}
  \caption{Comparison of visual backbone architectures for document understanding. \textbf{Left}: CNN-based backbones extract features at the entity/token or patch level using region-of-interest (RoI) pooling. \textbf{Right}: Vision Transformer (ViT)-style encoders operate directly on fixed-size image patches, enabling patch-level feature extraction across the entire document. Both approaches utilize pretrained models to obtain visual embeddings $\{ v_1, v_2, \ldots \}$ for downstream tasks.
}

  \label{fig:visual_representation}
\end{figure}
Visual information provides layout, structural insights, and rich contextual clues, making it easier for humans to interpret and prioritize content and resulting in a more comprehensive reading experience. Based on the methods used for encoding visual information, we categorize them into two main types: CNN-based and Vision Transformer-based approaches, as represented by Figure~\ref{fig:visual_representation}.

\textbf{CNN-based Vision Encoding.} Methods involve first acquiring Region of Interest (RoI) bounding boxes and then applying RoI-pooling and RoI-Align on pretrained CNN backbones (e.g., Faster-RCNN or Mask-RCNN) to extract the region features. Many frameworks \cite{layoutlm,selfdoc,docgcn,mmvqa,faststructext,pdfvqa,udoc} utilize word or entity-level RoIs to effectively extract visual features of target regions and leverage the implicit knowledge embedded in pretrained backbones. However, acquiring the bounding boxes of words or entities incurs additional costs. Therefore, several frameworks \cite{layoutlmv2,layoutxlm,wukong,docformer} directly use image patch bounding RoIs to extract visual features and learn contextually with other modalities.

\textbf{Tranformer-based Vision Encoding.} After acquiring the visual features, they are typically fed into a transformer framework to fuse multimodal information, which can create significant computational bottlenecks. Additionally, acquiring high-quality RoIs of words or entities requires supervised training. To address these challenges, \textbf{\textit{LayoutLMv3}}, inspired by ViT \cite{vit}, introduces a transformer-only framework. This approach applies a linear layer to project flattened patch pixels, which are then fed into a multimodal transformer to contextually learn with other modalities. This method reduces the number of parameters and simplifies the preprocessing steps, making it more efficient and adopted by many recent frameworks \cite{udop,docformerv2,vitlp}. However, this encoding method cannot take advantage of implicit knowledge in pre-trained frameworks and typically requires extensive pre-trained.

\subsubsection{Layout Representation}
\label{sec:positional_feature}
\begin{figure}[t]
  \centering
  \includegraphics[width=\linewidth]{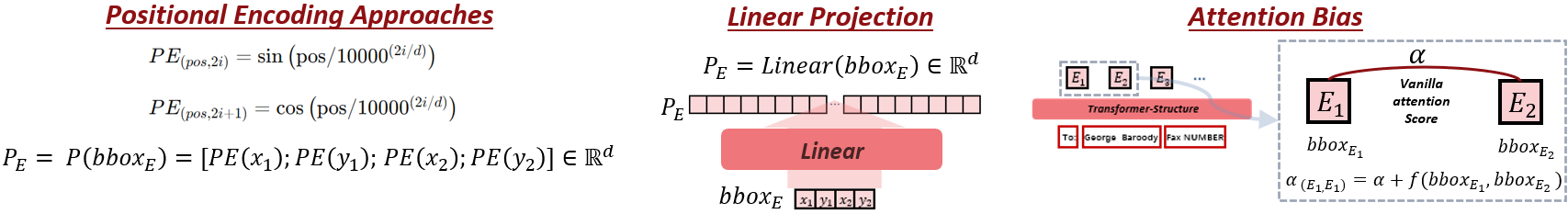}
  \caption{Overview of layout encoding methods. \textbf{Left}: Sinusoidal positional encoding applied to bounding box coordinates \( \text{PE}(bbox_E) = [\text{PE}(x_1); \text{PE}(y_1); \text{PE}(x_2); \text{PE}(y_2)] \in \mathbb{R}^d \). \textbf{Middle}: Linear projection of layout features \( P_E = \text{Linear}(bbox_E) \in \mathbb{R}^d \). \textbf{Right}: Attention bias, where layout-aware attention is modified via \( \alpha(E_1, E_2) = \alpha + f(bbox_{E_1}, bbox_{E_2}) \).
}
\label{fig:layout_representations}
\end{figure}

Layout information is crucial for understanding document elements' spatial arrangement, including words and entities. Enhance document representation by clarifying the spatial relationships between these elements, thereby aiding in the comprehension of the overall document structure. The coordinates of the bounding box (bbox) of the document elements serve as initial layout information. This layout information can then be encoded using methods such as positional encoding, linear projection, and spatial-aware attention bias.

\textbf{2D positional encoding}, first introduced by \textit{\textbf{LayoutLM}} \cite{layoutlm}, and widely adopted by many VRD-CU models, allows the model to be aware of the relative spatial positions within a document. In this approach, document elements are normalized and discretized into integer ranges, and two separate embedding layers are used to encode the x and y coordinates, respectively. Despite its widespread use in models like \cite{layoutlm,layoutlmv2,layoutxlm,layoutlmv3,layoutmask,lilt,wukong,doctr}this method encodes x and y coordinates individually, making it challenging to represent continuous 2D space and capture special correlations between document elements. 
Some models \cite{formnlu, docstruct} that follow the approach of \textbf{\textit{LXMERT}} \cite{lxmert} utilize linear projection to update the x and y coordinates of the normalized bounding box coordinates simultaneously. To address the limitations of absolute positional encoding and incorporate relative positional correlations, other models introduce spatial-aware attention mechanisms \cite{bros,docformer,lilt}. These mechanisms enable vanilla self-attention to learn spatial dependencies effectively.
\subsection{Multi-modality Fusion}
\begin{figure}[t]
  \centering
  \includegraphics[width=\linewidth]{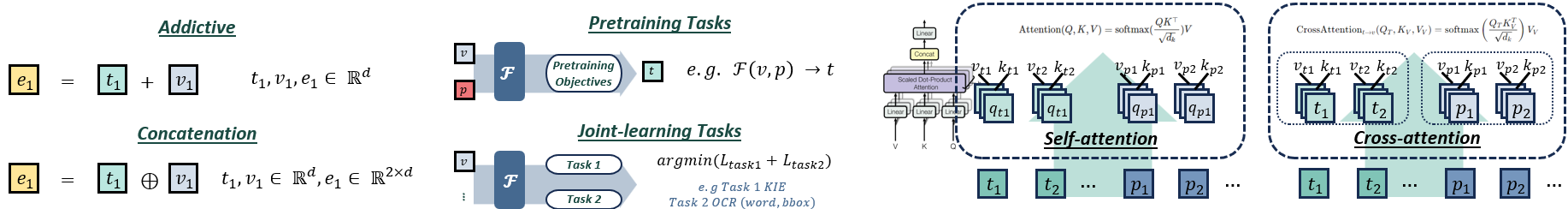}
  \caption{Overview of multimodal fusion strategies. \textbf{Left}: Additive (\( e_1 = t_1 + v_1 \)) and concatenation (\( e_1 = t_1 \oplus v_1 \)) fusion, with \( t_1, v_1 \in \mathbb{R}^d \). \textbf{Middle}: Pretraining via \( \mathcal{F}(v, p) \rightarrow t \) or joint-learning by minimizing \( \mathcal{L}_{\text{task1}} + \mathcal{L}_{\text{task2}} \). \textbf{Right}: Self-attention and cross-attention mechanisms using \( \text{Attention}(Q, K, V) = \text{softmax}\left(QK^\top/{\sqrt{d_k}}\right)V \).
}
    \vspace{-0.4cm}
  \label{fig:fusion}
\end{figure}

After acquiring multimodal representations, it is important to explore effective fusion methods to integrate textual, visual, and layout information. This integration improves document understanding and boosts performance on downstream tasks. The straightforward integration methods, as shown in Figure~\ref{fig:fusion}, include the additive and concatenation of the feature vectors. For example, additive integration sums layout information with corresponding textual or visual token representations \cite{layoutlm,layoutlmv2,layoutlmv3,selfdoc}, while concatenation merges visual and textual features of document entities (e.g. tables) \cite{mmvqa,pdfvqa}. However, these methods require one-to-one correlations and alternative approaches are needed when such correlations are not available. Consequently, self-attention and cross-attention mechanisms are widely adopted to enhance each modality by learning inter- and intramodality contexts. These mechanisms are commonly used in frameworks \cite{layoutlmv2,layoutlmv3,layoutmask,wukong,layoutxlm} that integrate patch-level visual embeddings with textual features for contextual learning. Novel self-attention \cite{docformer, lilt} and cross-attention \cite{udoc,selfdoc, faststructext} methods have been proposed to fuse multimodal information more effectively.
Apart from model-based fusion approaches, self-supervised and joint learning tasks are also effective for integrating multi-aspect features. Self-supervised pretraining tasks such as Masked Visual-Language Modeling \cite{lilt,layoutlm,layoutlmv2,structurallm,layoutlmv3,faststructext}, Text-Image Alignment \cite{docformer}, Text-Layout Pairing \cite{lilt}, and Text-Image Matching \cite{layoutlmv2,layoutxlm} can significantly enhance multimodal information fusion. These methods require large-scale pretraining to learn cross-modality semantic correlations.
Joint learning methods, often used in OCR-free frameworks \cite{ephoie,donut,dessurt,structextv2}, design auxiliary text detection or recognition tasks to fuse textual and visual information. This approach reduces pre-processing during inference and addresses mis-ordering sequence issues. However, these methods generally underperform compared to OCR-dependent models and involve additional training costs.

\subsection{Model Architecture}
\subsubsection{Transformer in VRD-CU}
Referring to the models introduced in Section~\ref{sec:mono_task} and Section~\ref{sec:multi_task}, transformers have become extensively utilized in VRD-CU tasks, attaining state-of-the-art performance due to several key advantages. Firstly, the attention mechanism effectively captures long-range dependencies within the multimodal information, including text, vision, and layout. Furthermore, the inherent scalability of transformers improves self-supervised learning on large-scale datasets (e.g. IIT-CDIP \cite{iitcdip}), allowing them to handle diverse types and formats of documents, capturing more intricate and comprehensive features of the documents. Based on the transformer architecture used, VRD-CU models can be divided into two categories: encoder-only and encoder-decoder-based models. The first encoder-only model, LayoutLM \cite{layoutlm}, was inspired by BERT \cite{bert} and uses various pretraining tasks to allow the bidirectional transformer encoder to capture more textual and layout information. Following LayoutLM, more pretrained VRD-CU models with encoder-only, layout-aware \cite{lilt,bros,xdoc,lilt,layoutmask,structurallm} or visual integrated \cite{layoutlmv2,layoutxlm,docformer,layoutlmv3} have been proposed. These models are pretrained on various tasks to enhance their understanding of document structures.
Encoder-only models demonstrate remarkable performance in sequence tagging and document classification tasks.
However, they face challenges related to heavy annotation requirements and low readability, and they struggle with generative tasks such as abstractive question answering. Additionally, OCR errors can complicate the extraction of accurate information from input text. Furthermore, the fixed maximum input length of encoder-only frameworks limits their ability to handle long document inputs effectively.
To overcome the limitations of encoder-only frameworks in generative QA and KIE, several encoder-decoder models \cite{tilt,udop,docformerv2,vitlp} have been developed, but they still depend on costly OCR tools which can introduce errors affecting performance. OCR-free frameworks \cite{donut,dessurt,serum,structextv2} address this issue by using vision encoders and text decoders for end-to-end processing. For long documents, T5-based encoder-decoder models \cite{t5} have been proposed to effectively handle multipage contexts.

\subsubsection{CNNs in VRD-CU} 
In VRD-CU, CNNs are employed as the core framework for extracting feature maps from document images \cite{yang2017learning,palm2019attend} and character grids \cite{chargrid}, benefiting from their strong local feature extraction capabilities. Joint-learning frameworks \cite{ephoie,ephoie} use CNNs as a backbone to integrate OCR and KIE tasks, combining visual and textual information through auxiliary tasks. Some OCR-free pretrained frameworks \cite{dessurt,structextv2} also utilize CNNs as vision encoders to extract visual feature maps. However, CNNs struggle to capture long-range dependencies due to their localized receptive fields. To address this, StrucTexTv2 \cite{structextv2} combines CNN-extracted feature maps with a transformer to capture global contextual information. Additionally, CNNs are commonly used to extract visual features from regions of interest (RoI) with RoI Align\cite{layoutlm,layoutlmv2,layoutxlm,layoutlmv3,layoutmask,lilt,wukong,doctr}. Although CNNs can accurately capture region-specific visual cues and leverage pre-trained knowledge from general domains, they require extra processing to obtain RoI bounding boxes, unlike vision transformers \cite{layoutlmv3,udop,docformerv2,vitlp}, which operate directly on patch pixel values. 

\subsubsection{Graphs in VRD-CU}
VRDs feature complex spatial and logical structures. The spatial structure shows the layout and positional relationships, such as a \textit{Title} above a \textit{Paragraph} and a \textit{Caption} near a \textit{Figure} or \textit{Table}. The logical structure denotes semantic and hierarchical connections, like a \textit{Title} being the parent of a \textit{Paragraph} and a \textit{Caption} describing a related \textit{Table} or \textit{Figure}. Graph-based frameworks explicitly encode these relationships using node and edge representations; therefore, GNNs are widely used in VRD-CU models \cite{formnet,layoutgcn,docgcn,zhang2021entity} to encode the spatial and logical representations. Although GNNs effectively capture domain-specific knowledge, they struggle with scalability and general domain knowledge pretraining. To address this, some frameworks \cite{graphlayoutlm,graphdoc} use attention masks or biases to mimic relationships between document elements, blending attention mechanisms with explicit relational modelling.

\subsection{Pretraining Mechanisms}
\label{sec:pretraining}
By performing various pretraining tasks, a model can enhance its generalization ability through extensive datasets and prior training. Inspired by advances in pre-training language \cite{bert,gpt3} and vision models \cite{resnet,vit}, numerous pretraining tasks for VRD-CU have been developed that are typically trained in large-scale document collections. This section will summarize the pretraining techniques and datasets commonly used for VRD-CU pretraining.
\subsubsection{Pretraining Tasks} \label{sec:pretraining_tasks}
Based on the purpose and pretraining targets, the pretraining methods can be categorised into Masked Information Modeling, Cross-modality Learning, Mono-modality Augmentation, and Contrastive Learning.

\textbf{\textit{Masked Information Modelling (MIM)}}  is first introduced by Masked Language Modelling in BERT \cite{bert}, which randomly masks 15\% workpiece tokens and requires the model to predict the masked tokens. Some models use multi-source inputs, such as XDoc \cite{xdoc} and MarkupLM \cite{markuplm}, which directly adopts \textit{Masked Language Modeling} as a pretraining task on plain text or markdown text subsets. Some methods optimise MLM by changing masking wordpieces to whole words (\textit{Whole Word Masking} by LayoutMask \cite{layoutmask}) or all tokens belong to one randomly generated text block, named \textit{Area-Masked Language Modelling} \cite{bros}. Similar strategies can also be applied to mask entity-level textual representation, e.g. \textit{Masked Sentence Modelling} \cite{udoc}. Those language-targeted masked information modelling to improve the language understanding ability of VRD-CU models. Except for language-focused masking strategies, vision \cite{udoc,structextv2} and layout-focused \cite{layoutmask,lilt} strategies are also adopted. Moreover, masked information modelling is an effective method to boost cross-modality understanding. LayoutLM \cite{layoutlm} introduces a \textit{Masked Visual-Language Modelling} which allows using kept visual/layout information and contextual text content to predict the masked word-pieces, adopted by many VRD-CU pretrained models \cite{lilt,layoutlmv2,structurallm,layoutlmv3,faststructext,docformer}. Similarly, some visual token masked models leverage multimodal information to reconstruct the view tokens, e.g. \textit{Learn to Reconstruct} \cite{docformer}, \textit{Masked Image Modeling} \cite{layoutlmv3}. Additionally, some models mask multimodal features simultaneously to conduct a cross-modality masking \cite{selfdoc,doctr}. Masking Information Modelling may have limitations on bias in masking strategies, thus some frameworks \cite{layoutmask} may try different masking ratios or strategies to improve the training effectiveness. Additionally, other common concerns about MIM also include training efficiency and lack of structural information. Thus, other pretraining methods are introduced to mitigate the limitations.  

\textbf{\textit{Cross-modality Aligning}} Although some \textit{Masked Information Modelling} methods could effectively boost the cross-modality understanding, implicit contextual learning is limited to capturing explicit alignment between different modalities. Thus, few cross-modality aligning methods are introduced to enhance the modality interaction. To enhance the text-image interactive learning, Text-Image Alignment is adopted \cite{layoutlmv2,layoutxlm} which covers the image region of token lines to predict whether the image region of the target token line is covered or not. LayoutLMv3 \cite{layoutlmv3} expands from covering the image region only to image/text tokens to further enhance interactive learning. Vision-language alignment \cite{udoc} and Text-Image Matching \cite{layoutlmv2,layoutxlm} target to predict whether image-text features belong to the same region or not. DocFormer \cite{docformer} tends to predict the text content of the paired images. 

\textbf{\textit{Other Pretraining Techniques}}: Some pretraining tasks are introduced to further enhance the understanding of specific modalities. To enhance layout information understanding, StructuralLM \cite{structurallm} and WOKONG-READER \cite{wukong} introduce \textit{Cell Position Classification} and \textit{Textline Grid Matching} to predict which located grids are of each cell or textline. Fast-StrucText \cite{faststructext} introduces a graph-based token relation method to predict a spatial correlation between token pairs. MarkupLM \cite{markuplm} leverage the benefits from markup files to predict the logical parent-child relation between nodes by introducing a \textit{Node Relation Prediction}. Contrastive learning-based strategies are adopted to conduct single \cite{udoc} or cross-modality contrastive (\textit{Textline-Region Contrastive Learning}) \cite{wukong} learning. 

\subsubsection{Pretraining Datasets} \label{sec:pretraining_datasets}
To perform the aforementioned tasks, large-scale document collections are essential for conducting self-supervised learning. Different pretraining datasets are adopted by various models.
The most widely used pretraining dataset is the IIT-CDIP Test Collection 1.0 \cite{iitcdip}, which contains more than 6 million documents with over 11 million scanned document images. Since it contains a cross-domain and large number of unannotated documents, it is used by the majority of models \cite{dit,bros,donut,layoutlm,layoutlmv2,layoutlmv3,layoutxlm,wukong,docformer,doctr,lilt,udoc,faststructext,structurallm,geolayoutlm,xdoc,docformerv2}. As the original IIT-CDIP dataset provides the text content without layout information, off-the-shelf OCR tools are normally used to acquire the bounding box information of each document.
Some models \cite{structext,selfdoc,mgdoc} use relatively smaller pretraining datasets like RVL-CDIP \cite{rvlcdip}, which contains 400,000 evenly distributed documents in 16 types, to reduce the cost of pretraining. The multi-source model XDoc \cite{xdoc} also leverages many plain text corpora for pretraining, such as BookCORPUS, CC-NEWS, OPENWEBTEXT, STORIES and HTML-sourced CommonCrawl datasets.
To address multilingual scenarios, both LiLT \cite{lilt} and LayoutXLM \cite{layoutxlm} follow the principles and policies of Common Crawl to gather large amounts of multilingual digitally-born PDF documents.

\subsection{\revise{LLM and MLLM into VRD-CU}}
\label{sec:llms}
\revise{In addition to frameworks specifically tailored for VRD-CU tasks, a growing number of general-purpose MLLMs have been proposed. These models aim to handle a wide range of modalities and domains—including plain text, natural scenes, documents, charts, and tables—and have demonstrated strong performance on various VRD-CU benchmarks. Many of these models incorporate specialized architectural mechanisms, pretraining or instruction tuning  to support broad vision-language understanding across tasks. 
These general-purpose models can be broadly classified into two categories based on their reliance on text extraction: OCR-dependent and OCR-free LLMs/MLLMs.}

\revise{\subsubsection{OCR-Dependent LLMs/MLLMs}}
\revise{OCR-dependent LLMs/MLLMs rely on off-the-shelf OCR tools to extract text and layout information, typically in the form of tokens and bounding boxes, from scanned or digital documents. These extracted elements are either directly embedded into LLM prompts (e.g., ICL-D3IE \cite{icld3ie}, DocLLM \cite{docllm}) or fused with visual features using multimodal encoders like ViT \cite{vit} or LayoutLMv3 \cite{layoutlmv3} (e.g., LayoutLLM \cite{layoutllm}, DoCo \cite{li2024enhancing}, GPE \cite{gpe}). To incorporate layout cues, models commonly use 2D positional encodings, bounding box-aware prompts, or auxiliary encoders, with some frameworks verbalizing spatial coordinates or using layout-specific tokens to enrich input representations. While this modular design reduces the need for costly end-to-end pretraining on text recognition, it introduces vulnerabilities: OCR errors, especially in noisy or handwritten documents, can cascade through the pipeline, and unordered extracted text often requires additional processing like XY-cut reordering. Furthermore, layout representations are often implicit or rule-based \cite{lapdoc,icld3ie}, which may fail to capture fine-grained structural dependencies. The use of low-resolution images to reduce computation can also limit the expressiveness of visual features, leading to degraded multimodal fusion and overall performance, especially on layout-sensitive tasks.}

\revise{\subsubsection{OCR-Free LLMs/MLLMs}} \revise{OCR-free LLMs/MLLMs have emerged as a promising direction when applying in end-to-end VRD-CU, eliminating the dependency on explicit text extraction by operating directly on document images. These models typically employ one or more vision encoders to extract visual features, which are then fused with user queries and decoded by LLMs to generate task-specific outputs such as, Donut, mPLUG-DocOwl, and UReader~\cite{donut,mplugdocowl,ureader}.
Achieving fine-grained textual understanding in OCR-free settings requires high-resolution input, which results in long visual sequences and necessitates the use of visual compression modules to maintain computational tractability~\cite{hrvda,mplugdocowl15}. Additionally, effective integration of textual semantics and document layout often hinges on large-scale pretraining. This is typically achieved through auxiliary tasks such as text recognition~\cite{textmonkey} and image captioning~\cite{feng2024docpedia}, which facilitate multimodal alignment. However, the paradigm is heavily reliant on extensive dataset construction and significant computational resources, presenting considerable challenges for scalability and real-world deployment.}


\subsection{\revise{Performance Analysis}}

\begin{table}[t]
\centering
\begin{adjustbox}{width=\textwidth, totalheight=\textheight, keepaspectratio}
\begin{tabular}{llllllll|ccccccc|cccc}
\toprule
\multicolumn{7}{c}{\textbf{Model Attributes}} & \multicolumn{8}{c}{\textbf{Architectural Details}} & \multicolumn{4}{c}{\textbf{Performance Scores}} \\
\cmidrule(lr){1-8} \cmidrule(lr){9-15} \cmidrule(lr){16-19}
\textbf{Model} & \textbf{Venues} & \textbf{Year} & \textbf{Var} & \textbf{Size} & \textbf{PT} & \textbf{Tasks} & \textbf{PT Size} & \textbf{Modality} & \textbf{OCR} & \textbf{Frmwk.} & \textbf{Grained} & \textbf{V.} & \textbf{L.} & \textbf{Fusion} & \textbf{FUNSD} & \textbf{SROIE} & \textbf{CROD} & \textbf{DocVQA} \\
\midrule
\rowcolor{gray!20} \multicolumn{19}{l}{\textbf{Text-only Baselines}} \\
\midrule
\multirow{2}{*}{BERT} & \multirow{2}{*}{NAACL} & \multirow{2}{*}{2018}& base & 110M & \multirow{2}{*}{Y} & QA, KIE, & \multirow{2}{*}{3.3B*} & \multirow{2}{*}{T} & \multirow{2}{*}{Y} & \multirow{2}{*}{T-E} & \multirow{2}{*}{Fine} &\multirow{2}{*}{N/A} & \multirow{2}{*}{N/A} & \multirow{2}{*}{N/A} & 60.26 & 90.99 & 89.68 & 63.72 \\
& & & large & 340M & & LA, DC & &  &  &  &  &  &  &  & 65.63 & 92 & 90.25 & 67.45 \\
\midrule
\multirow{2}{*}{RoBERTa} & \multirow{2}{*}{Preprint} & \multirow{2}{*}{2018} & base & 125M &  \multirow{2}{*}{Y}&  QA, KIE, &  \multirow{2}{*}{30B*} &  \multirow{2}{*}{T} &  \multirow{2}{*}{Y} &  \multirow{2}{*}{T-E} &  \multirow{2}{*}{Fine} &  \multirow{2}{*}{N/A} &  \multirow{2}{*}{N/A} &  \multirow{2}{*}{N/A} & 66.48 & / & 93.54 & 66.42 \\
& & & large & 355M & & LA, DC &  & & & & & & & & 70.72 & / & 93.86 & 69.52 \\
\midrule

\rowcolor{gray!20} \multicolumn{19}{l}{\textbf{Mono-Task (KIE) Frameworks}} \\
\midrule
PICK & ICPR & 2020 & -  & -& N & KIE & N/A & T+L+V & Y & Graph & Fine & CNN & Linear & Concat & / & 96.12 & / & / \\
\midrule
TRIE & MM & 2020 & -  & -& N & KIE & N/A & T+L+V & Y & RNN & Coarse & CNN & PE & Self-Att. & / & 96.18 & / & / \\
\midrule
VIES & AAAI & 2021 & -  & -& N & KIE & N/A & T+L+V & Y & Self-atten. & Coarse & CNN & Linear & Sum & / & 96.12 & / & / \\
\midrule
SPADE & ACL & 2021 & -  & 110M & N& KIE & N/A & T+L+Y & Y & Graph & Fine & N/A & PE & Sum & 70.59 & 91.5 & / & / \\
\midrule
XYLayoutLM & CVPR & 2022 & -& - & B & KIE & N/A & T+L+V & Y & T-E & Fine & CNN & PE & Rule* & 83.35 & / & / & / \\
\midrule
FormNet & ACL & 2022 & - & 217M & Y& KIE & 0.7M  & T+L+Y & Y & T-E & Fine & N/A & Graph & Graph & 84.69 & / & 97.28 & / \\
\midrule
FormNetv2 & ACL & 2023 & - & 204M & Y& KIE & 11M  & T+L+V & Y & T-E & Fine & CNN & Graph & Graph & 86.35 & 98.31 & 97.37 & / \\
\midrule
DocTR & CVPR & 2023 & - & 153M & Y& KIE & 11M  & T+L+V & Y & T-E/T-D & Fine & CNN & PE & Cross-Att. & 84 & 98.2 & / & / \\
\midrule

\rowcolor{gray!20} \multicolumn{19}{l}{\textbf{Layout-aware Pretrained Frameworks}} \\
\midrule
\multirow{2}{*}{LayoutLM} & \multirow{2}{*}{KDD} & \multirow{2}{*}{2020}& base & 160M & \multirow{2}{*}{Y} & QA, KIE,& \multirow{2}{*}{11M}  & \multirow{2}{*}{T+L} & \multirow{2}{*}{Y} & \multirow{2}{*}{T-E} & \multirow{2}{*}{Fine} & \multirow{2}{*}{N/A} & \multirow{2}{*}{PE} & \multirow{2}{*}{Concat} & 78.66 & 94.67 & 94.72 & 69.79 \\
& & & large & 343M & & LA, DC &   &  &  &  &  &  &  &  & 78.95 & 95.24 & 94.93 & 72.59 \\
\midrule
StructuralLM & ACL & 2021 & - & 355M & Y & KIE, QA, DC & 11M  & T+L & Y & T-E & Coarse & N/A & PE & Self-Att. & 85.14 & / & / & 83.94 \\
\midrule
XDocE & EMNLP & 2022 & - & 146M & Y & KIE, LA, DC & 42M+ *  & T+L & Y & T-E & Fine & N/A & PE & Sum & 89.4 & / & 92.05 & 72.7 \\
\midrule
\multirow{2}{*}{BROS} & \multirow{2}{*}{AAAI} & \multirow{2}{*}{2022} & base & 110M & \multirow{2}{*}{Y} & IE, RP & \multirow{2}{*}{10.6M} & \multirow{2}{*}{T+L} & \multirow{2}{*}{Y} & \multirow{2}{*}{T-E} & \multirow{2}{*}{Fine} & \multirow{2}{*}{N/A} & \multirow{2}{*}{PE} & \multirow{2}{*}{Self-Att.} & 83.05 & 96.28 & 95.73 & / \\
& & & large & 340M & & IE, RP &  &  &  &  &  & &  &  & 84.52 & 96.62 & 96.07 & / \\
\midrule
LiLT & ACL & 2022 & - & - & Y & KIE, LA, DC & 11M  & T+L & Y & T-E & Fine & N/A & PE & Cross-Att. & 88.41 & 96.4 & 96.07 & / \\
\midrule
\multirow{2}{*}{LayoutMask} & \multirow{2}{*}{ACL} & \multirow{2}{*}{2023} & base & 182M & \multirow{2}{*}{Y} & \multirow{2}{*}{KIE, DC} & \multirow{2}{*}{10M} & \multirow{2}{*}{T+L} & \multirow{2}{*}{Y} & \multirow{2}{*}{T-E} & \multirow{2}{*}{Fine} & \multirow{2}{*}{N/A} & \multirow{2}{*}{PE} & \multirow{2}{*}{Self-Att.} & 92.91 & 96.87 & 96.99 & / \\
& & & large & 404M & & &  &  &  &  &  &  &  &  & 93.2 & 97.27 & 97.19 & / \\
\midrule

\rowcolor{gray!20} \multicolumn{19}{l}{\textbf{Vision-integrated Pretrained Frameworks}} \\
\midrule
\rowcolor{gray!7}  \multicolumn{19}{c}{Fine-grained Frameworks} \\
\midrule
\multirow{2}{*}{LayoutLMv2} & \multirow{2}{*}{ACL} & \multirow{2}{*}{2021} & base & 200M & \multirow{2}{*}{Y}& QA, KIE,& \multirow{2}{*}{50M}  & \multirow{2}{*}{T+L+V} & \multirow{2}{*}{Y} & \multirow{2}{*}{T-E} & \multirow{2}{*}{Fine} & \multirow{2}{*}{CNN} & \multirow{2}{*}{PE} & \multirow{2}{*}{Concat} & 82.76 & 96.25 & 94.95 & 78.08 \\
& & &large & 426M & & LA, DC &   &  &  &  &  &  &  &  & 84.2 & 97.81 & 96.01 & 83.48 \\
\midrule
\multirow{2}{*}{DocFormer} & \multirow{2}{*}{ICCV} & \multirow{2}{*}{2021} & base & 183M& \multirow{2}{*}{Y}  & \multirow{2}{*}{KIE, LA, DC} & \multirow{2}{*}{5M} & \multirow{2}{*}{T+L+V} & \multirow{2}{*}{Y} & \multirow{2}{*}{T-E} & \multirow{2}{*}{Fine} & \multirow{2}{*}{CNN} & \multirow{2}{*}{PE} & \multirow{2}{*}{Self-Att.} & 83.34 & 96.88 & 96.33 & / \\
& & & large & 536M & & &  &  &  &  &  &  & &  & 84.55 & 96.28 & 96.99 & / \\
\midrule
\multirow{2}{*}{LayoutLMv3} & \multirow{2}{*}{MM} & \multirow{2}{*}{2022}& base & 133M & \multirow{2}{*}{Y} & QA, KIE,& \multirow{2}{*}{50M}  & \multirow{2}{*}{T+L+V} & \multirow{2}{*}{Y} & \multirow{2}{*}{T-E} & \multirow{2}{*}{Fine} & \multirow{2}{*}{ViT} & \multirow{2}{*}{PE} & \multirow{2}{*}{Self-Att.} & 90.29 & 96.12 & 96.56 & 78.76 \\
& & & large & 368M & & LA, DC &   & &  &  &  &  &  &  & 92.08 & 96.18 & 97.46 & 83.37 \\
\midrule

\rowcolor{gray!7} \multicolumn{19}{c}{Coarse-grained Frameworks} \\
\midrule
SelfDoc & CVPR & 2021 & - & -& Y & KIE, LA, DC & 320K & T+L+V & Y & T-E & Coarse & CNN & PE & Cross-Att. & 83.36 & / & / & / \\
\midrule
UDoc & NeurIPS & 2021 & - & 272M & Y & KIE, LA, DC & 1M & T+L+V & Y & T-E & Coarse & CNN & Linear & Cross-Att. & 87.93 & / & 98.94 & / \\
\midrule
MGDoc & EMNLP & 2022 & - & 203M & Y & KIE, DC & 11M  & T+L+V & Y & T-E & Coarse & CNN & PE & Concat & 89.44 & / & 97.11 & / \\
\midrule

\rowcolor{gray!7} \multicolumn{19}{c}{Joint-grained Frameworks} \\
\midrule
StructText & MM & 2021 & - & 107M & Y & KIE, RP & 0.9M & T+L+V & Y & T-E & Joint & CNN & Linear & Self-Att. & 83.09 & 96.88 & / & / \\
\midrule
\multirow{2}{*}{mmLayout} & \multirow{2}{*}{MM} & \multirow{2}{*}{2022} & base & 236M & \multirow{2}{*}{Y} & \multirow{2}{*}{KIE, LA} & \multirow{2}{*}{N/A} & \multirow{2}{*}{T+L+V} & \multirow{2}{*}{Y} & \multirow{2}{*}{Graph} & \multirow{2}{*}{Joint} & \multirow{2}{*}{CNN} & \multirow{2}{*}{PE} & \multirow{2}{*}{Concat} & 86.02 & 97.63 & 97.23 & 79.15 \\
& & & large & -  &  &  &  &  &  &  &  &  &  &  & 86.49 & 97.91 & 97.38 & 83.66 \\
\midrule
Fast-StrucTexT & IJCAI & 2023 & - & 116M & Y  & KIE, RP & 11M  & T+L+V & Y & T-E & Joint & CNN & PE & Cross-Att. & 90.35 & 97.55 & 97.15 & / \\
\midrule
\multirow{2}{*}{Wukong} & \multirow{2}{*}{ACL}& \multirow{2}{*}{2023}  & base & 237M& \multirow{2}{*}{Y} & \multirow{2}{*}{KIE, LA, DC} & \multirow{2}{*}{11M} & \multirow{2}{*}{T+L+V} & \multirow{2}{*}{Y} & \multirow{2}{*}{T-E} & \multirow{2}{*}{Joint} & \multirow{2}{*}{ViT} & \multirow{2}{*}{PE} & \multirow{2}{*}{Self-Att.} & 91.52 & 96.88 & 96.54 & / \\
& & & large & 470M&  &  &  &  &  &  &  &  &  &  & 93.62 & 98.15 & 97.27 & / \\
\midrule
GeoLayoutLM & CVPR & 2023 & - & 399M & Y & KIE, RP & 11M  & T+L+V & Y & T-E & Joint & ViT & PE & Cross-Att. & 92.86 & / & 97.97 & / \\
\midrule

\rowcolor{gray!20} \multicolumn{19}{l}{\textbf{Encoder-Decoder based Frameworks}} \\
\midrule
\multirow{2}{*}{TILT} & \multirow{2}{*}{ICDAR}  & \multirow{2}{*}{2021}& base & 230M & \multirow{2}{*}{Y}& \multirow{2}{*}{QA, KIE} & \multirow{2}{*}{1.1M} & \multirow{2}{*}{T+L+V} & \multirow{2}{*}{Y} & \multirow{2}{*}{T-E/T-D} & \multirow{2}{*}{Fine} & \multirow{2}{*}{CNN} & \multirow{2}{*}{PE} & \multirow{2}{*}{Concat} & / & 97.65 & 95.11 & 83.92 \\
& & & large & 780M & &  &  &  &  &  &  &  &  &  & / & 98.1 & 96.33 & 87.05 \\
\midrule
Dessurt & ECCV & 2022 & - & - & Y & KIE, QA, DC & N/A & T+V & N & T-E/T-D & Fine & CNN & N/A & Cross-Att. & 65 & / & / & 63.2 \\ \midrule
Donut & ECCV & 2022 & - & 176M & Y & KIE, QA, DC & 13M & T+V & N & T-E/T-D & Fine & Swin & N/A & Cross-Att. & / & / & 84.1 & 72.1 \\
\midrule
SeRum & ICCV & 2023 & - & 136M & Y & KIE, QA & 13M+ & T+V & N & T-E/T-D & Fine & Swin & N/A & Cross-Att. & / & 85.8 & 84.9 & 71.9 \\
\midrule
UDOP & CVPR & 2023 & - & 742M & Y  & KIE, QA, DC & 11M  & T+L+V & Y & T-E/T-D & Fine & ViT & PE & Concat & 91.62 & / & 97.58 & 84.7 \\
\midrule
ViTLP & NAACL & 2024 & - & 253M & Y & KIE, QA, DC & 13M & T+L+V & Y & T-E/T-D & Fine & ViT & PE & Cross-Att. & 87.61 & / & 95.59 & 65.9 \\
\midrule

\rowcolor{gray!20} \multicolumn{19}{l}{\textbf{LLM-based Frameworks}} \\
\midrule
LapDoc & ICDAR & 2024 & - & - & N & KIE, QA & N/A & T+L & Y & T-D & Fine & N/A & Prompt & Prompt & / & 79.9 & / & 79.8 \\ \midrule
DocLLM & ACL & 2024 & - & 7B & Y & KIE, QA, DC & 3.8M & T+L & Y & T-D & Fine & N/A & PE & Cross-Att. & 51.8 & 91.9 & 67.4 & 69.5 \\ \midrule
LayoutLLM & CVPR & 2024 & - & 7B & Y & KIE, QA & 6M+ & T+L+V & Y & T-E/T-D & Fine & Fine & PE & Self-Att. & 79.98 & 72.12 & 63.1 & 74.27 \\ \midrule
HRVDA & CVPR & 2024 & - & 7.2B+ & Y & KIE, QA & 13M+ & T+L+V & Y & VT-E/T-D & Fine & Swin & PE & LoRA & / & 91 & 89.3 & 72.1 \\ \midrule

DocLayLLM & CVPR & 2025 & - & 8B & Y & KIE, QA & 3.1M & T+L+V & Y & VT-E/T-D & Fine & ViT & Prompt & Self-Att. & 84.11 & 84.36 & 71.34 & 78.36 \\ \midrule
LayTokenLLM & CVPR&  2025 & - & 8B & Y & KIE, QA & 5.3M & T+L & Y & T-D & Fine & N/A & PE & LoRA & 81.62 &  / & 78.3 & 85.11 \\ \midrule
GPE & ICLR & 2025 & - & 7B & Y & KIE, QA & 0.5M & T+L & Y & T-D & Fine & N/A & PE & Sum & 82.6 & 97.8 & 86.9 & 78.1 \\ 

\bottomrule
\end{tabular}
\end{adjustbox}
\small \caption{\revise{Performance and Attribute Comparison of Leading VRD-CU Models.
This table provides a comprehensive overview of various VRD-CU models, detailing their published venue, architectural Variant (Var), and pretraining (PT) size. It outlines their supported Tasks, including Key Information Extraction (KIE), Question Answering (QA), Layout Analysis (LA), Document Classification (DC), and Relation Prediction (RP). The table also specifies their input Modality (Text (T), Layout (L), Visual (V)), Optical Character Recognition (OCR) tool requirement, and Framework (Frmwk.) types (e.g., Transformer Encoder (T-E), Self-attention (Self-atten.), Transformer Decoder (T-D)). Further architectural details include granularity levels (Grained), visual encoder (V.), layout representation (L.) (e.g., Linear projected (Linear), Position Encoding (PE)), and fusion mechanisms (e.g., Concatenation (Concat), Self-attention (Self-Att.), Cross-Attention (Cross-Att.)). Performance scores are reported on the FUNSD, SROIE, CORD, and DocVQA datasets, using F1-score for FUNSD, SROIE, and CORD, and ANLS for DocVQA. }}

\label{tab:document_understanding_models}
\end{table}

\revise{Table~\ref{tab:document_understanding_models} provides a structured comparison of key VRD-CU models, enabling us to identify performance trends, limitations, and trade-offs across different architectural designs, task categories, and dataset domains.}

\revise{\textbf{Text-only vs. Multimodal Models.} Early baselines such as BERT~\cite{bert} and RoBERTa~\cite{roberta}, while foundational for natural language understanding, perform significantly worse on visually-rich document tasks due to their inability to model spatial and visual structures. On tasks like KIE (e.g., FUNSD) and document QA (e.g., DocVQA), these text-only models underperform by 15–30 percentage points compared to multimodal alternatives. This performance gap underscores the importance of incorporating layout (L) and visual (V) modalities alongside text (T). LayoutLMv3~\cite{layoutlmv3} and GeoLayoutLM~\cite{geolayoutlm} exemplify this shift, offering gains of over 10 points on datasets such as SROIE and XFUND through layout-aware and multilingual pretraining.}

\revise{\textbf{Task-specific vs. Multi-task Models.} Mono-task models like SPADE~\cite{hwang2021spatial} and FormNetv2~\cite{formnetv2} are tailored for KIE, often achieving high performance on structured forms (e.g., 98.2\% on SROIE). However, these models typically show reduced performance on QA or classification tasks, indicating limited generalisation capacity. In contrast, multi-task frameworks such as Wukong~\cite{wukong} and UDoc~\cite{udoc} demonstrate broader versatility, achieving consistently high scores across both KIE and QA tasks (e.g., UDoc reports 98.94\% on CROD and 85.3\% on DocVQA). This suggests that task-agnostic pretraining with cross-modal alignment confers better robustness and generalisation, particularly when applied to real-world, multi-format documents.}

\revise{\textbf{OCR-based vs. OCR-free Models.} OCR-dependent models remain dominant in terms of accuracy on text-intensive tasks. For instance, LayoutLMv3 achieves over 80\% on DocVQA and near-perfect accuracy on SROIE due to reliable text extraction. Meanwhile, OCR-free models like Donut~\cite{donut} and Dessurt~\cite{dessurt} show promising results in terms of end-to-end pipeline simplification and robustness to OCR noise, but they generally fall short by 5–15 percentage points on benchmarks such as FUNSD and XFUND. This performance drop is particularly notable for multilingual or handwritten datasets where precise text recognition remains a bottleneck. Therefore, OCR-free models offer an elegant alternative but require further advances in visual text understanding to close the performance gap.}

\revise{\textbf{Model Capacity and Efficiency Trade-offs.} Models like Wukong-large and LayoutMask-large demonstrate that increasing model capacity enhances performance, especially on high-complexity tasks, with LayoutMask-large reaching 93.2\% on FUNSD. However, such improvements come at the cost of computational demands, memory usage, and slower inference, which may hinder practical deployment in resource-constrained settings. Lightweight models or those leveraging parameter-efficient tuning (e.g., adapters, LoRA) are emerging as viable solutions, though they currently underperform on complex multi-modal reasoning tasks.}

\revise{In summary, performance trends across VRD-CU models reveal that (1) multimodal integration consistently outperforms text-only approaches, (2) multi-task learning improves cross-domain generalisation, (3) OCR-free methods offer architectural simplification at the cost of task accuracy, and (4) higher model capacity often improves performance but exacerbates scalability and deployment challenges. These observations not only provide critical insights into the trade-offs inherent in current designs but also guide future directions in balancing performance, generalisability, and efficiency.}

\section{\revise{Critical Discussion about Future Trend and Practical Challenges}}
\label{sec:challenges}

\subsection{\revise{Challenges in Multimodal Information Representation}}
\revise{The inherently multimodal nature of VRD-CU tasks is well established \cite{vrdiu}, with recent frameworks increasingly integrating multimodal information through varied strategies. In the text modality, a primary challenge lies in accurately extracting content from scanned and handwritten documents. Both OCR-dependent \cite{layoutlmv3,docformerv2} and OCR-free methods \cite{donut,dessurt} continue to struggle with this issue, which remains contingent on advances in text spotting and recognition. Another key limitation is the application of these frameworks to low-resource languages, where the scarcity of annotated datasets \cite{layoutxlm} and the reduced performance of LLMs significantly undermine effectiveness. On the visual side, most pretrained VRD-CU models rely on low-resolution images to limit the number of visual tokens, often at the cost of fine-grained details—particularly problematic for OCR-free systems. Although high-resolution-compatible methods (e.g., adaptive cropping \cite{ureader}, visual compressors \cite{mplugdocowl15}, and resampler \cite{Qwen2-VL}) have emerged, they face persistent trade-offs between token efficiency and content preservation, especially when handling lengthy documents. Finally, layout encoding has evolved from simple 2D positional encodings to RoPE-based methods \cite{gpe} compatible with LLMs. Recent approaches increasingly embed layout features during pretraining and instruction tuning rather than relying solely on prompts or external inputs. Nonetheless, scaling layout encoding to long documents remains difficult, particularly in preserving spatial coherence across pages.}

\subsection{\revise{Challenges in Multimodal Information Fusion}}
\revise{Beyond challenges in feature representation, the effective fusion of encoded multimodal features remains an open research problem. Direct fusion methods, such as simple summation or concatenation, lack the capacity to capture complex inter-modal dependencies and often underperform in scenarios requiring fine-grained alignment. Additionally, as model sizes grow and input sequences expand—especially due to the incorporation of fine-grained visual and layout features—self-attention and cross-attention mechanisms become computationally expensive, posing scalability challenges for large-scale pretraining and instruction tuning in LLM-based frameworks. To mitigate this, lightweight modules such as adapters and LoRA have been adopted to reduce training costs \cite{feng2024docpedia,texthawk,texthawk2}. However, LoRA-style methods often exhibit limited flexibility and generalization, requiring task-specific supervised fine-tuning and performing inconsistently across diverse document types. Additionally, while several well-defined pretraining and instruction-tuning tasks have been proposed to enhance multimodal fusion especially for OCR-free frameworks \cite{vary,marten,dockylin}, even LLM-based models trained on large-scale datasets for text spotting and recognition still underperform compared to OCR-dependent or small-scale pretrained VRD-CU models.}

\subsection{\revise{Challenges with Multi-page and Long Documents}}
\revise{In real-world scenarios, VRDs often span multiple pages—such as academic papers, financial statements, or industrial reports—yet most existing frameworks are designed for single-page inputs and are constrained by token limits (e.g., 200 visual and 512 textual tokens in LayoutLM-series \cite{layoutlm,layoutlmv2,layoutlmv3,layoutxlm}), making it difficult to process text-dense page or multi-page documents. Existing multi-page approaches (e.g.,\cite{mpdocvqa,gram}) typically rely on retrieval mechanisms to select relevant pages, which are then processed independently by MLLM-based VRD-CU systems; however, such pipelines often fail to capture semantic and logical dependencies across pages, leading to incomplete contextual understanding. While Hu et al.~\cite{mplugdocowl2} introduce multi-page scale pretraining to enhance cross-page comprehension, their focus remains primarily on text recognition, with limited modeling of semantic and logical relationships. Additionally, managing long input sequences—especially with high-resolution images—poses challenges for effective multimodal retrieval, compression, and fusion, as maintaining spatial and logical coherence across long text-visual sequences remains difficult. Moreover, existing multi-page benchmarks \cite{mmvqa,dude,mpdocvqa} are mostly limited to extractive tasks with low inter-page complexity, leaving critical challenges like multi-hop and multimodal reasoning under-explored.}

\subsection{\revise{Challenges in Data Efficiency: Synthetic, Few/Zero-shot Learning}}
\revise{In real-world scenarios, acquiring high-quality, manually curated datasets for new document collections is time-consuming and labor-intensive, often requiring domain experts to annotate and ensure data quality \cite{gandhi2024better}. While a few studies \cite{chen2023task,laser} have proposed few-shot learning frameworks, they remain limited—particularly in KIE tasks—with constrained sets of pre-defined keys and generally underperform compared to fully fine-tuned models. To address data scarcity and improve few-shot and zero-shot performance, recent LLM/MLLM-based approaches \cite{llama,mplugdocowl15} have turned to synthetically generating large-scale instruction-response pairs for instruction tuning. This strategy helps models better understand user queries in low-resource settings, yet noticeable performance gaps persist, as shown in Table~\ref{tab:document_understanding_models}. A recent study \cite{david} further investigates tuning models with synthetic datasets to bridge domain gaps between general-domain pretraining and downstream document tasks. However, synthetic data often lacks the quality, complexity, and diversity of manually curated datasets \cite{gandhi2024better}, limiting its effectiveness for robust few-shot and zero-shot generalization. As such, future work should explore active learning, human-in-the-loop pipelines and reinforcement learning to improve the authenticity, task alignment, and transferability of synthetic training data \cite{li2024survey}.}

\begin{table}[ht]
\centering
\resizebox{\textwidth}{!}{%
\begin{tabular}{p{2.8cm}|p{4.8cm}|p{3.5cm}|p{5.2cm}}
\toprule
\textbf{Aspect} & \textbf{Description} & \textbf{Affected Domains} & \textbf{Remarks} \\
\midrule
OCR Noise & Noisy scans, handwriting, and low-resolution inputs affect text quality & Legal, Healthcare & OCR-free models still struggle with poor-quality or multilingual handwritten inputs \\
\hline
Multilinguality & Mixed scripts and cross-lingual layout inconsistencies & Government, Finance, Education & Limited multilingual pretraining; lack of annotated cross-lingual datasets \\
\hline
Latency and Efficiency & Long documents and high-resolution inputs increase inference cost & RPA, Mobile Apps, Enterprise Systems & Transformer-based models have high memory/latency demands; few studies address real-time use \\
\hline
Privacy and Data Sensitivity & Sensitive information restricts data use and sharing & Healthcare, Finance & Requires privacy-preserving learning (e.g., federated learning) and explainability \\
\hline
Domain Adaptation & Models fail to generalise across domains with unique templates and semantics & Legal, Insurance, Logistics & Domain-adaptive pretraining and few-shot adaptation remain under-developed \\
\hline
Explainability & Lack of interpretable outputs for extracted fields or decisions & Legal, Healthcare, Finance & Needed for compliance, auditability, and user trust \\

\bottomrule
\end{tabular}%
}
\caption{\revise{Summary of Domain-Specific Challenges in VRD-CU Applications}}
\label{tab:domain_challenges}
\end{table}

\subsection{\revise{Challenges in Domain-specific Applications}}
\revise{While most benchmark-driven VRD-CU research has focused on general-purpose or synthetic datasets, real-world deployment in domain-specific settings introduces a host of practical challenges that remain under-explored. In fields such as law, finance, healthcare, and robotic process automation (RPA), documents often exhibit unique structural conventions, sensitive content, and operational constraints that demand specialised adaptation. A summary of these domain-specific challenges and their implications is presented in Table~\ref{tab:domain_challenges}.}

\revise{One critical issue is the presence of OCR noise in low-quality scans, handwritten forms, or degraded faxes, which are particularly common in archival legal documents or hospital records. Although recent OCR-free models demonstrate robustness under certain conditions, they still struggle with noisy inputs, particularly in multilingual or script-heavy environments. For example, forms from multilingual regions, such as the EU or India, often contain mixed scripts (e.g., Hindi-English or French-German), making tokenisation, layout modelling, and language alignment significantly more complex. Despite promising efforts such as LiLT \cite{lilt} and XFUND \cite{layoutxlm}, limitations in cross-lingual pretraining and a lack of high-quality multilingual annotations hinder progress.}

\revise{Latency and computational cost are additional concerns for latency-sensitive applications, such as RPA workflows or mobile-based document scanning tools. Transformer-based VRD-CU models, particularly those requiring high-resolution images or long-sequence attention, pose challenges in real-time scenarios due to memory and processing overhead. Lightweight alternatives, such as LoRA or early-exit mechanisms, have not yet been comprehensively validated in production-grade settings where speed, interpretability, and robustness must be balanced.
Moreover, in regulated domains such as finance and healthcare, data privacy and explainability are essential. The use of large-scale pretrained models raises concerns around data leakage, reproducibility, and the opaque nature of predictions. Current VRD-CU systems often lack mechanisms to provide transparent rationales for extracted content or document-level decisions, posing a barrier for adoption in high-stakes settings.}

\revise{Overall, greater attention must be paid to domain-specific customisation, multilingual document handling, and real-world constraints such as noise, latency, and privacy. Future directions include domain-adaptive pretraining, federated learning for privacy-preserving modelling, and the development of cross-lingual VRD-CU benchmarks that reflect the diversity and complexity of real-world documents.
We also present Table~\ref{tab:model_usecase_mapping}, which illustrates how different classes of VRD-CU models are aligned with domain-specific applications such as finance, healthcare, multilingual administration, and archiving.}

\begin{table}[ht]
\centering
\resizebox{\textwidth}{!}{%
\begin{tabular}{p{3.5cm}|p{4.5cm}|p{6.8cm}}
\toprule
\textbf{Model Class} & \textbf{Representative Models} & \textbf{Practical Application Domains} \\
\midrule
Layout-aware Transformers & LayoutLM, LayoutLMv3, DocFormer & Invoice analysis, form classification, information extraction in finance and healthcare \\
\hline
OCR-free Vision-Language Models & Donut, DESSURT, SimulDoc & Digital archiving of historical documents, handwritten note recognition, multi-script forms \\
\hline
Multilingual and Cross-lingual Models & LiLT, LayoutXLM, Xfund-adapted LayoutLMv3 & Processing multilingual government forms, EU compliance documents, Indian tax filings \\
\hline
High-Resolution Document Models & mPLUG-Owl, UReader, Qwen-VL & Industrial reports, technical diagrams, patents with dense layout or large images \\
\hline
Instruction-tuned Multimodal LLMs & DocPedia, mPLUG-DocOwl, DocGLM-IT & Robotic Process Automation (RPA), document QA, interactive document agents \\
\bottomrule
\end{tabular}%
}
\caption{\revise{Mapping of Model Classes to Real-world VRD-CU Applications}}
\label{tab:model_usecase_mapping}
\end{table}

\section{Conclusion}
\label{sec:conclusion}
This paper comprehensively reviews deep learning-based models for visually rich document content understanding, encompassing both mono-task frameworks designed for specific VRD-CU downstream tasks and multi-task frameworks that support multiple VRD-CU downstream tasks. Beyond introducing the novelties of each model, the limitations of these frameworks are summarized at the end of each section, offering a thorough trend analysis. Additionally, this paper summarizes existing VRD content understanding datasets, pointing out future trends and demands for VRD-CU. To provide a systematic review, we critically discuss various techniques, highlighting their strengths and limitations. We believe that this survey offers a comprehensive overview of the development of VRD-CU, catering to the needs of both the academic and industrial sectors.

\backmatter

\bibliography{sn-bibliography}


\begin{thebibliography}{172}
\ifx \bisbn   \undefined \def \bisbn  #1{ISBN #1}\fi
\ifx \binits  \undefined \def \binits#1{#1}\fi
\ifx \bauthor  \undefined \def \bauthor#1{#1}\fi
\ifx \batitle  \undefined \def \batitle#1{#1}\fi
\ifx \bjtitle  \undefined \def \bjtitle#1{#1}\fi
\ifx \bvolume  \undefined \def \bvolume#1{\textbf{#1}}\fi
\ifx \byear  \undefined \def \byear#1{#1}\fi
\ifx \bissue  \undefined \def \bissue#1{#1}\fi
\ifx \bfpage  \undefined \def \bfpage#1{#1}\fi
\ifx \blpage  \undefined \def \blpage #1{#1}\fi
\ifx \burl  \undefined \def \burl#1{\textsf{#1}}\fi
\ifx \doiurl  \undefined \def \doiurl#1{\url{https://doi.org/#1}}\fi
\ifx \betal  \undefined \def \betal{\textit{et al.}}\fi
\ifx \binstitute  \undefined \def \binstitute#1{#1}\fi
\ifx \binstitutionaled  \undefined \def \binstitutionaled#1{#1}\fi
\ifx \bctitle  \undefined \def \bctitle#1{#1}\fi
\ifx \beditor  \undefined \def \beditor#1{#1}\fi
\ifx \bpublisher  \undefined \def \bpublisher#1{#1}\fi
\ifx \bbtitle  \undefined \def \bbtitle#1{#1}\fi
\ifx \bedition  \undefined \def \bedition#1{#1}\fi
\ifx \bseriesno  \undefined \def \bseriesno#1{#1}\fi
\ifx \blocation  \undefined \def \blocation#1{#1}\fi
\ifx \bsertitle  \undefined \def \bsertitle#1{#1}\fi
\ifx \bsnm \undefined \def \bsnm#1{#1}\fi
\ifx \bsuffix \undefined \def \bsuffix#1{#1}\fi
\ifx \bparticle \undefined \def \bparticle#1{#1}\fi
\ifx \barticle \undefined \def \barticle#1{#1}\fi
\bibcommenthead
\ifx \bconfdate \undefined \def \bconfdate #1{#1}\fi
\ifx \botherref \undefined \def \botherref #1{#1}\fi
\ifx \url \undefined \def \url#1{\textsf{#1}}\fi
\ifx \bchapter \undefined \def \bchapter#1{#1}\fi
\ifx \bbook \undefined \def \bbook#1{#1}\fi
\ifx \bcomment \undefined \def \bcomment#1{#1}\fi
\ifx \oauthor \undefined \def \oauthor#1{#1}\fi
\ifx \citeauthoryear \undefined \def \citeauthoryear#1{#1}\fi
\ifx \endbibitem  \undefined \def \endbibitem {}\fi
\ifx \bconflocation  \undefined \def \bconflocation#1{#1}\fi
\ifx \arxivurl  \undefined \def \arxivurl#1{\textsf{#1}}\fi
\csname PreBibitemsHook\endcsname

\bibitem[\protect\citeauthoryear{Ding et~al.}{2024}]{mmvqa}
\begin{botherref}
\oauthor{\bsnm{Ding}, \binits{Y.}},
\oauthor{\bsnm{Ren}, \binits{K.}},
\oauthor{\bsnm{Huang}, \binits{J.}},
\oauthor{\bsnm{Luo}, \binits{S.}},
\oauthor{\bsnm{Han}, \binits{S.C.}}:
Mvqa: A dataset for multimodal information retrieval in pdf-based visual question answering.
arXiv preprint arXiv:2404.12720
(2024)
\end{botherref}
\endbibitem

\bibitem[\protect\citeauthoryear{Ding et~al.}{2023}]{pdfvqa}
\begin{bchapter}
\bauthor{\bsnm{Ding}, \binits{Y.}},
\bauthor{\bsnm{Luo}, \binits{S.}},
\bauthor{\bsnm{Chung}, \binits{H.}},
\bauthor{\bsnm{Han}, \binits{S.C.}}:
\bctitle{Vqa: A new dataset for real-world vqa on pdf documents}.
In: \bbtitle{Joint European Conference on Machine Learning and Knowledge Discovery in Databases},
pp. \bfpage{585}--\blpage{601}
(\byear{2023}).
\bcomment{Springer}
\end{bchapter}
\endbibitem

\bibitem[\protect\citeauthoryear{Li et~al.}{2020}]{docbank}
\begin{bchapter}
\bauthor{\bsnm{Li}, \binits{M.}},
\bauthor{\bsnm{Xu}, \binits{Y.}},
\bauthor{\bsnm{Cui}, \binits{L.}},
\bauthor{\bsnm{Huang}, \binits{S.}},
\bauthor{\bsnm{Wei}, \binits{F.}},
\bauthor{\bsnm{Li}, \binits{Z.}},
\bauthor{\bsnm{Zhou}, \binits{M.}}:
\bctitle{Docbank: A benchmark dataset for document layout analysis}.
In: \bbtitle{Proceedings of the 28th International Conference on Computational Linguistics},
pp. \bfpage{949}--\blpage{960}
(\byear{2020})
\end{bchapter}
\endbibitem

\bibitem[\protect\citeauthoryear{Ding et~al.}{2025}]{vrdiu}
\begin{botherref}
\oauthor{\bsnm{Ding}, \binits{Y.}},
\oauthor{\bsnm{Han}, \binits{S.C.}},
\oauthor{\bsnm{Li}, \binits{Y.}},
\oauthor{\bsnm{Poon}, \binits{J.}}:
Vrd-iu: Lessons from visually rich document intelligence and understanding.
arXiv preprint arXiv:2506.01388
(2025)
\end{botherref}
\endbibitem

\bibitem[\protect\citeauthoryear{Watanabe et~al.}{1995}]{watanabe1995layout}
\begin{barticle}
\bauthor{\bsnm{Watanabe}, \binits{T.}},
\bauthor{\bsnm{Luo}, \binits{Q.}},
\bauthor{\bsnm{Sugie}, \binits{N.}}:
\batitle{Layout recognition of multi-kinds of table-form documents}.
\bjtitle{IEEE Transactions on Pattern Analysis and Machine Intelligence}
\bvolume{17}(\bissue{4}),
\bfpage{432}--\blpage{445}
(\byear{1995})
\end{barticle}
\endbibitem

\bibitem[\protect\citeauthoryear{Seki et~al.}{2007}]{seki2007information}
\begin{bchapter}
\bauthor{\bsnm{Seki}, \binits{M.}},
\bauthor{\bsnm{Fujio}, \binits{M.}},
\bauthor{\bsnm{Nagasaki}, \binits{T.}},
\bauthor{\bsnm{Shinjo}, \binits{H.}},
\bauthor{\bsnm{Marukawa}, \binits{K.}}:
\bctitle{Information management system using structure analysis of paper/electronic documents and its applications}.
In: \bbtitle{Ninth International Conference on Document Analysis and Recognition (ICDAR 2007)},
vol. \bseriesno{2},
pp. \bfpage{689}--\blpage{693}
(\byear{2007}).
\bcomment{IEEE}
\end{bchapter}
\endbibitem

\bibitem[\protect\citeauthoryear{Rusinol et~al.}{2013}]{rusinol2013field}
\begin{bchapter}
\bauthor{\bsnm{Rusinol}, \binits{M.}},
\bauthor{\bsnm{Benkhelfallah}, \binits{T.}},
\bauthor{\bsnm{Poulain~dAndecy}, \binits{V.}}:
\bctitle{Field extraction from administrative documents by incremental structural templates}.
In: \bbtitle{2013 12th International Conference on Document Analysis and Recognition},
pp. \bfpage{1100}--\blpage{1104}
(\byear{2013}).
\bcomment{IEEE}
\end{bchapter}
\endbibitem

\bibitem[\protect\citeauthoryear{Oliveira and Viana}{2017}]{oliveira2017fast}
\begin{bchapter}
\bauthor{\bsnm{Oliveira}, \binits{D.A.B.}},
\bauthor{\bsnm{Viana}, \binits{M.P.}}:
\bctitle{Fast cnn-based document layout analysis}.
In: \bbtitle{2017 IEEE International Conference on Computer Vision Workshops (ICCVW)},
pp. \bfpage{1173}--\blpage{1180}
(\byear{2017}).
\bcomment{IEEE}
\end{bchapter}
\endbibitem

\bibitem[\protect\citeauthoryear{Katti et~al.}{2018}]{chargrid}
\begin{bchapter}
\bauthor{\bsnm{Katti}, \binits{A.R.}},
\bauthor{\bsnm{Reisswig}, \binits{C.}},
\bauthor{\bsnm{Guder}, \binits{C.}},
\bauthor{\bsnm{Brarda}, \binits{S.}},
\bauthor{\bsnm{Bickel}, \binits{S.}},
\bauthor{\bsnm{H{\"o}hne}, \binits{J.}},
\bauthor{\bsnm{Faddoul}, \binits{J.B.}}:
\bctitle{Chargrid: Towards understanding 2d documents}.
In: \bbtitle{Proceedings of the 2018 Conference on Empirical Methods in Natural Language Processing},
pp. \bfpage{4459}--\blpage{4469}
(\byear{2018})
\end{bchapter}
\endbibitem

\bibitem[\protect\citeauthoryear{Denk and Reisswig}{2019}]{bertgrid}
\begin{botherref}
\oauthor{\bsnm{Denk}, \binits{T.I.}},
\oauthor{\bsnm{Reisswig}, \binits{C.}}:
Bertgrid: Contextualized embedding for 2d document representation and understanding.
arXiv preprint arXiv:1909.04948
(2019)
\end{botherref}
\endbibitem

\bibitem[\protect\citeauthoryear{Yu et~al.}{2021}]{pick}
\begin{bchapter}
\bauthor{\bsnm{Yu}, \binits{W.}},
\bauthor{\bsnm{Lu}, \binits{N.}},
\bauthor{\bsnm{Qi}, \binits{X.}},
\bauthor{\bsnm{Gong}, \binits{P.}},
\bauthor{\bsnm{Xiao}, \binits{R.}}:
\bctitle{Pick: processing key information extraction from documents using improved graph learning-convolutional networks}.
In: \bbtitle{2020 25th International Conference on Pattern Recognition (ICPR)},
pp. \bfpage{4363}--\blpage{4370}
(\byear{2021}).
\bcomment{IEEE}
\end{bchapter}
\endbibitem

\bibitem[\protect\citeauthoryear{Zhang et~al.}{2020}]{trie}
\begin{bchapter}
\bauthor{\bsnm{Zhang}, \binits{P.}},
\bauthor{\bsnm{Xu}, \binits{Y.}},
\bauthor{\bsnm{Cheng}, \binits{Z.}},
\bauthor{\bsnm{Pu}, \binits{S.}},
\bauthor{\bsnm{Lu}, \binits{J.}},
\bauthor{\bsnm{Qiao}, \binits{L.}},
\bauthor{\bsnm{Niu}, \binits{Y.}},
\bauthor{\bsnm{Wu}, \binits{F.}}:
\bctitle{Trie: end-to-end text reading and information extraction for document understanding}.
In: \bbtitle{Proceedings of the 28th ACM International Conference on Multimedia},
pp. \bfpage{1413}--\blpage{1422}
(\byear{2020})
\end{bchapter}
\endbibitem

\bibitem[\protect\citeauthoryear{Wang et~al.}{2021}]{ephoie}
\begin{bchapter}
\bauthor{\bsnm{Wang}, \binits{J.}},
\bauthor{\bsnm{Liu}, \binits{C.}},
\bauthor{\bsnm{Jin}, \binits{L.}},
\bauthor{\bsnm{Tang}, \binits{G.}},
\bauthor{\bsnm{Zhang}, \binits{J.}},
\bauthor{\bsnm{Zhang}, \binits{S.}},
\bauthor{\bsnm{Wang}, \binits{Q.}},
\bauthor{\bsnm{Wu}, \binits{Y.}},
\bauthor{\bsnm{Cai}, \binits{M.}}:
\bctitle{Towards robust visual information extraction in real world: new dataset and novel solution}.
In: \bbtitle{Proceedings of the AAAI Conference on Artificial Intelligence},
vol. \bseriesno{35},
pp. \bfpage{2738}--\blpage{2745}
(\byear{2021})
\end{bchapter}
\endbibitem

\bibitem[\protect\citeauthoryear{Gu et~al.}{2021}]{udoc}
\begin{barticle}
\bauthor{\bsnm{Gu}, \binits{J.}},
\bauthor{\bsnm{Kuen}, \binits{J.}},
\bauthor{\bsnm{Morariu}, \binits{V.I.}},
\bauthor{\bsnm{Zhao}, \binits{H.}},
\bauthor{\bsnm{Jain}, \binits{R.}},
\bauthor{\bsnm{Barmpalios}, \binits{N.}},
\bauthor{\bsnm{Nenkova}, \binits{A.}},
\bauthor{\bsnm{Sun}, \binits{T.}}:
\batitle{Unidoc: Unified pretraining framework for document understanding}.
\bjtitle{Advances in Neural Information Processing Systems}
\bvolume{34},
\bfpage{39}--\blpage{50}
(\byear{2021})
\end{barticle}
\endbibitem

\bibitem[\protect\citeauthoryear{Li et~al.}{2021a}]{selfdoc}
\begin{bchapter}
\bauthor{\bsnm{Li}, \binits{P.}},
\bauthor{\bsnm{Gu}, \binits{J.}},
\bauthor{\bsnm{Kuen}, \binits{J.}},
\bauthor{\bsnm{Morariu}, \binits{V.I.}},
\bauthor{\bsnm{Zhao}, \binits{H.}},
\bauthor{\bsnm{Jain}, \binits{R.}},
\bauthor{\bsnm{Manjunatha}, \binits{V.}},
\bauthor{\bsnm{Liu}, \binits{H.}}:
\bctitle{Selfdoc: Self-supervised document representation learning}.
In: \bbtitle{Proceedings of the IEEE/CVF Conference on Computer Vision and Pattern Recognition},
pp. \bfpage{5652}--\blpage{5660}
(\byear{2021})
\end{bchapter}
\endbibitem

\bibitem[\protect\citeauthoryear{Li et~al.}{2021b}]{structext}
\begin{bchapter}
\bauthor{\bsnm{Li}, \binits{Y.}},
\bauthor{\bsnm{Qian}, \binits{Y.}},
\bauthor{\bsnm{Yu}, \binits{Y.}},
\bauthor{\bsnm{Qin}, \binits{X.}},
\bauthor{\bsnm{Zhang}, \binits{C.}},
\bauthor{\bsnm{Liu}, \binits{Y.}},
\bauthor{\bsnm{Yao}, \binits{K.}},
\bauthor{\bsnm{Han}, \binits{J.}},
\bauthor{\bsnm{Liu}, \binits{J.}},
\bauthor{\bsnm{Ding}, \binits{E.}}:
\bctitle{Structext: Structured text understanding with multi-modal transformers}.
In: \bbtitle{Proceedings of the 29th ACM International Conference on Multimedia},
pp. \bfpage{1912}--\blpage{1920}
(\byear{2021})
\end{bchapter}
\endbibitem

\bibitem[\protect\citeauthoryear{Yu et~al.}{2022}]{structextv2}
\begin{bchapter}
\bauthor{\bsnm{Yu}, \binits{Y.}},
\bauthor{\bsnm{Li}, \binits{Y.}},
\bauthor{\bsnm{Zhang}, \binits{C.}},
\bauthor{\bsnm{Zhang}, \binits{X.}},
\bauthor{\bsnm{Guo}, \binits{Z.}},
\bauthor{\bsnm{Qin}, \binits{X.}},
\bauthor{\bsnm{Yao}, \binits{K.}},
\bauthor{\bsnm{Han}, \binits{J.}},
\bauthor{\bsnm{Ding}, \binits{E.}},
\bauthor{\bsnm{Wang}, \binits{J.}}:
\bctitle{Structextv2: Masked visual-textual prediction for document image pre-training}.
In: \bbtitle{The Eleventh International Conference on Learning Representations}
(\byear{2022})
\end{bchapter}
\endbibitem

\bibitem[\protect\citeauthoryear{Bai et~al.}{2022}]{wukong}
\begin{botherref}
\oauthor{\bsnm{Bai}, \binits{H.}},
\oauthor{\bsnm{Liu}, \binits{Z.}},
\oauthor{\bsnm{Meng}, \binits{X.}},
\oauthor{\bsnm{Li}, \binits{W.}},
\oauthor{\bsnm{Liu}, \binits{S.}},
\oauthor{\bsnm{Xie}, \binits{N.}},
\oauthor{\bsnm{Zheng}, \binits{R.}},
\oauthor{\bsnm{Wang}, \binits{L.}},
\oauthor{\bsnm{Hou}, \binits{L.}},
\oauthor{\bsnm{Wei}, \binits{J.}}, et al.:
Wukong-reader: Multi-modal pre-training for fine-grained visual document understanding.
arXiv preprint arXiv:2212.09621
(2022)
\end{botherref}
\endbibitem

\bibitem[\protect\citeauthoryear{Luo et~al.}{2024}]{layoutllm}
\begin{bchapter}
\bauthor{\bsnm{Luo}, \binits{C.}},
\bauthor{\bsnm{Shen}, \binits{Y.}},
\bauthor{\bsnm{Zhu}, \binits{Z.}},
\bauthor{\bsnm{Zheng}, \binits{Q.}},
\bauthor{\bsnm{Yu}, \binits{Z.}},
\bauthor{\bsnm{Yao}, \binits{C.}}:
\bctitle{Layoutllm: Layout instruction tuning with large language models for document understanding}.
In: \bbtitle{Proceedings of the IEEE/CVF Conference on Computer Vision and Pattern Recognition},
pp. \bfpage{15630}--\blpage{15640}
(\byear{2024})
\end{bchapter}
\endbibitem

\bibitem[\protect\citeauthoryear{Liu et~al.}{2024}]{hrvda}
\begin{bchapter}
\bauthor{\bsnm{Liu}, \binits{C.}},
\bauthor{\bsnm{Yin}, \binits{K.}},
\bauthor{\bsnm{Cao}, \binits{H.}},
\bauthor{\bsnm{Jiang}, \binits{X.}},
\bauthor{\bsnm{Li}, \binits{X.}},
\bauthor{\bsnm{Liu}, \binits{Y.}},
\bauthor{\bsnm{Jiang}, \binits{D.}},
\bauthor{\bsnm{Sun}, \binits{X.}},
\bauthor{\bsnm{Xu}, \binits{L.}}:
\bctitle{Hrvda: High-resolution visual document assistant}.
In: \bbtitle{Proceedings of the IEEE/CVF Conference on Computer Vision and Pattern Recognition},
pp. \bfpage{15534}--\blpage{15545}
(\byear{2024})
\end{bchapter}
\endbibitem

\bibitem[\protect\citeauthoryear{Subramani et~al.}{2020}]{subramani2020survey}
\begin{botherref}
\oauthor{\bsnm{Subramani}, \binits{N.}},
\oauthor{\bsnm{Matton}, \binits{A.}},
\oauthor{\bsnm{Greaves}, \binits{M.}},
\oauthor{\bsnm{Lam}, \binits{A.}}:
A survey of deep learning approaches for ocr and document understanding.
arXiv preprint arXiv:2011.13534
(2020)
\end{botherref}
\endbibitem

\bibitem[\protect\citeauthoryear{Liu et~al.}{2023}]{liu2023tabular}
\begin{barticle}
\bauthor{\bsnm{Liu}, \binits{J.}},
\bauthor{\bsnm{Chabot}, \binits{Y.}},
\bauthor{\bsnm{Troncy}, \binits{R.}},
\bauthor{\bsnm{Huynh}, \binits{V.-P.}},
\bauthor{\bsnm{Labb{\'e}}, \binits{T.}},
\bauthor{\bsnm{Monnin}, \binits{P.}}:
\batitle{From tabular data to knowledge graphs: A survey of semantic table interpretation tasks and methods}.
\bjtitle{Journal of Web Semantics}
\bvolume{76},
\bfpage{100761}
(\byear{2023})
\end{barticle}
\endbibitem

\bibitem[\protect\citeauthoryear{Ehrmann et~al.}{2023}]{ehrmann2023named}
\begin{barticle}
\bauthor{\bsnm{Ehrmann}, \binits{M.}},
\bauthor{\bsnm{Hamdi}, \binits{A.}},
\bauthor{\bsnm{Pontes}, \binits{E.L.}},
\bauthor{\bsnm{Romanello}, \binits{M.}},
\bauthor{\bsnm{Doucet}, \binits{A.}}:
\batitle{Named entity recognition and classification in historical documents: A survey}.
\bjtitle{ACM Computing Surveys}
\bvolume{56}(\bissue{2}),
\bfpage{1}--\blpage{47}
(\byear{2023})
\end{barticle}
\endbibitem

\bibitem[\protect\citeauthoryear{Lombardi and Marinai}{2020}]{lombardi2020deep}
\begin{barticle}
\bauthor{\bsnm{Lombardi}, \binits{F.}},
\bauthor{\bsnm{Marinai}, \binits{S.}}:
\batitle{Deep learning for historical document analysis and recognition—a survey}.
\bjtitle{Journal of Imaging}
\bvolume{6}(\bissue{10}),
\bfpage{110}
(\byear{2020})
\end{barticle}
\endbibitem

\bibitem[\protect\citeauthoryear{Saout et~al.}{2024}]{saout2024overview}
\begin{botherref}
\oauthor{\bsnm{Saout}, \binits{T.}},
\oauthor{\bsnm{Lardeux}, \binits{F.}},
\oauthor{\bsnm{Saubion}, \binits{F.}}:
An overview of data extraction from invoices.
IEEE Access
(2024)
\end{botherref}
\endbibitem

\bibitem[\protect\citeauthoryear{Cui et~al.}{2021}]{cui2021document}
\begin{botherref}
\oauthor{\bsnm{Cui}, \binits{L.}},
\oauthor{\bsnm{Xu}, \binits{Y.}},
\oauthor{\bsnm{Lv}, \binits{T.}},
\oauthor{\bsnm{Wei}, \binits{F.}}:
Document ai: Benchmarks, models and applications.
arXiv preprint arXiv:2111.08609
(2021)
\end{botherref}
\endbibitem

\bibitem[\protect\citeauthoryear{O'Gorman}{1993}]{o1993document}
\begin{barticle}
\bauthor{\bsnm{O'Gorman}, \binits{L.}}:
\batitle{The document spectrum for page layout analysis}.
\bjtitle{IEEE Transactions on pattern analysis and machine intelligence}
\bvolume{15}(\bissue{11}),
\bfpage{1162}--\blpage{1173}
(\byear{1993})
\end{barticle}
\endbibitem

\bibitem[\protect\citeauthoryear{Yang et~al.}{2017}]{yang2017learning}
\begin{bchapter}
\bauthor{\bsnm{Yang}, \binits{X.}},
\bauthor{\bsnm{Yumer}, \binits{E.}},
\bauthor{\bsnm{Asente}, \binits{P.}},
\bauthor{\bsnm{Kraley}, \binits{M.}},
\bauthor{\bsnm{Kifer}, \binits{D.}},
\bauthor{\bsnm{Lee~Giles}, \binits{C.}}:
\bctitle{Learning to extract semantic structure from documents using multimodal fully convolutional neural networks}.
In: \bbtitle{Proceedings of the IEEE Conference on Computer Vision and Pattern Recognition},
pp. \bfpage{5315}--\blpage{5324}
(\byear{2017})
\end{bchapter}
\endbibitem

\bibitem[\protect\citeauthoryear{Devlin}{2018}]{bert}
\begin{bchapter}
\bauthor{\bsnm{Devlin}, \binits{J.}}:
\bctitle{Bert: Pre-training of deep bidirectional transformers for language understanding.}
In: \bbtitle{Proceedings of NAACL-HLT},
vol. \bseriesno{2019},
p. \bfpage{4171}
(\byear{2018})
\end{bchapter}
\endbibitem

\bibitem[\protect\citeauthoryear{Liu et~al.}{2019}]{roberta}
\begin{botherref}
\oauthor{\bsnm{Liu}, \binits{Y.}},
\oauthor{\bsnm{Ott}, \binits{M.}},
\oauthor{\bsnm{Goyal}, \binits{N.}},
\oauthor{\bsnm{Du}, \binits{J.}},
\oauthor{\bsnm{Joshi}, \binits{M.}},
\oauthor{\bsnm{Chen}, \binits{D.}},
\oauthor{\bsnm{Levy}, \binits{O.}},
\oauthor{\bsnm{Lewis}, \binits{M.}},
\oauthor{\bsnm{Zettlemoyer}, \binits{L.}},
\oauthor{\bsnm{Stoyanov}, \binits{V.}}:
Roberta: A robustly optimized bert pretraining approach.
arXiv preprint arXiv:1907.11692
(2019)
\end{botherref}
\endbibitem

\bibitem[\protect\citeauthoryear{Ren et~al.}{2015}]{fasterrcnn}
\begin{botherref}
\oauthor{\bsnm{Ren}, \binits{S.}},
\oauthor{\bsnm{He}, \binits{K.}},
\oauthor{\bsnm{Girshick}, \binits{R.}},
\oauthor{\bsnm{Sun}, \binits{J.}}:
Faster r-cnn: Towards real-time object detection with region proposal networks.
Advances in neural information processing systems
\textbf{28}
(2015)
\end{botherref}
\endbibitem

\bibitem[\protect\citeauthoryear{He et~al.}{2017}]{maskrcnn}
\begin{bchapter}
\bauthor{\bsnm{He}, \binits{K.}},
\bauthor{\bsnm{Gkioxari}, \binits{G.}},
\bauthor{\bsnm{Doll{\'a}r}, \binits{P.}},
\bauthor{\bsnm{Girshick}, \binits{R.}}:
\bctitle{Mask r-cnn}.
In: \bbtitle{Proceedings of the IEEE International Conference on Computer Vision},
pp. \bfpage{2961}--\blpage{2969}
(\byear{2017})
\end{bchapter}
\endbibitem

\bibitem[\protect\citeauthoryear{Wang et~al.}{2020}]{docstruct}
\begin{botherref}
\oauthor{\bsnm{Wang}, \binits{Z.}},
\oauthor{\bsnm{Zhan}, \binits{M.}},
\oauthor{\bsnm{Liu}, \binits{X.}},
\oauthor{\bsnm{Liang}, \binits{D.}}:
Docstruct: A multimodal method to extract hierarchy structure in document for general form understanding.
arXiv preprint arXiv:2010.11685
(2020)
\end{botherref}
\endbibitem

\bibitem[\protect\citeauthoryear{Xu et~al.}{2020}]{layoutlm}
\begin{bchapter}
\bauthor{\bsnm{Xu}, \binits{Y.}},
\bauthor{\bsnm{Li}, \binits{M.}},
\bauthor{\bsnm{Cui}, \binits{L.}},
\bauthor{\bsnm{Huang}, \binits{S.}},
\bauthor{\bsnm{Wei}, \binits{F.}},
\bauthor{\bsnm{Zhou}, \binits{M.}}:
\bctitle{Layoutlm: Pre-training of text and layout for document image understanding}.
In: \bbtitle{Proceedings of the 26th ACM SIGKDD International Conference on Knowledge Discovery \& Data Mining},
pp. \bfpage{1192}--\blpage{1200}
(\byear{2020})
\end{bchapter}
\endbibitem

\bibitem[\protect\citeauthoryear{Lee et~al.}{2022}]{formnet}
\begin{bchapter}
\bauthor{\bsnm{Lee}, \binits{C.-Y.}},
\bauthor{\bsnm{Li}, \binits{C.-L.}},
\bauthor{\bsnm{Dozat}, \binits{T.}},
\bauthor{\bsnm{Perot}, \binits{V.}},
\bauthor{\bsnm{Su}, \binits{G.}},
\bauthor{\bsnm{Hua}, \binits{N.}},
\bauthor{\bsnm{Ainslie}, \binits{J.}},
\bauthor{\bsnm{Wang}, \binits{R.}},
\bauthor{\bsnm{Fujii}, \binits{Y.}},
\bauthor{\bsnm{Pfister}, \binits{T.}}:
\bctitle{Formnet: Structural encoding beyond sequential modeling in form document information extraction}.
In: \bbtitle{Proceedings of the 60th Annual Meeting of the Association for Computational Linguistics (Volume 1: Long Papers)},
pp. \bfpage{3735}--\blpage{3754}
(\byear{2022})
\end{bchapter}
\endbibitem

\bibitem[\protect\citeauthoryear{Majumder et~al.}{2020}]{majumder2020representation}
\begin{bchapter}
\bauthor{\bsnm{Majumder}, \binits{B.P.}},
\bauthor{\bsnm{Potti}, \binits{N.}},
\bauthor{\bsnm{Tata}, \binits{S.}},
\bauthor{\bsnm{Wendt}, \binits{J.B.}},
\bauthor{\bsnm{Zhao}, \binits{Q.}},
\bauthor{\bsnm{Najork}, \binits{M.}}:
\bctitle{Representation learning for information extraction from form-like documents}.
In: \bbtitle{Proceedings of the 58th Annual Meeting of the Association for Computational Linguistics},
pp. \bfpage{6495}--\blpage{6504}
(\byear{2020})
\end{bchapter}
\endbibitem

\bibitem[\protect\citeauthoryear{He et~al.}{2023}]{icld3ie}
\begin{bchapter}
\bauthor{\bsnm{He}, \binits{J.}},
\bauthor{\bsnm{Wang}, \binits{L.}},
\bauthor{\bsnm{Hu}, \binits{Y.}},
\bauthor{\bsnm{Liu}, \binits{N.}},
\bauthor{\bsnm{Liu}, \binits{H.}},
\bauthor{\bsnm{Xu}, \binits{X.}},
\bauthor{\bsnm{Shen}, \binits{H.T.}}:
\bctitle{Icl-d3ie: In-context learning with diverse demonstrations updating for document information extraction}.
In: \bbtitle{Proceedings of the IEEE/CVF International Conference on Computer Vision},
pp. \bfpage{19485}--\blpage{19494}
(\byear{2023})
\end{bchapter}
\endbibitem

\bibitem[\protect\citeauthoryear{Lee et~al.}{2023}]{formnetv2}
\begin{botherref}
\oauthor{\bsnm{Lee}, \binits{C.-Y.}},
\oauthor{\bsnm{Li}, \binits{C.-L.}},
\oauthor{\bsnm{Zhang}, \binits{H.}},
\oauthor{\bsnm{Dozat}, \binits{T.}},
\oauthor{\bsnm{Perot}, \binits{V.}},
\oauthor{\bsnm{Su}, \binits{G.}},
\oauthor{\bsnm{Zhang}, \binits{X.}},
\oauthor{\bsnm{Sohn}, \binits{K.}},
\oauthor{\bsnm{Glushnev}, \binits{N.}},
\oauthor{\bsnm{Wang}, \binits{R.}}, et al.:
Formnetv2: Multimodal graph contrastive learning for form document information extraction.
arXiv preprint arXiv:2305.02549
(2023)
\end{botherref}
\endbibitem

\bibitem[\protect\citeauthoryear{Wang and Shang}{2022}]{laser}
\begin{bchapter}
\bauthor{\bsnm{Wang}, \binits{Z.}},
\bauthor{\bsnm{Shang}, \binits{J.}}:
\bctitle{Towards few-shot entity recognition in document images: A label-aware sequence-to-sequence framework}.
In: \bbtitle{Findings of the Association for Computational Linguistics: ACL 2022},
pp. \bfpage{4174}--\blpage{4186}
(\byear{2022})
\end{bchapter}
\endbibitem

\bibitem[\protect\citeauthoryear{Chen et~al.}{2023}]{chen2023task}
\begin{bchapter}
\bauthor{\bsnm{Chen}, \binits{J.}},
\bauthor{\bsnm{Dai}, \binits{H.}},
\bauthor{\bsnm{Dai}, \binits{B.}},
\bauthor{\bsnm{Zhang}, \binits{A.}},
\bauthor{\bsnm{Wei}, \binits{W.}}:
\bctitle{On task-personalized multimodal few-shot learning for visually-rich document entity retrieval}.
In: \bbtitle{Findings of the Association for Computational Linguistics: EMNLP 2023},
pp. \bfpage{9006}--\blpage{9025}
(\byear{2023})
\end{bchapter}
\endbibitem

\bibitem[\protect\citeauthoryear{Cao et~al.}{2023}]{genkie}
\begin{bchapter}
\bauthor{\bsnm{Cao}, \binits{P.}},
\bauthor{\bsnm{Wang}, \binits{Y.}},
\bauthor{\bsnm{Zhang}, \binits{Q.}},
\bauthor{\bsnm{Meng}, \binits{Z.}}:
\bctitle{Genkie: Robust generative multimodal document key information extraction}.
In: \bbtitle{Findings of the Association for Computational Linguistics: EMNLP 2023},
pp. \bfpage{14702}--\blpage{14713}
(\byear{2023})
\end{bchapter}
\endbibitem

\bibitem[\protect\citeauthoryear{Zhang et~al.}{2021}]{zhang2021entity}
\begin{bchapter}
\bauthor{\bsnm{Zhang}, \binits{Y.}},
\bauthor{\bsnm{Bo}, \binits{Z.}},
\bauthor{\bsnm{Wang}, \binits{R.}},
\bauthor{\bsnm{Cao}, \binits{J.}},
\bauthor{\bsnm{Li}, \binits{C.}},
\bauthor{\bsnm{Bao}, \binits{Z.}}:
\bctitle{Entity relation extraction as dependency parsing in visually rich documents}.
In: \bbtitle{Proceedings of the 2021 Conference on Empirical Methods in Natural Language Processing},
pp. \bfpage{2759}--\blpage{2768}
(\byear{2021})
\end{bchapter}
\endbibitem

\bibitem[\protect\citeauthoryear{Hu et~al.}{2023}]{kvpformer}
\begin{bchapter}
\bauthor{\bsnm{Hu}, \binits{K.}},
\bauthor{\bsnm{Wu}, \binits{Z.}},
\bauthor{\bsnm{Zhong}, \binits{Z.}},
\bauthor{\bsnm{Lin}, \binits{W.}},
\bauthor{\bsnm{Sun}, \binits{L.}},
\bauthor{\bsnm{Huo}, \binits{Q.}}:
\bctitle{A question-answering approach to key value pair extraction from form-like document images}.
In: \bbtitle{Proceedings of the AAAI Conference on Artificial Intelligence},
vol. \bseriesno{37},
pp. \bfpage{12899}--\blpage{12906}
(\byear{2023})
\end{bchapter}
\endbibitem

\bibitem[\protect\citeauthoryear{Carbonell et~al.}{2021}]{carbonell2021named}
\begin{bchapter}
\bauthor{\bsnm{Carbonell}, \binits{M.}},
\bauthor{\bsnm{Riba}, \binits{P.}},
\bauthor{\bsnm{Villegas}, \binits{M.}},
\bauthor{\bsnm{Forn{\'e}s}, \binits{A.}},
\bauthor{\bsnm{Llad{\'o}s}, \binits{J.}}:
\bctitle{Named entity recognition and relation extraction with graph neural networks in semi structured documents}.
In: \bbtitle{2020 25th International Conference on Pattern Recognition (ICPR)},
pp. \bfpage{9622}--\blpage{9627}
(\byear{2021}).
\bcomment{IEEE}
\end{bchapter}
\endbibitem

\bibitem[\protect\citeauthoryear{Wang et~al.}{2022}]{lilt}
\begin{bchapter}
\bauthor{\bsnm{Wang}, \binits{J.}},
\bauthor{\bsnm{Jin}, \binits{L.}},
\bauthor{\bsnm{Ding}, \binits{K.}}:
\bctitle{Lilt: A simple yet effective language-independent layout transformer for structured document understanding}.
In: \bbtitle{Proceedings of the 60th Annual Meeting of the Association for Computational Linguistics (Volume 1: Long Papers)},
pp. \bfpage{7747}--\blpage{7757}
(\byear{2022})
\end{bchapter}
\endbibitem

\bibitem[\protect\citeauthoryear{Tu et~al.}{2023}]{layoutmask}
\begin{bchapter}
\bauthor{\bsnm{Tu}, \binits{Y.}},
\bauthor{\bsnm{Guo}, \binits{Y.}},
\bauthor{\bsnm{Chen}, \binits{H.}},
\bauthor{\bsnm{Tang}, \binits{J.}}:
\bctitle{Layoutmask: Enhance text-layout interaction in multi-modal pre-training for document understanding}.
In: \bbtitle{Proceedings of the 61st Annual Meeting of the Association for Computational Linguistics (Volume 1: Long Papers)},
pp. \bfpage{15200}--\blpage{15212}
(\byear{2023})
\end{bchapter}
\endbibitem

\bibitem[\protect\citeauthoryear{Xu et~al.}{2020}]{layoutlmv2}
\begin{botherref}
\oauthor{\bsnm{Xu}, \binits{Y.}},
\oauthor{\bsnm{Xu}, \binits{Y.}},
\oauthor{\bsnm{Lv}, \binits{T.}},
\oauthor{\bsnm{Cui}, \binits{L.}},
\oauthor{\bsnm{Wei}, \binits{F.}},
\oauthor{\bsnm{Wang}, \binits{G.}},
\oauthor{\bsnm{Lu}, \binits{Y.}},
\oauthor{\bsnm{Florencio}, \binits{D.}},
\oauthor{\bsnm{Zhang}, \binits{C.}},
\oauthor{\bsnm{Che}, \binits{W.}}, et al.:
Layoutlmv2: Multi-modal pre-training for visually-rich document understanding.
arXiv preprint arXiv:2012.14740
(2020)
\end{botherref}
\endbibitem

\bibitem[\protect\citeauthoryear{Huang et~al.}{2022}]{layoutlmv3}
\begin{bchapter}
\bauthor{\bsnm{Huang}, \binits{Y.}},
\bauthor{\bsnm{Lv}, \binits{T.}},
\bauthor{\bsnm{Cui}, \binits{L.}},
\bauthor{\bsnm{Lu}, \binits{Y.}},
\bauthor{\bsnm{Wei}, \binits{F.}}:
\bctitle{Layoutlmv3: Pre-training for document ai with unified text and image masking}.
In: \bbtitle{Proceedings of the 30th ACM International Conference on Multimedia},
pp. \bfpage{4083}--\blpage{4091}
(\byear{2022})
\end{bchapter}
\endbibitem

\bibitem[\protect\citeauthoryear{Jaume et~al.}{2019}]{funsd}
\begin{bchapter}
\bauthor{\bsnm{Jaume}, \binits{G.}},
\bauthor{\bsnm{Ekenel}, \binits{H.K.}},
\bauthor{\bsnm{Thiran}, \binits{J.-P.}}:
\bctitle{Funsd: A dataset for form understanding in noisy scanned documents}.
In: \bbtitle{2019 International Conference on Document Analysis and Recognition Workshops (ICDARW)},
vol. \bseriesno{2},
pp. \bfpage{1}--\blpage{6}
(\byear{2019}).
\bcomment{IEEE}
\end{bchapter}
\endbibitem

\bibitem[\protect\citeauthoryear{Park et~al.}{2019}]{cord}
\begin{bchapter}
\bauthor{\bsnm{Park}, \binits{S.}},
\bauthor{\bsnm{Shin}, \binits{S.}},
\bauthor{\bsnm{Lee}, \binits{B.}},
\bauthor{\bsnm{Lee}, \binits{J.}},
\bauthor{\bsnm{Surh}, \binits{J.}},
\bauthor{\bsnm{Seo}, \binits{M.}},
\bauthor{\bsnm{Lee}, \binits{H.}}:
\bctitle{Cord: a consolidated receipt dataset for post-ocr parsing}.
In: \bbtitle{Workshop on Document Intelligence at NeurIPS 2019}
(\byear{2019})
\end{bchapter}
\endbibitem

\bibitem[\protect\citeauthoryear{Huang et~al.}{2019}]{sroie}
\begin{bchapter}
\bauthor{\bsnm{Huang}, \binits{Z.}},
\bauthor{\bsnm{Chen}, \binits{K.}},
\bauthor{\bsnm{He}, \binits{J.}},
\bauthor{\bsnm{Bai}, \binits{X.}},
\bauthor{\bsnm{Karatzas}, \binits{D.}},
\bauthor{\bsnm{Lu}, \binits{S.}},
\bauthor{\bsnm{Jawahar}, \binits{C.}}:
\bctitle{Icdar2019 competition on scanned receipt ocr and information extraction}.
In: \bbtitle{2019 International Conference on Document Analysis and Recognition (ICDAR)},
pp. \bfpage{1516}--\blpage{1520}
(\byear{2019}).
\bcomment{IEEE}
\end{bchapter}
\endbibitem

\bibitem[\protect\citeauthoryear{Mathew et~al.}{2021}]{docvqa}
\begin{bchapter}
\bauthor{\bsnm{Mathew}, \binits{M.}},
\bauthor{\bsnm{Karatzas}, \binits{D.}},
\bauthor{\bsnm{Jawahar}, \binits{C.}}:
\bctitle{Docvqa: A dataset for vqa on document images}.
In: \bbtitle{Proceedings of the IEEE/CVF Winter Conference on Applications of Computer Vision},
pp. \bfpage{2200}--\blpage{2209}
(\byear{2021})
\end{bchapter}
\endbibitem

\bibitem[\protect\citeauthoryear{Harley et~al.}{2015}]{rvlcdip}
\begin{bchapter}
\bauthor{\bsnm{Harley}, \binits{A.W.}},
\bauthor{\bsnm{Ufkes}, \binits{A.}},
\bauthor{\bsnm{Derpanis}, \binits{K.G.}}:
\bctitle{Evaluation of deep convolutional nets for document image classification and retrieval}.
In: \bbtitle{2015 13th International Conference on Document Analysis and Recognition (ICDAR)},
pp. \bfpage{991}--\blpage{995}
(\byear{2015}).
\bcomment{IEEE}
\end{bchapter}
\endbibitem

\bibitem[\protect\citeauthoryear{Tang et~al.}{2022}]{udop}
\begin{botherref}
\oauthor{\bsnm{Tang}, \binits{Z.}},
\oauthor{\bsnm{Yang}, \binits{Z.}},
\oauthor{\bsnm{Wang}, \binits{G.}},
\oauthor{\bsnm{Fang}, \binits{Y.}},
\oauthor{\bsnm{Liu}, \binits{Y.}},
\oauthor{\bsnm{Zhu}, \binits{C.}},
\oauthor{\bsnm{Zeng}, \binits{M.}},
\oauthor{\bsnm{Zhang}, \binits{C.}},
\oauthor{\bsnm{Bansal}, \binits{M.}}:
Unifying vision, text, and layout for universal document processing.
arXiv preprint arXiv:2212.02623
(2022)
\end{botherref}
\endbibitem

\bibitem[\protect\citeauthoryear{Kim et~al.}{2022}]{donut}
\begin{bchapter}
\bauthor{\bsnm{Kim}, \binits{G.}},
\bauthor{\bsnm{Hong}, \binits{T.}},
\bauthor{\bsnm{Yim}, \binits{M.}},
\bauthor{\bsnm{Nam}, \binits{J.}},
\bauthor{\bsnm{Park}, \binits{J.}},
\bauthor{\bsnm{Yim}, \binits{J.}},
\bauthor{\bsnm{Hwang}, \binits{W.}},
\bauthor{\bsnm{Yun}, \binits{S.}},
\bauthor{\bsnm{Han}, \binits{D.}},
\bauthor{\bsnm{Park}, \binits{S.}}:
\bctitle{Ocr-free document understanding transformer}.
In: \bbtitle{Computer Vision--ECCV 2022: 17th European Conference, Tel Aviv, Israel, October 23--27, 2022, Proceedings, Part XXVIII},
pp. \bfpage{498}--\blpage{517}
(\byear{2022}).
\bcomment{Springer}
\end{bchapter}
\endbibitem

\bibitem[\protect\citeauthoryear{Davis et~al.}{2022}]{dessurt}
\begin{bchapter}
\bauthor{\bsnm{Davis}, \binits{B.}},
\bauthor{\bsnm{Morse}, \binits{B.}},
\bauthor{\bsnm{Price}, \binits{B.}},
\bauthor{\bsnm{Tensmeyer}, \binits{C.}},
\bauthor{\bsnm{Wigington}, \binits{C.}},
\bauthor{\bsnm{Morariu}, \binits{V.}}:
\bctitle{End-to-end document recognition and understanding with dessurt}.
In: \bbtitle{European Conference on Computer Vision},
pp. \bfpage{280}--\blpage{296}
(\byear{2022}).
\bcomment{Springer}
\end{bchapter}
\endbibitem

\bibitem[\protect\citeauthoryear{Dosovitskiy et~al.}{2020}]{vit}
\begin{bchapter}
\bauthor{\bsnm{Dosovitskiy}, \binits{A.}},
\bauthor{\bsnm{Beyer}, \binits{L.}},
\bauthor{\bsnm{Kolesnikov}, \binits{A.}},
\bauthor{\bsnm{Weissenborn}, \binits{D.}},
\bauthor{\bsnm{Zhai}, \binits{X.}},
\bauthor{\bsnm{Unterthiner}, \binits{T.}},
\bauthor{\bsnm{Dehghani}, \binits{M.}},
\bauthor{\bsnm{Minderer}, \binits{M.}},
\bauthor{\bsnm{Heigold}, \binits{G.}},
\bauthor{\bsnm{Gelly}, \binits{S.}}, \betal:
\bctitle{An image is worth 16x16 words: Transformers for image recognition at scale}.
In: \bbtitle{International Conference on Learning Representations}
(\byear{2020})
\end{bchapter}
\endbibitem

\bibitem[\protect\citeauthoryear{Liu et~al.}{2021}]{liu2021swin}
\begin{bchapter}
\bauthor{\bsnm{Liu}, \binits{Z.}},
\bauthor{\bsnm{Lin}, \binits{Y.}},
\bauthor{\bsnm{Cao}, \binits{Y.}},
\bauthor{\bsnm{Hu}, \binits{H.}},
\bauthor{\bsnm{Wei}, \binits{Y.}},
\bauthor{\bsnm{Zhang}, \binits{Z.}},
\bauthor{\bsnm{Lin}, \binits{S.}},
\bauthor{\bsnm{Guo}, \binits{B.}}:
\bctitle{Swin transformer: Hierarchical vision transformer using shifted windows}.
In: \bbtitle{Proceedings of the IEEE/CVF International Conference on Computer Vision},
pp. \bfpage{10012}--\blpage{10022}
(\byear{2021})
\end{bchapter}
\endbibitem

\bibitem[\protect\citeauthoryear{Zhao et~al.}{2019}]{cutie}
\begin{botherref}
\oauthor{\bsnm{Zhao}, \binits{X.}},
\oauthor{\bsnm{Niu}, \binits{E.}},
\oauthor{\bsnm{Wu}, \binits{Z.}},
\oauthor{\bsnm{Wang}, \binits{X.}}:
Cutie: Learning to understand documents with convolutional universal text information extractor.
arXiv preprint arXiv:1903.12363
(2019)
\end{botherref}
\endbibitem

\bibitem[\protect\citeauthoryear{Palm et~al.}{2019}]{palm2019attend}
\begin{bchapter}
\bauthor{\bsnm{Palm}, \binits{R.B.}},
\bauthor{\bsnm{Laws}, \binits{F.}},
\bauthor{\bsnm{Winther}, \binits{O.}}:
\bctitle{Attend, copy, parse end-to-end information extraction from documents}.
In: \bbtitle{2019 International Conference on Document Analysis and Recognition (ICDAR)},
pp. \bfpage{329}--\blpage{336}
(\byear{2019}).
\bcomment{IEEE}
\end{bchapter}
\endbibitem

\bibitem[\protect\citeauthoryear{Gu et~al.}{2022}]{xylayoutlm}
\begin{bchapter}
\bauthor{\bsnm{Gu}, \binits{Z.}},
\bauthor{\bsnm{Meng}, \binits{C.}},
\bauthor{\bsnm{Wang}, \binits{K.}},
\bauthor{\bsnm{Lan}, \binits{J.}},
\bauthor{\bsnm{Wang}, \binits{W.}},
\bauthor{\bsnm{Gu}, \binits{M.}},
\bauthor{\bsnm{Zhang}, \binits{L.}}:
\bctitle{Xylayoutlm: Towards layout-aware multimodal networks for visually-rich document understanding}.
In: \bbtitle{Proceedings of the IEEE/CVF Conference on Computer Vision and Pattern Recognition},
pp. \bfpage{4583}--\blpage{4592}
(\byear{2022})
\end{bchapter}
\endbibitem

\bibitem[\protect\citeauthoryear{Liu et~al.}{2019}]{liu2019graph}
\begin{bchapter}
\bauthor{\bsnm{Liu}, \binits{X.}},
\bauthor{\bsnm{Gao}, \binits{F.}},
\bauthor{\bsnm{Zhang}, \binits{Q.}},
\bauthor{\bsnm{Zhao}, \binits{H.}}:
\bctitle{Graph convolution for multimodal information extraction from visually rich documents}.
In: \bbtitle{Proceedings of the 2019 Conference of the North American Chapter of the Association for Computational Linguistics: Human Language Technologies, Volume 2 (Industry Papers)},
pp. \bfpage{32}--\blpage{39}
(\byear{2019})
\end{bchapter}
\endbibitem

\bibitem[\protect\citeauthoryear{Cheng et~al.}{2020}]{cheng2020one}
\begin{bchapter}
\bauthor{\bsnm{Cheng}, \binits{M.}},
\bauthor{\bsnm{Qiu}, \binits{M.}},
\bauthor{\bsnm{Shi}, \binits{X.}},
\bauthor{\bsnm{Huang}, \binits{J.}},
\bauthor{\bsnm{Lin}, \binits{W.}}:
\bctitle{One-shot text field labeling using attention and belief propagation for structure information extraction}.
In: \bbtitle{Proceedings of the 28th ACM International Conference on Multimedia},
pp. \bfpage{340}--\blpage{348}
(\byear{2020})
\end{bchapter}
\endbibitem

\bibitem[\protect\citeauthoryear{Wang et~al.}{2023}]{queryform}
\begin{bchapter}
\bauthor{\bsnm{Wang}, \binits{Z.}},
\bauthor{\bsnm{Zhang}, \binits{Z.}},
\bauthor{\bsnm{Devlin}, \binits{J.}},
\bauthor{\bsnm{Lee}, \binits{C.-Y.}},
\bauthor{\bsnm{Su}, \binits{G.}},
\bauthor{\bsnm{Zhang}, \binits{H.}},
\bauthor{\bsnm{Dy}, \binits{J.}},
\bauthor{\bsnm{Perot}, \binits{V.}},
\bauthor{\bsnm{Pfister}, \binits{T.}}:
\bctitle{Queryform: A simple zero-shot form entity query framework}.
In: \bbtitle{Findings of the Association for Computational Linguistics: ACL 2023},
pp. \bfpage{4146}--\blpage{4159}
(\byear{2023})
\end{bchapter}
\endbibitem

\bibitem[\protect\citeauthoryear{Perot et~al.}{2023}]{lmdx}
\begin{botherref}
\oauthor{\bsnm{Perot}, \binits{V.}},
\oauthor{\bsnm{Kang}, \binits{K.}},
\oauthor{\bsnm{Luisier}, \binits{F.}},
\oauthor{\bsnm{Su}, \binits{G.}},
\oauthor{\bsnm{Sun}, \binits{X.}},
\oauthor{\bsnm{Boppana}, \binits{R.S.}},
\oauthor{\bsnm{Wang}, \binits{Z.}},
\oauthor{\bsnm{Mu}, \binits{J.}},
\oauthor{\bsnm{Zhang}, \binits{H.}},
\oauthor{\bsnm{Hua}, \binits{N.}}:
Lmdx: Language model-based document information extraction and localization.
arXiv preprint arXiv:2309.10952
(2023)
\end{botherref}
\endbibitem

\bibitem[\protect\citeauthoryear{Hwang et~al.}{2021}]{hwang2021spatial}
\begin{bchapter}
\bauthor{\bsnm{Hwang}, \binits{W.}},
\bauthor{\bsnm{Yim}, \binits{J.}},
\bauthor{\bsnm{Park}, \binits{S.}},
\bauthor{\bsnm{Yang}, \binits{S.}},
\bauthor{\bsnm{Seo}, \binits{M.}}:
\bctitle{Spatial dependency parsing for semi-structured document information extraction}.
In: \bbtitle{Findings of the Association for Computational Linguistics: ACL-IJCNLP 2021},
pp. \bfpage{330}--\blpage{343}
(\byear{2021})
\end{bchapter}
\endbibitem

\bibitem[\protect\citeauthoryear{Liao et~al.}{2023}]{doctr}
\begin{bchapter}
\bauthor{\bsnm{Liao}, \binits{H.}},
\bauthor{\bsnm{RoyChowdhury}, \binits{A.}},
\bauthor{\bsnm{Li}, \binits{W.}},
\bauthor{\bsnm{Bansal}, \binits{A.}},
\bauthor{\bsnm{Zhang}, \binits{Y.}},
\bauthor{\bsnm{Tu}, \binits{Z.}},
\bauthor{\bsnm{Satzoda}, \binits{R.K.}},
\bauthor{\bsnm{Manmatha}, \binits{R.}},
\bauthor{\bsnm{Mahadevan}, \binits{V.}}:
\bctitle{Doctr: Document transformer for structured information extraction in documents}.
In: \bbtitle{Proceedings of the IEEE/CVF International Conference on Computer Vision},
pp. \bfpage{19584}--\blpage{19594}
(\byear{2023})
\end{bchapter}
\endbibitem

\bibitem[\protect\citeauthoryear{Tito et~al.}{2023}]{mpdocvqa}
\begin{barticle}
\bauthor{\bsnm{Tito}, \binits{R.}},
\bauthor{\bsnm{Karatzas}, \binits{D.}},
\bauthor{\bsnm{Valveny}, \binits{E.}}:
\batitle{Hierarchical multimodal transformers for multipage docvqa}.
\bjtitle{Pattern Recognition}
\bvolume{144},
\bfpage{109834}
(\byear{2023})
\end{barticle}
\endbibitem

\bibitem[\protect\citeauthoryear{Blau et~al.}{2024}]{gram}
\begin{bchapter}
\bauthor{\bsnm{Blau}, \binits{T.}},
\bauthor{\bsnm{Fogel}, \binits{S.}},
\bauthor{\bsnm{Ronen}, \binits{R.}},
\bauthor{\bsnm{Golts}, \binits{A.}},
\bauthor{\bsnm{Ganz}, \binits{R.}},
\bauthor{\bsnm{Ben~Avraham}, \binits{E.}},
\bauthor{\bsnm{Aberdam}, \binits{A.}},
\bauthor{\bsnm{Tsiper}, \binits{S.}},
\bauthor{\bsnm{Litman}, \binits{R.}}:
\bctitle{Gram: Global reasoning for multi-page vqa}.
In: \bbtitle{Proceedings of the IEEE/CVF Conference on Computer Vision and Pattern Recognition},
pp. \bfpage{15598}--\blpage{15607}
(\byear{2024})
\end{bchapter}
\endbibitem

\bibitem[\protect\citeauthoryear{Kang et~al.}{2024}]{kang2024multi}
\begin{botherref}
\oauthor{\bsnm{Kang}, \binits{L.}},
\oauthor{\bsnm{Tito}, \binits{R.}},
\oauthor{\bsnm{Valveny}, \binits{E.}},
\oauthor{\bsnm{Karatzas}, \binits{D.}}:
Multi-page document visual question answering using self-attention scoring mechanism.
arXiv preprint arXiv:2404.19024
(2024)
\end{botherref}
\endbibitem

\bibitem[\protect\citeauthoryear{Lan et~al.}{2019}]{albert}
\begin{botherref}
\oauthor{\bsnm{Lan}, \binits{Z.}},
\oauthor{\bsnm{Chen}, \binits{M.}},
\oauthor{\bsnm{Goodman}, \binits{S.}},
\oauthor{\bsnm{Gimpel}, \binits{K.}},
\oauthor{\bsnm{Sharma}, \binits{P.}},
\oauthor{\bsnm{Soricut}, \binits{R.}}:
Albert: A lite bert for self-supervised learning of language representations.
arXiv preprint arXiv:1909.11942
(2019)
\end{botherref}
\endbibitem

\bibitem[\protect\citeauthoryear{Hong et~al.}{2021}]{bros}
\begin{botherref}
\oauthor{\bsnm{Hong}, \binits{T.}},
\oauthor{\bsnm{Kim}, \binits{D.}},
\oauthor{\bsnm{Ji}, \binits{M.}},
\oauthor{\bsnm{Hwang}, \binits{W.}},
\oauthor{\bsnm{Nam}, \binits{D.}},
\oauthor{\bsnm{Park}, \binits{S.}}:
Bros: A pre-trained language model for understanding texts in document
(2021)
\end{botherref}
\endbibitem

\bibitem[\protect\citeauthoryear{Xu et~al.}{2021}]{layoutxlm}
\begin{botherref}
\oauthor{\bsnm{Xu}, \binits{Y.}},
\oauthor{\bsnm{Lv}, \binits{T.}},
\oauthor{\bsnm{Cui}, \binits{L.}},
\oauthor{\bsnm{Wang}, \binits{G.}},
\oauthor{\bsnm{Lu}, \binits{Y.}},
\oauthor{\bsnm{Florencio}, \binits{D.}},
\oauthor{\bsnm{Zhang}, \binits{C.}},
\oauthor{\bsnm{Wei}, \binits{F.}}:
Layoutxlm: Multimodal pre-training for multilingual visually-rich document understanding.
arXiv preprint arXiv:2104.08836
(2021)
\end{botherref}
\endbibitem

\bibitem[\protect\citeauthoryear{Lample et~al.}{2016}]{lample2016neural}
\begin{bchapter}
\bauthor{\bsnm{Lample}, \binits{G.}},
\bauthor{\bsnm{Ballesteros}, \binits{M.}},
\bauthor{\bsnm{Subramanian}, \binits{S.}},
\bauthor{\bsnm{Kawakami}, \binits{K.}},
\bauthor{\bsnm{Dyer}, \binits{C.}}:
\bctitle{Neural architectures for named entity recognition}.
In: \bbtitle{Proceedings of the 2016 Conference of the North American Chapter of the Association for Computational Linguistics: Human Language Technologies},
pp. \bfpage{260}--\blpage{270}
(\byear{2016})
\end{bchapter}
\endbibitem

\bibitem[\protect\citeauthoryear{Sohn}{2016}]{sohn2016improved}
\begin{botherref}
\oauthor{\bsnm{Sohn}, \binits{K.}}:
Improved deep metric learning with multi-class n-pair loss objective.
Advances in neural information processing systems
\textbf{29}
(2016)
\end{botherref}
\endbibitem

\bibitem[\protect\citeauthoryear{Wang et~al.}{2021}]{layoutreader}
\begin{bchapter}
\bauthor{\bsnm{Wang}, \binits{Z.}},
\bauthor{\bsnm{Xu}, \binits{Y.}},
\bauthor{\bsnm{Cui}, \binits{L.}},
\bauthor{\bsnm{Shang}, \binits{J.}},
\bauthor{\bsnm{Wei}, \binits{F.}}:
\bctitle{Layoutreader: Pre-training of text and layout for reading order detection}.
In: \bbtitle{Proceedings of the 2021 Conference on Empirical Methods in Natural Language Processing},
pp. \bfpage{4735}--\blpage{4744}
(\byear{2021})
\end{bchapter}
\endbibitem

\bibitem[\protect\citeauthoryear{Snell et~al.}{2017}]{snell2017prototypical}
\begin{botherref}
\oauthor{\bsnm{Snell}, \binits{J.}},
\oauthor{\bsnm{Swersky}, \binits{K.}},
\oauthor{\bsnm{Zemel}, \binits{R.}}:
Prototypical networks for few-shot learning.
Advances in neural information processing systems
\textbf{30}
(2017)
\end{botherref}
\endbibitem

\bibitem[\protect\citeauthoryear{Oreshkin et~al.}{2018}]{oreshkin2018tadam}
\begin{botherref}
\oauthor{\bsnm{Oreshkin}, \binits{B.}},
\oauthor{\bsnm{Rodr{\'\i}guez~L{\'o}pez}, \binits{P.}},
\oauthor{\bsnm{Lacoste}, \binits{A.}}:
Tadam: Task dependent adaptive metric for improved few-shot learning.
Advances in neural information processing systems
\textbf{31}
(2018)
\end{botherref}
\endbibitem

\bibitem[\protect\citeauthoryear{OpenAI}{2023}]{chatgpt}
\begin{botherref}
\oauthor{\bsnm{OpenAI}}:
ChatGPT: A conversational agent
(2023).
\url{https://www.openai.com/chatgpt}
\end{botherref}
\endbibitem

\bibitem[\protect\citeauthoryear{Touvron et~al.}{2023}]{llama}
\begin{botherref}
\oauthor{\bsnm{Touvron}, \binits{H.}},
\oauthor{\bsnm{Lavril}, \binits{T.}},
\oauthor{\bsnm{Izacard}, \binits{G.}},
\oauthor{\bsnm{Martinet}, \binits{X.}},
\oauthor{\bsnm{Lachaux}, \binits{M.-A.}},
\oauthor{\bsnm{Lacroix}, \binits{T.}},
\oauthor{\bsnm{Rozi{\`e}re}, \binits{B.}},
\oauthor{\bsnm{Goyal}, \binits{N.}},
\oauthor{\bsnm{Hambro}, \binits{E.}},
\oauthor{\bsnm{Azhar}, \binits{F.}}, et al.:
Llama: Open and efficient foundation language models.
arXiv preprint arXiv:2302.13971
(2023)
\end{botherref}
\endbibitem

\bibitem[\protect\citeauthoryear{Liu et~al.}{2024}]{llava}
\begin{botherref}
\oauthor{\bsnm{Liu}, \binits{H.}},
\oauthor{\bsnm{Li}, \binits{C.}},
\oauthor{\bsnm{Wu}, \binits{Q.}},
\oauthor{\bsnm{Lee}, \binits{Y.J.}}:
Visual instruction tuning.
Advances in neural information processing systems
\textbf{36}
(2024)
\end{botherref}
\endbibitem

\bibitem[\protect\citeauthoryear{Brown et~al.}{2020}]{gpt3}
\begin{barticle}
\bauthor{\bsnm{Brown}, \binits{T.}},
\bauthor{\bsnm{Mann}, \binits{B.}},
\bauthor{\bsnm{Ryder}, \binits{N.}},
\bauthor{\bsnm{Subbiah}, \binits{M.}},
\bauthor{\bsnm{Kaplan}, \binits{J.D.}},
\bauthor{\bsnm{Dhariwal}, \binits{P.}},
\bauthor{\bsnm{Neelakantan}, \binits{A.}},
\bauthor{\bsnm{Shyam}, \binits{P.}},
\bauthor{\bsnm{Sastry}, \binits{G.}},
\bauthor{\bsnm{Askell}, \binits{A.}}, \betal:
\batitle{Language models are few-shot learners}.
\bjtitle{Advances in neural information processing systems}
\bvolume{33},
\bfpage{1877}--\blpage{1901}
(\byear{2020})
\end{barticle}
\endbibitem

\bibitem[\protect\citeauthoryear{He et~al.}{2016}]{resnet}
\begin{bchapter}
\bauthor{\bsnm{He}, \binits{K.}},
\bauthor{\bsnm{Zhang}, \binits{X.}},
\bauthor{\bsnm{Ren}, \binits{S.}},
\bauthor{\bsnm{Sun}, \binits{J.}}:
\bctitle{Deep residual learning for image recognition}.
In: \bbtitle{Proceedings of the IEEE Conference on Computer Vision and Pattern Recognition},
pp. \bfpage{770}--\blpage{778}
(\byear{2016})
\end{bchapter}
\endbibitem

\bibitem[\protect\citeauthoryear{Wang et~al.}{2022}]{wang2022ofa}
\begin{bchapter}
\bauthor{\bsnm{Wang}, \binits{P.}},
\bauthor{\bsnm{Yang}, \binits{A.}},
\bauthor{\bsnm{Men}, \binits{R.}},
\bauthor{\bsnm{Lin}, \binits{J.}},
\bauthor{\bsnm{Bai}, \binits{S.}},
\bauthor{\bsnm{Li}, \binits{Z.}},
\bauthor{\bsnm{Ma}, \binits{J.}},
\bauthor{\bsnm{Zhou}, \binits{C.}},
\bauthor{\bsnm{Zhou}, \binits{J.}},
\bauthor{\bsnm{Yang}, \binits{H.}}:
\bctitle{Ofa: Unifying architectures, tasks, and modalities through a simple sequence-to-sequence learning framework}.
In: \bbtitle{International Conference on Machine Learning},
pp. \bfpage{23318}--\blpage{23340}
(\byear{2022}).
\bcomment{PMLR}
\end{bchapter}
\endbibitem

\bibitem[\protect\citeauthoryear{Reimers and Gurevych}{2019}]{sentencebert}
\begin{bchapter}
\bauthor{\bsnm{Reimers}, \binits{N.}},
\bauthor{\bsnm{Gurevych}, \binits{I.}}:
\bctitle{Sentence-bert: Sentence embeddings using siamese bert-networks}.
In: \bbtitle{Proceedings of the 2019 Conference on Empirical Methods in Natural Language Processing and the 9th International Joint Conference on Natural Language Processing (EMNLP-IJCNLP)},
pp. \bfpage{3982}--\blpage{3992}
(\byear{2019})
\end{bchapter}
\endbibitem

\bibitem[\protect\citeauthoryear{Wang et~al.}{2019}]{wang2019words}
\begin{bchapter}
\bauthor{\bsnm{Wang}, \binits{Y.}},
\bauthor{\bsnm{Shen}, \binits{Y.}},
\bauthor{\bsnm{Liu}, \binits{Z.}},
\bauthor{\bsnm{Liang}, \binits{P.P.}},
\bauthor{\bsnm{Zadeh}, \binits{A.}},
\bauthor{\bsnm{Morency}, \binits{L.-P.}}:
\bctitle{Words can shift: Dynamically adjusting word representations using nonverbal behaviors}.
In: \bbtitle{Proceedings of the AAAI Conference on Artificial Intelligence},
vol. \bseriesno{33},
pp. \bfpage{7216}--\blpage{7223}
(\byear{2019})
\end{bchapter}
\endbibitem

\bibitem[\protect\citeauthoryear{Mikolov et~al.}{2013}]{mikolov2013distributed}
\begin{botherref}
\oauthor{\bsnm{Mikolov}, \binits{T.}},
\oauthor{\bsnm{Sutskever}, \binits{I.}},
\oauthor{\bsnm{Chen}, \binits{K.}},
\oauthor{\bsnm{Corrado}, \binits{G.S.}},
\oauthor{\bsnm{Dean}, \binits{J.}}:
Distributed representations of words and phrases and their compositionality.
Advances in neural information processing systems
\textbf{26}
(2013)
\end{botherref}
\endbibitem

\bibitem[\protect\citeauthoryear{Carion et~al.}{2020}]{carion2020end}
\begin{bchapter}
\bauthor{\bsnm{Carion}, \binits{N.}},
\bauthor{\bsnm{Massa}, \binits{F.}},
\bauthor{\bsnm{Synnaeve}, \binits{G.}},
\bauthor{\bsnm{Usunier}, \binits{N.}},
\bauthor{\bsnm{Kirillov}, \binits{A.}},
\bauthor{\bsnm{Zagoruyko}, \binits{S.}}:
\bctitle{End-to-end object detection with transformers}.
In: \bbtitle{European Conference on Computer Vision},
pp. \bfpage{213}--\blpage{229}
(\byear{2020}).
\bcomment{Springer}
\end{bchapter}
\endbibitem

\bibitem[\protect\citeauthoryear{Ding et~al.}{2022}]{vdoc}
\begin{bchapter}
\bauthor{\bsnm{Ding}, \binits{Y.}},
\bauthor{\bsnm{Huang}, \binits{Z.}},
\bauthor{\bsnm{Wang}, \binits{R.}},
\bauthor{\bsnm{Zhang}, \binits{Y.}},
\bauthor{\bsnm{Chen}, \binits{X.}},
\bauthor{\bsnm{Ma}, \binits{Y.}},
\bauthor{\bsnm{Chung}, \binits{H.}},
\bauthor{\bsnm{Han}, \binits{S.C.}}:
\bctitle{V-doc: Visual questions answers with documents}.
In: \bbtitle{Proceedings of the IEEE/CVF Conference on Computer Vision and Pattern Recognition},
pp. \bfpage{21492}--\blpage{21498}
(\byear{2022})
\end{bchapter}
\endbibitem

\bibitem[\protect\citeauthoryear{Li et~al.}{2019}]{visualbert}
\begin{botherref}
\oauthor{\bsnm{Li}, \binits{L.H.}},
\oauthor{\bsnm{Yatskar}, \binits{M.}},
\oauthor{\bsnm{Yin}, \binits{D.}},
\oauthor{\bsnm{Hsieh}, \binits{C.-J.}},
\oauthor{\bsnm{Chang}, \binits{K.-W.}}:
Visualbert: A simple and performant baseline for vision and language.
arXiv preprint arXiv:1908.03557
(2019)
\end{botherref}
\endbibitem

\bibitem[\protect\citeauthoryear{Tan and Bansal}{2019}]{lxmert}
\begin{bchapter}
\bauthor{\bsnm{Tan}, \binits{H.}},
\bauthor{\bsnm{Bansal}, \binits{M.}}:
\bctitle{Lxmert: Learning cross-modality encoder representations from transformers}.
In: \bbtitle{Proceedings of the 2019 Conference on Empirical Methods in Natural Language Processing and the 9th International Joint Conference on Natural Language Processing (EMNLP-IJCNLP)},
pp. \bfpage{5100}--\blpage{5111}
(\byear{2019})
\end{bchapter}
\endbibitem

\bibitem[\protect\citeauthoryear{Kim et~al.}{2021}]{vilt}
\begin{bchapter}
\bauthor{\bsnm{Kim}, \binits{W.}},
\bauthor{\bsnm{Son}, \binits{B.}},
\bauthor{\bsnm{Kim}, \binits{I.}}:
\bctitle{Vilt: Vision-and-language transformer without convolution or region supervision}.
In: \bbtitle{International Conference on Machine Learning},
pp. \bfpage{5583}--\blpage{5594}
(\byear{2021}).
\bcomment{PMLR}
\end{bchapter}
\endbibitem

\bibitem[\protect\citeauthoryear{Raffel et~al.}{2020}]{t5}
\begin{barticle}
\bauthor{\bsnm{Raffel}, \binits{C.}},
\bauthor{\bsnm{Shazeer}, \binits{N.}},
\bauthor{\bsnm{Roberts}, \binits{A.}},
\bauthor{\bsnm{Lee}, \binits{K.}},
\bauthor{\bsnm{Narang}, \binits{S.}},
\bauthor{\bsnm{Matena}, \binits{M.}},
\bauthor{\bsnm{Zhou}, \binits{Y.}},
\bauthor{\bsnm{Li}, \binits{W.}},
\bauthor{\bsnm{Liu}, \binits{P.J.}}:
\batitle{Exploring the limits of transfer learning with a unified text-to-text transformer}.
\bjtitle{Journal of machine learning research}
\bvolume{21}(\bissue{140}),
\bfpage{1}--\blpage{67}
(\byear{2020})
\end{barticle}
\endbibitem

\bibitem[\protect\citeauthoryear{Biten et~al.}{2022}]{biten2022latr}
\begin{bchapter}
\bauthor{\bsnm{Biten}, \binits{A.F.}},
\bauthor{\bsnm{Litman}, \binits{R.}},
\bauthor{\bsnm{Xie}, \binits{Y.}},
\bauthor{\bsnm{Appalaraju}, \binits{S.}},
\bauthor{\bsnm{Manmatha}, \binits{R.}}:
\bctitle{Latr: Layout-aware transformer for scene-text vqa}.
In: \bbtitle{Proceedings of the IEEE/CVF Conference on Computer Vision and Pattern Recognition},
pp. \bfpage{16548}--\blpage{16558}
(\byear{2022})
\end{bchapter}
\endbibitem

\bibitem[\protect\citeauthoryear{Li et~al.}{2022}]{dit}
\begin{bchapter}
\bauthor{\bsnm{Li}, \binits{J.}},
\bauthor{\bsnm{Xu}, \binits{Y.}},
\bauthor{\bsnm{Lv}, \binits{T.}},
\bauthor{\bsnm{Cui}, \binits{L.}},
\bauthor{\bsnm{Zhang}, \binits{C.}},
\bauthor{\bsnm{Wei}, \binits{F.}}:
\bctitle{Dit: Self-supervised pre-training for document image transformer}.
In: \bbtitle{Proceedings of the 30th ACM International Conference on Multimedia},
pp. \bfpage{3530}--\blpage{3539}
(\byear{2022})
\end{bchapter}
\endbibitem

\bibitem[\protect\citeauthoryear{Appalaraju et~al.}{2024}]{docformerv2}
\begin{bchapter}
\bauthor{\bsnm{Appalaraju}, \binits{S.}},
\bauthor{\bsnm{Tang}, \binits{P.}},
\bauthor{\bsnm{Dong}, \binits{Q.}},
\bauthor{\bsnm{Sankaran}, \binits{N.}},
\bauthor{\bsnm{Zhou}, \binits{Y.}},
\bauthor{\bsnm{Manmatha}, \binits{R.}}:
\bctitle{Docformerv2: Local features for document understanding}.
In: \bbtitle{Proceedings of the AAAI Conference on Artificial Intelligence},
vol. \bseriesno{38},
pp. \bfpage{709}--\blpage{718}
(\byear{2024})
\end{bchapter}
\endbibitem

\bibitem[\protect\citeauthoryear{Press et~al.}{2021}]{press2021train}
\begin{bchapter}
\bauthor{\bsnm{Press}, \binits{O.}},
\bauthor{\bsnm{Smith}, \binits{N.}},
\bauthor{\bsnm{Lewis}, \binits{M.}}:
\bctitle{Train short, test long: Attention with linear biases enables input length extrapolation}.
In: \bbtitle{International Conference on Learning Representations}
(\byear{2021})
\end{bchapter}
\endbibitem

\bibitem[\protect\citeauthoryear{Lee et~al.}{2022}]{pix2struct}
\begin{botherref}
\oauthor{\bsnm{Lee}, \binits{K.}},
\oauthor{\bsnm{Joshi}, \binits{M.}},
\oauthor{\bsnm{Turc}, \binits{I.}},
\oauthor{\bsnm{Hu}, \binits{H.}},
\oauthor{\bsnm{Liu}, \binits{F.}},
\oauthor{\bsnm{Eisenschlos}, \binits{J.}},
\oauthor{\bsnm{Khandelwal}, \binits{U.}},
\oauthor{\bsnm{Shaw}, \binits{P.}},
\oauthor{\bsnm{Chang}, \binits{M.-W.}},
\oauthor{\bsnm{Toutanova}, \binits{K.}}:
Pix2struct: Screenshot parsing as pretraining for visual language understanding.
arXiv preprint arXiv:2210.03347
(2022)
\end{botherref}
\endbibitem

\bibitem[\protect\citeauthoryear{Chen et~al.}{2022}]{xdoc}
\begin{botherref}
\oauthor{\bsnm{Chen}, \binits{J.}},
\oauthor{\bsnm{Lv}, \binits{T.}},
\oauthor{\bsnm{Cui}, \binits{L.}},
\oauthor{\bsnm{Zhang}, \binits{C.}},
\oauthor{\bsnm{Wei}, \binits{F.}}:
Xdoc: Unified pre-training for cross-format document understanding.
arXiv preprint arXiv:2210.02849
(2022)
\end{botherref}
\endbibitem

\bibitem[\protect\citeauthoryear{Appalaraju et~al.}{2021}]{docformer}
\begin{bchapter}
\bauthor{\bsnm{Appalaraju}, \binits{S.}},
\bauthor{\bsnm{Jasani}, \binits{B.}},
\bauthor{\bsnm{Kota}, \binits{B.U.}},
\bauthor{\bsnm{Xie}, \binits{Y.}},
\bauthor{\bsnm{Manmatha}, \binits{R.}}:
\bctitle{Docformer: End-to-end transformer for document understanding}.
In: \bbtitle{Proceedings of the IEEE/CVF International Conference on Computer Vision},
pp. \bfpage{993}--\blpage{1003}
(\byear{2021})
\end{bchapter}
\endbibitem

\bibitem[\protect\citeauthoryear{Zhai et~al.}{2023}]{faststructext}
\begin{bchapter}
\bauthor{\bsnm{Zhai}, \binits{M.}},
\bauthor{\bsnm{Li}, \binits{Y.}},
\bauthor{\bsnm{Qin}, \binits{X.}},
\bauthor{\bsnm{Yi}, \binits{C.}},
\bauthor{\bsnm{Xie}, \binits{Q.}},
\bauthor{\bsnm{Zhang}, \binits{C.}},
\bauthor{\bsnm{Yao}, \binits{K.}},
\bauthor{\bsnm{Wu}, \binits{Y.}},
\bauthor{\bsnm{Jia}, \binits{Y.}}:
\bctitle{Fast-structext: an efficient hourglass transformer with modality-guided dynamic token merge for document understanding}.
In: \bbtitle{Proceedings of the Thirty-Second International Joint Conference on Artificial Intelligence},
pp. \bfpage{5269}--\blpage{5277}
(\byear{2023})
\end{bchapter}
\endbibitem

\bibitem[\protect\citeauthoryear{Wang et~al.}{2022}]{mgdoc}
\begin{bchapter}
\bauthor{\bsnm{Wang}, \binits{Z.}},
\bauthor{\bsnm{Gu}, \binits{J.}},
\bauthor{\bsnm{Tensmeyer}, \binits{C.}},
\bauthor{\bsnm{Barmpalios}, \binits{N.}},
\bauthor{\bsnm{Nenkova}, \binits{A.}},
\bauthor{\bsnm{Sun}, \binits{T.}},
\bauthor{\bsnm{Shang}, \binits{J.}},
\bauthor{\bsnm{Morariu}, \binits{V.}}:
\bctitle{Mgdoc: Pre-training with multi-granular hierarchy for document image understanding}.
In: \bbtitle{Proceedings of the 2022 Conference on Empirical Methods in Natural Language Processing},
pp. \bfpage{3984}--\blpage{3993}
(\byear{2022})
\end{bchapter}
\endbibitem

\bibitem[\protect\citeauthoryear{Luo et~al.}{2023}]{geolayoutlm}
\begin{bchapter}
\bauthor{\bsnm{Luo}, \binits{C.}},
\bauthor{\bsnm{Cheng}, \binits{C.}},
\bauthor{\bsnm{Zheng}, \binits{Q.}},
\bauthor{\bsnm{Yao}, \binits{C.}}:
\bctitle{Geolayoutlm: Geometric pre-training for visual information extraction}.
In: \bbtitle{Proceedings of the IEEE/CVF Conference on Computer Vision and Pattern Recognition},
pp. \bfpage{7092}--\blpage{7101}
(\byear{2023})
\end{bchapter}
\endbibitem

\bibitem[\protect\citeauthoryear{Mao et~al.}{2024}]{vitlp}
\begin{bchapter}
\bauthor{\bsnm{Mao}, \binits{Z.}},
\bauthor{\bsnm{Bai}, \binits{H.}},
\bauthor{\bsnm{Hou}, \binits{L.}},
\bauthor{\bsnm{Shang}, \binits{L.}},
\bauthor{\bsnm{Jiang}, \binits{X.}},
\bauthor{\bsnm{Liu}, \binits{Q.}},
\bauthor{\bsnm{Wong}, \binits{K.-F.}}:
\bctitle{Visually guided generative text-layout pre-training for document intelligence}.
In: \bbtitle{Proceedings of the 2024 Conference of the North American Chapter of the Association for Computational Linguistics: Human Language Technologies (Volume 1: Long Papers)},
pp. \bfpage{4713}--\blpage{4730}
(\byear{2024})
\end{bchapter}
\endbibitem

\bibitem[\protect\citeauthoryear{Cao et~al.}{2023}]{serum}
\begin{bchapter}
\bauthor{\bsnm{Cao}, \binits{H.}},
\bauthor{\bsnm{Bao}, \binits{C.}},
\bauthor{\bsnm{Liu}, \binits{C.}},
\bauthor{\bsnm{Chen}, \binits{H.}},
\bauthor{\bsnm{Yin}, \binits{K.}},
\bauthor{\bsnm{Liu}, \binits{H.}},
\bauthor{\bsnm{Liu}, \binits{Y.}},
\bauthor{\bsnm{Jiang}, \binits{D.}},
\bauthor{\bsnm{Sun}, \binits{X.}}:
\bctitle{Attention where it matters: Rethinking visual document understanding with selective region concentration}.
In: \bbtitle{Proceedings of the IEEE/CVF International Conference on Computer Vision},
pp. \bfpage{19517}--\blpage{19527}
(\byear{2023})
\end{bchapter}
\endbibitem

\bibitem[\protect\citeauthoryear{Du et~al.}{2022}]{calm}
\begin{bchapter}
\bauthor{\bsnm{Du}, \binits{Q.}},
\bauthor{\bsnm{Wang}, \binits{Q.}},
\bauthor{\bsnm{Li}, \binits{K.}},
\bauthor{\bsnm{Tian}, \binits{J.}},
\bauthor{\bsnm{Xiao}, \binits{L.}},
\bauthor{\bsnm{Jin}, \binits{Y.}}:
\bctitle{Calm: commen-sense knowledge augmentation for document image understanding}.
In: \bbtitle{Proceedings of the 30th ACM International Conference on Multimedia},
pp. \bfpage{3282}--\blpage{3290}
(\byear{2022})
\end{bchapter}
\endbibitem

\bibitem[\protect\citeauthoryear{Shi et~al.}{2023}]{layoutgcn}
\begin{bchapter}
\bauthor{\bsnm{Shi}, \binits{D.}},
\bauthor{\bsnm{Liu}, \binits{S.}},
\bauthor{\bsnm{Du}, \binits{J.}},
\bauthor{\bsnm{Zhu}, \binits{H.}}:
\bctitle{Layoutgcn: A lightweight architecture for visually rich document understanding}.
In: \bbtitle{International Conference on Document Analysis and Recognition},
pp. \bfpage{149}--\blpage{165}
(\byear{2023}).
\bcomment{Springer}
\end{bchapter}
\endbibitem

\bibitem[\protect\citeauthoryear{Joshi et~al.}{2020}]{joshi2020spanbert}
\begin{barticle}
\bauthor{\bsnm{Joshi}, \binits{M.}},
\bauthor{\bsnm{Chen}, \binits{D.}},
\bauthor{\bsnm{Liu}, \binits{Y.}},
\bauthor{\bsnm{Weld}, \binits{D.S.}},
\bauthor{\bsnm{Zettlemoyer}, \binits{L.}},
\bauthor{\bsnm{Levy}, \binits{O.}}:
\batitle{Spanbert: Improving pre-training by representing and predicting spans}.
\bjtitle{Transactions of the association for computational linguistics}
\bvolume{8},
\bfpage{64}--\blpage{77}
(\byear{2020})
\end{barticle}
\endbibitem

\bibitem[\protect\citeauthoryear{Li et~al.}{2021}]{structurallm}
\begin{bchapter}
\bauthor{\bsnm{Li}, \binits{C.}},
\bauthor{\bsnm{Bi}, \binits{B.}},
\bauthor{\bsnm{Yan}, \binits{M.}},
\bauthor{\bsnm{Wang}, \binits{W.}},
\bauthor{\bsnm{Huang}, \binits{S.}},
\bauthor{\bsnm{Huang}, \binits{F.}},
\bauthor{\bsnm{Si}, \binits{L.}}:
\bctitle{Structurallm: Structural pre-training for form understanding}.
In: \bbtitle{Proceedings of the 59th Annual Meeting of the Association for Computational Linguistics and the 11th International Joint Conference on Natural Language Processing (Volume 1: Long Papers)},
pp. \bfpage{6309}--\blpage{6318}
(\byear{2021})
\end{bchapter}
\endbibitem

\bibitem[\protect\citeauthoryear{Bao et~al.}{2021}]{bao2021beit}
\begin{bchapter}
\bauthor{\bsnm{Bao}, \binits{H.}},
\bauthor{\bsnm{Dong}, \binits{L.}},
\bauthor{\bsnm{Piao}, \binits{S.}},
\bauthor{\bsnm{Wei}, \binits{F.}}:
\bctitle{Beit: Bert pre-training of image transformers}.
In: \bbtitle{International Conference on Learning Representations}
(\byear{2021})
\end{bchapter}
\endbibitem

\bibitem[\protect\citeauthoryear{Jegou et~al.}{2010}]{jegou2010product}
\begin{barticle}
\bauthor{\bsnm{Jegou}, \binits{H.}},
\bauthor{\bsnm{Douze}, \binits{M.}},
\bauthor{\bsnm{Schmid}, \binits{C.}}:
\batitle{Product quantization for nearest neighbor search}.
\bjtitle{IEEE transactions on pattern analysis and machine intelligence}
\bvolume{33}(\bissue{1}),
\bfpage{117}--\blpage{128}
(\byear{2010})
\end{barticle}
\endbibitem

\bibitem[\protect\citeauthoryear{Wei et~al.}{2020}]{wei2020robust}
\begin{bchapter}
\bauthor{\bsnm{Wei}, \binits{M.}},
\bauthor{\bsnm{He}, \binits{Y.}},
\bauthor{\bsnm{Zhang}, \binits{Q.}}:
\bctitle{Robust layout-aware ie for visually rich documents with pre-trained language models}.
In: \bbtitle{Proceedings of the 43rd International ACM SIGIR Conference on Research and Development in Information Retrieval},
pp. \bfpage{2367}--\blpage{2376}
(\byear{2020})
\end{bchapter}
\endbibitem

\bibitem[\protect\citeauthoryear{Yao et~al.}{2021}]{yao2021filip}
\begin{bchapter}
\bauthor{\bsnm{Yao}, \binits{L.}},
\bauthor{\bsnm{Huang}, \binits{R.}},
\bauthor{\bsnm{Hou}, \binits{L.}},
\bauthor{\bsnm{Lu}, \binits{G.}},
\bauthor{\bsnm{Niu}, \binits{M.}},
\bauthor{\bsnm{Xu}, \binits{H.}},
\bauthor{\bsnm{Liang}, \binits{X.}},
\bauthor{\bsnm{Li}, \binits{Z.}},
\bauthor{\bsnm{Jiang}, \binits{X.}},
\bauthor{\bsnm{Xu}, \binits{C.}}:
\bctitle{Filip: Fine-grained interactive language-image pre-training}.
In: \bbtitle{International Conference on Learning Representations}
(\byear{2021})
\end{bchapter}
\endbibitem

\bibitem[\protect\citeauthoryear{Powalski et~al.}{2021}]{tilt}
\begin{bchapter}
\bauthor{\bsnm{Powalski}, \binits{R.}},
\bauthor{\bsnm{Borchmann}, \binits{{\L}.}},
\bauthor{\bsnm{Jurkiewicz}, \binits{D.}},
\bauthor{\bsnm{Dwojak}, \binits{T.}},
\bauthor{\bsnm{Pietruszka}, \binits{M.}},
\bauthor{\bsnm{Pa{\l}ka}, \binits{G.}}:
\bctitle{Going full-tilt boogie on document understanding with text-image-layout transformer}.
In: \bbtitle{Document Analysis and Recognition--ICDAR 2021: 16th International Conference, Lausanne, Switzerland, September 5--10, 2021, Proceedings, Part II 16},
pp. \bfpage{732}--\blpage{747}
(\byear{2021}).
\bcomment{Springer}
\end{bchapter}
\endbibitem

\bibitem[\protect\citeauthoryear{Guu et~al.}{2020}]{guu2020retrieval}
\begin{bchapter}
\bauthor{\bsnm{Guu}, \binits{K.}},
\bauthor{\bsnm{Lee}, \binits{K.}},
\bauthor{\bsnm{Tung}, \binits{Z.}},
\bauthor{\bsnm{Pasupat}, \binits{P.}},
\bauthor{\bsnm{Chang}, \binits{M.}}:
\bctitle{Retrieval augmented language model pre-training}.
In: \bbtitle{International Conference on Machine Learning},
pp. \bfpage{3929}--\blpage{3938}
(\byear{2020}).
\bcomment{PMLR}
\end{bchapter}
\endbibitem

\bibitem[\protect\citeauthoryear{Stanis{\l}awek et~al.}{2021}]{nda}
\begin{bchapter}
\bauthor{\bsnm{Stanis{\l}awek}, \binits{T.}},
\bauthor{\bsnm{Grali{\'n}ski}, \binits{F.}},
\bauthor{\bsnm{Wr{\'o}blewska}, \binits{A.}},
\bauthor{\bsnm{Lipi{\'n}ski}, \binits{D.}},
\bauthor{\bsnm{Kaliska}, \binits{A.}},
\bauthor{\bsnm{Rosalska}, \binits{P.}},
\bauthor{\bsnm{Topolski}, \binits{B.}},
\bauthor{\bsnm{Biecek}, \binits{P.}}:
\bctitle{Kleister: key information extraction datasets involving long documents with complex layouts}.
In: \bbtitle{Document Analysis and Recognition--ICDAR 2021: 16th International Conference, Lausanne, Switzerland, September 5--10, 2021, Proceedings, Part I},
pp. \bfpage{564}--\blpage{579}
(\byear{2021}).
\bcomment{Springer}
\end{bchapter}
\endbibitem

\bibitem[\protect\citeauthoryear{Tanaka et~al.}{2021}]{visualmrc}
\begin{bchapter}
\bauthor{\bsnm{Tanaka}, \binits{R.}},
\bauthor{\bsnm{Nishida}, \binits{K.}},
\bauthor{\bsnm{Yoshida}, \binits{S.}}:
\bctitle{Visualmrc: Machine reading comprehension on document images}.
In: \bbtitle{Proceedings of the AAAI Conference on Artificial Intelligence},
vol. \bseriesno{35},
pp. \bfpage{13878}--\blpage{13888}
(\byear{2021})
\end{bchapter}
\endbibitem

\bibitem[\protect\citeauthoryear{Zhong et~al.}{2019}]{publaynet}
\begin{bchapter}
\bauthor{\bsnm{Zhong}, \binits{X.}},
\bauthor{\bsnm{Tang}, \binits{J.}},
\bauthor{\bsnm{Yepes}, \binits{A.J.}}:
\bctitle{Publaynet: largest dataset ever for document layout analysis}.
In: \bbtitle{2019 International Conference on Document Analysis and Recognition (ICDAR)},
pp. \bfpage{1015}--\blpage{1022}
(\byear{2019}).
\bcomment{IEEE}
\end{bchapter}
\endbibitem

\bibitem[\protect\citeauthoryear{Lewis et~al.}{2020}]{bart}
\begin{bchapter}
\bauthor{\bsnm{Lewis}, \binits{M.}},
\bauthor{\bsnm{Liu}, \binits{Y.}},
\bauthor{\bsnm{Goyal}, \binits{N.}},
\bauthor{\bsnm{Ghazvininejad}, \binits{M.}},
\bauthor{\bsnm{Mohamed}, \binits{A.}},
\bauthor{\bsnm{Levy}, \binits{O.}},
\bauthor{\bsnm{Stoyanov}, \binits{V.}},
\bauthor{\bsnm{Zettlemoyer}, \binits{L.}}:
\bctitle{Bart: Denoising sequence-to-sequence pre-training for natural language generation, translation, and comprehension}.
In: \bbtitle{Proceedings of the 58th Annual Meeting of the Association for Computational Linguistics},
pp. \bfpage{7871}--\blpage{7880}
(\byear{2020})
\end{bchapter}
\endbibitem

\bibitem[\protect\citeauthoryear{Lin et~al.}{2017}]{lin2017feature}
\begin{bchapter}
\bauthor{\bsnm{Lin}, \binits{T.-Y.}},
\bauthor{\bsnm{Doll{\'a}r}, \binits{P.}},
\bauthor{\bsnm{Girshick}, \binits{R.}},
\bauthor{\bsnm{He}, \binits{K.}},
\bauthor{\bsnm{Hariharan}, \binits{B.}},
\bauthor{\bsnm{Belongie}, \binits{S.}}:
\bctitle{Feature pyramid networks for object detection}.
In: \bbtitle{Proceedings of the IEEE Conference on Computer Vision and Pattern Recognition},
pp. \bfpage{2117}--\blpage{2125}
(\byear{2017})
\end{bchapter}
\endbibitem

\bibitem[\protect\citeauthoryear{Bai et~al.}{2023}]{qwen}
\begin{botherref}
\oauthor{\bsnm{Bai}, \binits{J.}},
\oauthor{\bsnm{Bai}, \binits{S.}},
\oauthor{\bsnm{Yang}, \binits{S.}},
\oauthor{\bsnm{Wang}, \binits{S.}},
\oauthor{\bsnm{Tan}, \binits{S.}},
\oauthor{\bsnm{Wang}, \binits{P.}},
\oauthor{\bsnm{Lin}, \binits{J.}},
\oauthor{\bsnm{Zhou}, \binits{C.}},
\oauthor{\bsnm{Zhou}, \binits{J.}}:
Qwen-VL: A Versatile Vision-Language Model for Understanding, Localization, Text Reading, and Beyond
(2023).
\url{https://arxiv.org/abs/2308.12966}
\end{botherref}
\endbibitem

\bibitem[\protect\citeauthoryear{Wang et~al.}{2024}]{docllm}
\begin{bchapter}
\bauthor{\bsnm{Wang}, \binits{D.}},
\bauthor{\bsnm{Raman}, \binits{N.}},
\bauthor{\bsnm{Sibue}, \binits{M.}},
\bauthor{\bsnm{Ma}, \binits{Z.}},
\bauthor{\bsnm{Babkin}, \binits{P.}},
\bauthor{\bsnm{Kaur}, \binits{S.}},
\bauthor{\bsnm{Pei}, \binits{Y.}},
\bauthor{\bsnm{Nourbakhsh}, \binits{A.}},
\bauthor{\bsnm{Liu}, \binits{X.}}:
\bctitle{Docllm: A layout-aware generative language model for multimodal document understanding}.
In: \bbtitle{Proceedings of the 62nd Annual Meeting of the Association for Computational Linguistics (Volume 1: Long Papers)},
pp. \bfpage{8529}--\blpage{8548}.
\bpublisher{Association for Computational Linguistics}, \blocation{???}
(\byear{2024}).
\burl{https://doi.org/10.18653/v1/2024.acl-long.463}
\end{bchapter}
\endbibitem

\bibitem[\protect\citeauthoryear{Lamott et~al.}{2024}]{lapdoc}
\begin{bchapter}
\bauthor{\bsnm{Lamott}, \binits{M.}},
\bauthor{\bsnm{Weweler}, \binits{Y.-N.}},
\bauthor{\bsnm{Ulges}, \binits{A.}},
\bauthor{\bsnm{Shafait}, \binits{F.}},
\bauthor{\bsnm{Krechel}, \binits{D.}},
\bauthor{\bsnm{Obradovic}, \binits{D.}}:
\bctitle{Lapdoc: Layout-aware prompting for documents}.
In: \bbtitle{International Conference on Document Analysis and Recognition},
pp. \bfpage{142}--\blpage{159}.
\bpublisher{Springer}, \blocation{???}
(\byear{2024}).
\burl{https://doi.org/10.1007/978-3-031-70546-5\_9}
\end{bchapter}
\endbibitem

\bibitem[\protect\citeauthoryear{Zheng et~al.}{2024}]{zheng2024judging}
\begin{botherref}
\oauthor{\bsnm{Zheng}, \binits{L.}},
\oauthor{\bsnm{Chiang}, \binits{W.-L.}},
\oauthor{\bsnm{Sheng}, \binits{Y.}},
\oauthor{\bsnm{Zhuang}, \binits{S.}},
\oauthor{\bsnm{Wu}, \binits{Z.}},
\oauthor{\bsnm{Zhuang}, \binits{Y.}},
\oauthor{\bsnm{Lin}, \binits{Z.}},
\oauthor{\bsnm{Li}, \binits{Z.}},
\oauthor{\bsnm{Li}, \binits{D.}},
\oauthor{\bsnm{Xing}, \binits{E.}}, et al.:
Judging llm-as-a-judge with mt-bench and chatbot arena.
Advances in Neural Information Processing Systems
\textbf{36}
(2024)
\end{botherref}
\endbibitem

\bibitem[\protect\citeauthoryear{Speer et~al.}{2017}]{speer2017conceptnet}
\begin{bchapter}
\bauthor{\bsnm{Speer}, \binits{R.}},
\bauthor{\bsnm{Chin}, \binits{J.}},
\bauthor{\bsnm{Havasi}, \binits{C.}}:
\bctitle{Conceptnet 5.5: An open multilingual graph of general knowledge}.
In: \bbtitle{Proceedings of the AAAI Conference on Artificial Intelligence},
vol. \bseriesno{31}
(\byear{2017})
\end{bchapter}
\endbibitem

\bibitem[\protect\citeauthoryear{Zhu et~al.}{2021}]{zhu2021mucko}
\begin{bchapter}
\bauthor{\bsnm{Zhu}, \binits{Z.}},
\bauthor{\bsnm{Yu}, \binits{J.}},
\bauthor{\bsnm{Wang}, \binits{Y.}},
\bauthor{\bsnm{Sun}, \binits{Y.}},
\bauthor{\bsnm{Hu}, \binits{Y.}},
\bauthor{\bsnm{Wu}, \binits{Q.}}:
\bctitle{Mucko: multi-layer cross-modal knowledge reasoning for fact-based visual question answering}.
In: \bbtitle{Proceedings of the Twenty-Ninth International Conference on International Joint Conferences on Artificial Intelligence},
pp. \bfpage{1097}--\blpage{1103}
(\byear{2021})
\end{bchapter}
\endbibitem

\bibitem[\protect\citeauthoryear{Kim}{2014}]{textcnn}
\begin{bchapter}
\bauthor{\bsnm{Kim}, \binits{Y.}}:
\bctitle{Convolutional neural networks for sentence classification}.
In: \beditor{\bsnm{Moschitti}, \binits{A.}},
\beditor{\bsnm{Pang}, \binits{B.}},
\beditor{\bsnm{Daelemans}, \binits{W.}} (eds.)
\bbtitle{Proceedings of the 2014 Conference on Empirical Methods in Natural Language Processing ({EMNLP})},
pp. \bfpage{1746}--\blpage{1751}.
\bpublisher{Association for Computational Linguistics},
\blocation{Doha, Qatar}
(\byear{2014}).
\doiurl{10.3115/v1/D14-1181} .
\burl{https://aclanthology.org/D14-1181}
\end{bchapter}
\endbibitem

\bibitem[\protect\citeauthoryear{Wang et~al.}{2020}]{wang2020cspnet}
\begin{bchapter}
\bauthor{\bsnm{Wang}, \binits{C.-Y.}},
\bauthor{\bsnm{Liao}, \binits{H.-Y.M.}},
\bauthor{\bsnm{Wu}, \binits{Y.-H.}},
\bauthor{\bsnm{Chen}, \binits{P.-Y.}},
\bauthor{\bsnm{Hsieh}, \binits{J.-W.}},
\bauthor{\bsnm{Yeh}, \binits{I.-H.}}:
\bctitle{Cspnet: A new backbone that can enhance learning capability of cnn}.
In: \bbtitle{Proceedings of the IEEE/CVF Conference on Computer Vision and Pattern Recognition Workshops},
pp. \bfpage{390}--\blpage{391}
(\byear{2020})
\end{bchapter}
\endbibitem

\bibitem[\protect\citeauthoryear{Nagy and Seth}{1984}]{nagy1984hierarchical}
\begin{botherref}
\oauthor{\bsnm{Nagy}, \binits{G.}},
\oauthor{\bsnm{Seth}, \binits{S.C.}}:
Hierarchical representation of optically scanned documents
(1984)
\end{botherref}
\endbibitem

\bibitem[\protect\citeauthoryear{Chu et~al.}{2022}]{chu2022conditional}
\begin{bchapter}
\bauthor{\bsnm{Chu}, \binits{X.}},
\bauthor{\bsnm{Tian}, \binits{Z.}},
\bauthor{\bsnm{Zhang}, \binits{B.}},
\bauthor{\bsnm{Wang}, \binits{X.}},
\bauthor{\bsnm{Shen}, \binits{C.}}:
\bctitle{Conditional positional encodings for vision transformers}.
In: \bbtitle{The Eleventh International Conference on Learning Representations}
(\byear{2022})
\end{bchapter}
\endbibitem

\bibitem[\protect\citeauthoryear{Yu and Koltun}{2015}]{yu2015multi}
\begin{botherref}
\oauthor{\bsnm{Yu}, \binits{F.}},
\oauthor{\bsnm{Koltun}, \binits{V.}}:
Multi-scale context aggregation by dilated convolutions.
arXiv preprint arXiv:1511.07122
(2015)
\end{botherref}
\endbibitem

\bibitem[\protect\citeauthoryear{Ding et~al.}{2024}]{3mvrd}
\begin{botherref}
\oauthor{\bsnm{Ding}, \binits{Y.}},
\oauthor{\bsnm{Vaiani}, \binits{L.}},
\oauthor{\bsnm{Han}, \binits{C.}},
\oauthor{\bsnm{Lee}, \binits{J.}},
\oauthor{\bsnm{Garza}, \binits{P.}},
\oauthor{\bsnm{Poon}, \binits{J.}},
\oauthor{\bsnm{Cagliero}, \binits{L.}}:
M3-vrd: Multimodal multi-task multi-teacher visually-rich form document understanding.
arXiv preprint arXiv:2402.17983
(2024)
\end{botherref}
\endbibitem

\bibitem[\protect\citeauthoryear{Majumder et~al.}{2020}]{payment}
\begin{bchapter}
\bauthor{\bsnm{Majumder}, \binits{B.P.}},
\bauthor{\bsnm{Potti}, \binits{N.}},
\bauthor{\bsnm{Tata}, \binits{S.}},
\bauthor{\bsnm{Wendt}, \binits{J.B.}},
\bauthor{\bsnm{Zhao}, \binits{Q.}},
\bauthor{\bsnm{Najork}, \binits{M.}}:
\bctitle{Representation learning for information extraction from form-like documents}.
In: \bbtitle{Proceedings of the 58th Annual Meeting of the Association for Computational Linguistics},
pp. \bfpage{6495}--\blpage{6504}
(\byear{2020})
\end{bchapter}
\endbibitem

\bibitem[\protect\citeauthoryear{Wang et~al.}{2023}]{vrdu}
\begin{bchapter}
\bauthor{\bsnm{Wang}, \binits{Z.}},
\bauthor{\bsnm{Zhou}, \binits{Y.}},
\bauthor{\bsnm{Wei}, \binits{W.}},
\bauthor{\bsnm{Lee}, \binits{C.-Y.}},
\bauthor{\bsnm{Tata}, \binits{S.}}:
\bctitle{Vrdu: A benchmark for visually-rich document understanding}.
In: \bbtitle{Proceedings of the 29th ACM SIGKDD Conference on Knowledge Discovery and Data Mining},
pp. \bfpage{5184}--\blpage{5193}
(\byear{2023})
\end{bchapter}
\endbibitem

\bibitem[\protect\citeauthoryear{Ding et~al.}{2023}]{formnlu}
\begin{bchapter}
\bauthor{\bsnm{Ding}, \binits{Y.}},
\bauthor{\bsnm{Long}, \binits{S.}},
\bauthor{\bsnm{Huang}, \binits{J.}},
\bauthor{\bsnm{Ren}, \binits{K.}},
\bauthor{\bsnm{Luo}, \binits{X.}},
\bauthor{\bsnm{Chung}, \binits{H.}},
\bauthor{\bsnm{Han}, \binits{S.C.}}:
\bctitle{Form-nlu: Dataset for the form natural language understanding}.
In: \bbtitle{Proceedings of the 46th International ACM SIGIR Conference on Research and Development in Information Retrieval},
pp. \bfpage{2807}--\blpage{2816}
(\byear{2023})
\end{bchapter}
\endbibitem

\bibitem[\protect\citeauthoryear{{\v{S}}imsa et~al.}{2023}]{docile}
\begin{bchapter}
\bauthor{\bsnm{{\v{S}}imsa}, \binits{{\v{S}}.}},
\bauthor{\bsnm{{\v{S}}ulc}, \binits{M.}},
\bauthor{\bsnm{U{\v{r}}i{\v{c}}{\'a}{\v{r}}}, \binits{M.}},
\bauthor{\bsnm{Patel}, \binits{Y.}},
\bauthor{\bsnm{Hamdi}, \binits{A.}},
\bauthor{\bsnm{Koci{\'a}n}, \binits{M.}},
\bauthor{\bsnm{Skalick{\`y}}, \binits{M.}},
\bauthor{\bsnm{Matas}, \binits{J.}},
\bauthor{\bsnm{Doucet}, \binits{A.}},
\bauthor{\bsnm{Coustaty}, \binits{M.}}, \betal:
\bctitle{Docile benchmark for document information localization and extraction}.
In: \bbtitle{International Conference on Document Analysis and Recognition},
pp. \bfpage{147}--\blpage{166}
(\byear{2023}).
\bcomment{Springer}
\end{bchapter}
\endbibitem

\bibitem[\protect\citeauthoryear{Rajpurkar et~al.}{2016}]{rajpurkar2016squad}
\begin{bchapter}
\bauthor{\bsnm{Rajpurkar}, \binits{P.}},
\bauthor{\bsnm{Zhang}, \binits{J.}},
\bauthor{\bsnm{Lopyrev}, \binits{K.}},
\bauthor{\bsnm{Liang}, \binits{P.}}:
\bctitle{Squad: 100,000+ questions for machine comprehension of text}.
In: \bbtitle{Proceedings of the 2016 Conference on Empirical Methods in Natural Language Processing},
pp. \bfpage{2383}--\blpage{2392}
(\byear{2016})
\end{bchapter}
\endbibitem

\bibitem[\protect\citeauthoryear{Biten et~al.}{2019a}]{biten2019scene}
\begin{bchapter}
\bauthor{\bsnm{Biten}, \binits{A.F.}},
\bauthor{\bsnm{Tito}, \binits{R.}},
\bauthor{\bsnm{Mafla}, \binits{A.}},
\bauthor{\bsnm{Gomez}, \binits{L.}},
\bauthor{\bsnm{Rusinol}, \binits{M.}},
\bauthor{\bsnm{Valveny}, \binits{E.}},
\bauthor{\bsnm{Jawahar}, \binits{C.}},
\bauthor{\bsnm{Karatzas}, \binits{D.}}:
\bctitle{Scene text visual question answering}.
In: \bbtitle{Proceedings of the IEEE/CVF International Conference on Computer Vision},
pp. \bfpage{4291}--\blpage{4301}
(\byear{2019})
\end{bchapter}
\endbibitem

\bibitem[\protect\citeauthoryear{Biten et~al.}{2019b}]{stvqa}
\begin{bchapter}
\bauthor{\bsnm{Biten}, \binits{A.F.}},
\bauthor{\bsnm{Tito}, \binits{R.}},
\bauthor{\bsnm{Mafla}, \binits{A.}},
\bauthor{\bsnm{Gomez}, \binits{L.}},
\bauthor{\bsnm{Rusinol}, \binits{M.}},
\bauthor{\bsnm{Mathew}, \binits{M.}},
\bauthor{\bsnm{Jawahar}, \binits{C.}},
\bauthor{\bsnm{Valveny}, \binits{E.}},
\bauthor{\bsnm{Karatzas}, \binits{D.}}:
\bctitle{Icdar 2019 competition on scene text visual question answering}.
In: \bbtitle{2019 International Conference on Document Analysis and Recognition (ICDAR)},
pp. \bfpage{1563}--\blpage{1570}
(\byear{2019}).
\bcomment{IEEE}
\end{bchapter}
\endbibitem

\bibitem[\protect\citeauthoryear{Zhu et~al.}{2022}]{tatdqa}
\begin{bchapter}
\bauthor{\bsnm{Zhu}, \binits{F.}},
\bauthor{\bsnm{Lei}, \binits{W.}},
\bauthor{\bsnm{Feng}, \binits{F.}},
\bauthor{\bsnm{Wang}, \binits{C.}},
\bauthor{\bsnm{Zhang}, \binits{H.}},
\bauthor{\bsnm{Chua}, \binits{T.-S.}}:
\bctitle{Towards complex document understanding by discrete reasoning}.
In: \bbtitle{Proceedings of the 30th ACM International Conference on Multimedia},
pp. \bfpage{4857}--\blpage{4866}
(\byear{2022})
\end{bchapter}
\endbibitem

\bibitem[\protect\citeauthoryear{Zhu et~al.}{2021}]{zhu2021tat}
\begin{bchapter}
\bauthor{\bsnm{Zhu}, \binits{F.}},
\bauthor{\bsnm{Lei}, \binits{W.}},
\bauthor{\bsnm{Huang}, \binits{Y.}},
\bauthor{\bsnm{Wang}, \binits{C.}},
\bauthor{\bsnm{Zhang}, \binits{S.}},
\bauthor{\bsnm{Lv}, \binits{J.}},
\bauthor{\bsnm{Feng}, \binits{F.}},
\bauthor{\bsnm{Chua}, \binits{T.-S.}}:
\bctitle{Tat-qa: A question answering benchmark on a hybrid of tabular and textual content in finance}.
In: \bbtitle{Proceedings of the 59th Annual Meeting of the Association for Computational Linguistics and the 11th International Joint Conference on Natural Language Processing (Volume 1: Long Papers)},
pp. \bfpage{3277}--\blpage{3287}
(\byear{2021})
\end{bchapter}
\endbibitem

\bibitem[\protect\citeauthoryear{Wu et~al.}{2022}]{rdvqa}
\begin{bchapter}
\bauthor{\bsnm{Wu}, \binits{X.}},
\bauthor{\bsnm{Zheng}, \binits{D.}},
\bauthor{\bsnm{Wang}, \binits{R.}},
\bauthor{\bsnm{Sun}, \binits{J.}},
\bauthor{\bsnm{Hu}, \binits{M.}},
\bauthor{\bsnm{Feng}, \binits{F.}},
\bauthor{\bsnm{Wang}, \binits{X.}},
\bauthor{\bsnm{Jiang}, \binits{H.}},
\bauthor{\bsnm{Yang}, \binits{F.}}:
\bctitle{A region-based document vqa}.
In: \bbtitle{Proceedings of the 30th ACM International Conference on Multimedia},
pp. \bfpage{4909}--\blpage{4920}
(\byear{2022})
\end{bchapter}
\endbibitem

\bibitem[\protect\citeauthoryear{Van~Landeghem et~al.}{2023}]{dude}
\begin{bchapter}
\bauthor{\bsnm{Van~Landeghem}, \binits{J.}},
\bauthor{\bsnm{Tito}, \binits{R.}},
\bauthor{\bsnm{Borchmann}, \binits{{\L}.}},
\bauthor{\bsnm{Pietruszka}, \binits{M.}},
\bauthor{\bsnm{Joziak}, \binits{P.}},
\bauthor{\bsnm{Powalski}, \binits{R.}},
\bauthor{\bsnm{Jurkiewicz}, \binits{D.}},
\bauthor{\bsnm{Coustaty}, \binits{M.}},
\bauthor{\bsnm{Anckaert}, \binits{B.}},
\bauthor{\bsnm{Valveny}, \binits{E.}}, \betal:
\bctitle{Document understanding dataset and evaluation (dude)}.
In: \bbtitle{Proceedings of the IEEE/CVF International Conference on Computer Vision},
pp. \bfpage{19528}--\blpage{19540}
(\byear{2023})
\end{bchapter}
\endbibitem

\bibitem[\protect\citeauthoryear{Guo et~al.}{2017}]{guo2017calibration}
\begin{bchapter}
\bauthor{\bsnm{Guo}, \binits{C.}},
\bauthor{\bsnm{Pleiss}, \binits{G.}},
\bauthor{\bsnm{Sun}, \binits{Y.}},
\bauthor{\bsnm{Weinberger}, \binits{K.Q.}}:
\bctitle{On calibration of modern neural networks}.
In: \bbtitle{International Conference on Machine Learning},
pp. \bfpage{1321}--\blpage{1330}
(\byear{2017}).
\bcomment{PMLR}
\end{bchapter}
\endbibitem

\bibitem[\protect\citeauthoryear{Geifman and El-Yaniv}{2017}]{geifman2017selective}
\begin{botherref}
\oauthor{\bsnm{Geifman}, \binits{Y.}},
\oauthor{\bsnm{El-Yaniv}, \binits{R.}}:
Selective classification for deep neural networks.
Advances in neural information processing systems
\textbf{30}
(2017)
\end{botherref}
\endbibitem

\bibitem[\protect\citeauthoryear{Jaeger et~al.}{2022}]{jaeger2022call}
\begin{bchapter}
\bauthor{\bsnm{Jaeger}, \binits{P.F.}},
\bauthor{\bsnm{L{\"u}th}, \binits{C.T.}},
\bauthor{\bsnm{Klein}, \binits{L.}},
\bauthor{\bsnm{Bungert}, \binits{T.J.}}:
\bctitle{A call to reflect on evaluation practices for failure detection in image classification}.
In: \bbtitle{The Eleventh International Conference on Learning Representations}
(\byear{2022})
\end{bchapter}
\endbibitem

\bibitem[\protect\citeauthoryear{Baek et~al.}{2020}]{baek2020cleval}
\begin{bchapter}
\bauthor{\bsnm{Baek}, \binits{Y.}},
\bauthor{\bsnm{Nam}, \binits{D.}},
\bauthor{\bsnm{Park}, \binits{S.}},
\bauthor{\bsnm{Lee}, \binits{J.}},
\bauthor{\bsnm{Shin}, \binits{S.}},
\bauthor{\bsnm{Baek}, \binits{J.}},
\bauthor{\bsnm{Lee}, \binits{C.Y.}},
\bauthor{\bsnm{Lee}, \binits{H.}}:
\bctitle{Cleval: Character-level evaluation for text detection and recognition tasks}.
In: \bbtitle{Proceedings of the IEEE/CVF Conference on Computer Vision and Pattern Recognition Workshops},
pp. \bfpage{564}--\blpage{565}
(\byear{2020})
\end{bchapter}
\endbibitem

\bibitem[\protect\citeauthoryear{Biten et~al.}{2019}]{biten2019icdar}
\begin{bchapter}
\bauthor{\bsnm{Biten}, \binits{A.F.}},
\bauthor{\bsnm{Tito}, \binits{R.}},
\bauthor{\bsnm{Mafla}, \binits{A.}},
\bauthor{\bsnm{Gomez}, \binits{L.}},
\bauthor{\bsnm{Rusinol}, \binits{M.}},
\bauthor{\bsnm{Mathew}, \binits{M.}},
\bauthor{\bsnm{Jawahar}, \binits{C.}},
\bauthor{\bsnm{Valveny}, \binits{E.}},
\bauthor{\bsnm{Karatzas}, \binits{D.}}:
\bctitle{Icdar 2019 competition on scene text visual question answering}.
In: \bbtitle{2019 International Conference on Document Analysis and Recognition (ICDAR)},
pp. \bfpage{1563}--\blpage{1570}
(\byear{2019}).
\bcomment{IEEE}
\end{bchapter}
\endbibitem

\bibitem[\protect\citeauthoryear{Mikolov et~al.}{2013}]{word2vec}
\begin{botherref}
\oauthor{\bsnm{Mikolov}, \binits{T.}},
\oauthor{\bsnm{Chen}, \binits{K.}},
\oauthor{\bsnm{Corrado}, \binits{G.}},
\oauthor{\bsnm{Dean}, \binits{J.}}:
Efficient estimation of word representations in vector space.
arXiv preprint arXiv:1301.3781
(2013)
\end{botherref}
\endbibitem

\bibitem[\protect\citeauthoryear{Pennington et~al.}{2014}]{glove}
\begin{bchapter}
\bauthor{\bsnm{Pennington}, \binits{J.}},
\bauthor{\bsnm{Socher}, \binits{R.}},
\bauthor{\bsnm{Manning}, \binits{C.D.}}:
\bctitle{Glove: Global vectors for word representation}.
In: \bbtitle{Proceedings of the 2014 Conference on Empirical Methods in Natural Language Processing (EMNLP)},
pp. \bfpage{1532}--\blpage{1543}
(\byear{2014})
\end{bchapter}
\endbibitem

\bibitem[\protect\citeauthoryear{Luo et~al.}{2022}]{docgcn}
\begin{bchapter}
\bauthor{\bsnm{Luo}, \binits{S.}},
\bauthor{\bsnm{Ding}, \binits{Y.}},
\bauthor{\bsnm{Long}, \binits{S.}},
\bauthor{\bsnm{Poon}, \binits{J.}},
\bauthor{\bsnm{Han}, \binits{S.C.}}:
\bctitle{Doc-gcn: Heterogeneous graph convolutional networks for document layout analysis}.
In: \bbtitle{Proceedings of the 29th International Conference on Computational Linguistics},
pp. \bfpage{2906}--\blpage{2916}
(\byear{2022})
\end{bchapter}
\endbibitem

\bibitem[\protect\citeauthoryear{Lewis et~al.}{2006}]{iitcdip}
\begin{bchapter}
\bauthor{\bsnm{Lewis}, \binits{D.}},
\bauthor{\bsnm{Agam}, \binits{G.}},
\bauthor{\bsnm{Argamon}, \binits{S.}},
\bauthor{\bsnm{Frieder}, \binits{O.}},
\bauthor{\bsnm{Grossman}, \binits{D.}},
\bauthor{\bsnm{Heard}, \binits{J.}}:
\bctitle{Building a test collection for complex document information processing}.
In: \bbtitle{Proceedings of the 29th Annual International ACM SIGIR Conference on Research and Development in Information Retrieval},
pp. \bfpage{665}--\blpage{666}
(\byear{2006})
\end{bchapter}
\endbibitem

\bibitem[\protect\citeauthoryear{Li et~al.}{2023}]{graphlayoutlm}
\begin{bchapter}
\bauthor{\bsnm{Li}, \binits{Q.}},
\bauthor{\bsnm{Li}, \binits{Z.}},
\bauthor{\bsnm{Cai}, \binits{X.}},
\bauthor{\bsnm{Du}, \binits{B.}},
\bauthor{\bsnm{Zhao}, \binits{H.}}:
\bctitle{Enhancing visually-rich document understanding via layout structure modeling}.
In: \bbtitle{Proceedings of the 31st ACM International Conference on Multimedia},
pp. \bfpage{4513}--\blpage{4523}
(\byear{2023})
\end{bchapter}
\endbibitem

\bibitem[\protect\citeauthoryear{Zhang et~al.}{2022}]{graphdoc}
\begin{botherref}
\oauthor{\bsnm{Zhang}, \binits{Z.}},
\oauthor{\bsnm{Ma}, \binits{J.}},
\oauthor{\bsnm{Du}, \binits{J.}},
\oauthor{\bsnm{Wang}, \binits{L.}},
\oauthor{\bsnm{Zhang}, \binits{J.}}:
Multimodal pre-training based on graph attention network for document understanding.
IEEE Transactions on Multimedia
(2022)
\end{botherref}
\endbibitem

\bibitem[\protect\citeauthoryear{Li et~al.}{2022}]{markuplm}
\begin{bchapter}
\bauthor{\bsnm{Li}, \binits{J.}},
\bauthor{\bsnm{Xu}, \binits{Y.}},
\bauthor{\bsnm{Cui}, \binits{L.}},
\bauthor{\bsnm{Wei}, \binits{F.}}:
\bctitle{Markuplm: Pre-training of text and markup language for visually rich document understanding}.
In: \bbtitle{Proceedings of the 60th Annual Meeting of the Association for Computational Linguistics (Volume 1: Long Papers)},
pp. \bfpage{6078}--\blpage{6087}
(\byear{2022})
\end{bchapter}
\endbibitem

\bibitem[\protect\citeauthoryear{Li et~al.}{2024}]{li2024enhancing}
\begin{bchapter}
\bauthor{\bsnm{Li}, \binits{X.}},
\bauthor{\bsnm{Wu}, \binits{Y.}},
\bauthor{\bsnm{Jiang}, \binits{X.}},
\bauthor{\bsnm{Guo}, \binits{Z.}},
\bauthor{\bsnm{Gong}, \binits{M.}},
\bauthor{\bsnm{Cao}, \binits{H.}},
\bauthor{\bsnm{Liu}, \binits{Y.}},
\bauthor{\bsnm{Jiang}, \binits{D.}},
\bauthor{\bsnm{Sun}, \binits{X.}}:
\bctitle{Enhancing visual document understanding with contrastive learning in large visual-language models}.
In: \bbtitle{Proceedings of the IEEE/CVF Conference on Computer Vision and Pattern Recognition},
pp. \bfpage{15546}--\blpage{15555}
(\byear{2024}).
\burl{https://doi.org/10.1109/CVPR52733.2024.01472}
\end{bchapter}
\endbibitem

\bibitem[\protect\citeauthoryear{Zhu et~al.}{2025}]{gpe}
\begin{bchapter}
\bauthor{\bsnm{Zhu}, \binits{Y.}},
\bauthor{\bsnm{Zhang}, \binits{Y.}},
\bauthor{\bsnm{Liu}, \binits{D.}},
\bauthor{\bsnm{Xie}, \binits{C.}},
\bauthor{\bsnm{Xiong}, \binits{Z.}},
\bauthor{\bsnm{Zheng}, \binits{B.}},
\bauthor{\bsnm{Guo}, \binits{S.}}:
\bctitle{Enhancing document understanding with group position embedding: A novel approach to incorporate layout information}.
In: \bbtitle{The Thirteenth International Conference on Learning Representations}
(\byear{2025}).
\burl{https://openreview.net/forum?id=Dj9a4zQsSl}
\end{bchapter}
\endbibitem

\bibitem[\protect\citeauthoryear{Ye et~al.}{2023a}]{mplugdocowl}
\begin{botherref}
\oauthor{\bsnm{Ye}, \binits{J.}},
\oauthor{\bsnm{Hu}, \binits{A.}},
\oauthor{\bsnm{Xu}, \binits{H.}},
\oauthor{\bsnm{Ye}, \binits{Q.}},
\oauthor{\bsnm{Yan}, \binits{M.}},
\oauthor{\bsnm{Dan}, \binits{Y.}},
\oauthor{\bsnm{Zhao}, \binits{C.}},
\oauthor{\bsnm{Xu}, \binits{G.}},
\oauthor{\bsnm{Li}, \binits{C.}},
\oauthor{\bsnm{Tian}, \binits{J.}},
\oauthor{\bsnm{Qi}, \binits{Q.}},
\oauthor{\bsnm{Zhang}, \binits{J.}},
\oauthor{\bsnm{Huang}, \binits{F.}}:
mPLUG-DocOwl: Modularized Multimodal Large Language Model for Document Understanding
(2023).
\url{https://arxiv.org/abs/2307.02499}
\end{botherref}
\endbibitem

\bibitem[\protect\citeauthoryear{Ye et~al.}{2023b}]{ureader}
\begin{bchapter}
\bauthor{\bsnm{Ye}, \binits{J.}},
\bauthor{\bsnm{Hu}, \binits{A.}},
\bauthor{\bsnm{Xu}, \binits{H.}},
\bauthor{\bsnm{Ye}, \binits{Q.}},
\bauthor{\bsnm{Yan}, \binits{M.}},
\bauthor{\bsnm{Xu}, \binits{G.}},
\bauthor{\bsnm{Li}, \binits{C.}},
\bauthor{\bsnm{Tian}, \binits{J.}},
\bauthor{\bsnm{Qian}, \binits{Q.}},
\bauthor{\bsnm{Zhang}, \binits{J.}}, \betal:
\bctitle{Ureader: Universal ocr-free visually-situated language understanding with multimodal large language model}.
In: \bbtitle{Findings of the Association for Computational Linguistics: EMNLP 2023},
pp. \bfpage{2841}--\blpage{2858}
(\byear{2023}).
\burl{https://doi.org/10.18653/v1/2023.findings-emnlp.187}
\end{bchapter}
\endbibitem

\bibitem[\protect\citeauthoryear{Hu et~al.}{2024}]{mplugdocowl15}
\begin{bchapter}
\bauthor{\bsnm{Hu}, \binits{A.}},
\bauthor{\bsnm{Xu}, \binits{H.}},
\bauthor{\bsnm{Ye}, \binits{J.}},
\bauthor{\bsnm{Yan}, \binits{M.}},
\bauthor{\bsnm{Zhang}, \binits{L.}},
\bauthor{\bsnm{Zhang}, \binits{B.}},
\bauthor{\bsnm{Zhang}, \binits{J.}},
\bauthor{\bsnm{Jin}, \binits{Q.}},
\bauthor{\bsnm{Huang}, \binits{F.}},
\bauthor{\bsnm{Zhou}, \binits{J.}}:
\bctitle{mplug-docowl 1.5: Unified structure learning for ocr-free document understanding}.
In: \bbtitle{Findings of the Association for Computational Linguistics: EMNLP 2024},
pp. \bfpage{3096}--\blpage{3120}
(\byear{2024}).
\burl{https://aclanthology.org/2024.findings-emnlp.175}
\end{bchapter}
\endbibitem

\bibitem[\protect\citeauthoryear{Liu et~al.}{2024}]{textmonkey}
\begin{botherref}
\oauthor{\bsnm{Liu}, \binits{Y.}},
\oauthor{\bsnm{Yang}, \binits{B.}},
\oauthor{\bsnm{Liu}, \binits{Q.}},
\oauthor{\bsnm{Li}, \binits{Z.}},
\oauthor{\bsnm{Ma}, \binits{Z.}},
\oauthor{\bsnm{Zhang}, \binits{S.}},
\oauthor{\bsnm{Bai}, \binits{X.}}:
TextMonkey: An OCR-Free Large Multimodal Model for Understanding Document
(2024).
\url{https://arxiv.org/abs/2403.04473}
\end{botherref}
\endbibitem

\bibitem[\protect\citeauthoryear{Feng et~al.}{2024}]{feng2024docpedia}
\begin{barticle}
\bauthor{\bsnm{Feng}, \binits{H.}},
\bauthor{\bsnm{Liu}, \binits{Q.}},
\bauthor{\bsnm{Liu}, \binits{H.}},
\bauthor{\bsnm{Tang}, \binits{J.}},
\bauthor{\bsnm{Zhou}, \binits{W.}},
\bauthor{\bsnm{Li}, \binits{H.}},
\bauthor{\bsnm{Huang}, \binits{C.}}:
\batitle{Docpedia: Unleashing the power of large multimodal model in the frequency domain for versatile document understanding}.
\bjtitle{Science China Information Sciences}
\bvolume{67}(\bissue{12}),
\bfpage{1}--\blpage{14}
(\byear{2024})
\end{barticle}
\endbibitem

\bibitem[\protect\citeauthoryear{Wang et~al.}{2024}]{Qwen2-VL}
\begin{botherref}
\oauthor{\bsnm{Wang}, \binits{P.}},
\oauthor{\bsnm{Bai}, \binits{S.}},
\oauthor{\bsnm{Tan}, \binits{S.}},
\oauthor{\bsnm{Wang}, \binits{S.}},
\oauthor{\bsnm{Fan}, \binits{Z.}},
\oauthor{\bsnm{Bai}, \binits{J.}},
\oauthor{\bsnm{Chen}, \binits{K.}},
\oauthor{\bsnm{Liu}, \binits{X.}},
\oauthor{\bsnm{Wang}, \binits{J.}},
\oauthor{\bsnm{Ge}, \binits{W.}},
\oauthor{\bsnm{Fan}, \binits{Y.}},
\oauthor{\bsnm{Dang}, \binits{K.}},
\oauthor{\bsnm{Du}, \binits{M.}},
\oauthor{\bsnm{Ren}, \binits{X.}},
\oauthor{\bsnm{Men}, \binits{R.}},
\oauthor{\bsnm{Liu}, \binits{D.}},
\oauthor{\bsnm{Zhou}, \binits{C.}},
\oauthor{\bsnm{Zhou}, \binits{J.}},
\oauthor{\bsnm{Lin}, \binits{J.}}:
Qwen2-vl: Enhancing vision-language model's perception of the world at any resolution.
arXiv preprint arXiv:2409.12191
(2024)
\end{botherref}
\endbibitem

\bibitem[\protect\citeauthoryear{Yu et~al.}{2024a}]{texthawk}
\begin{botherref}
\oauthor{\bsnm{Yu}, \binits{Y.-Q.}},
\oauthor{\bsnm{Liao}, \binits{M.}},
\oauthor{\bsnm{Wu}, \binits{J.}},
\oauthor{\bsnm{Liao}, \binits{Y.}},
\oauthor{\bsnm{Zheng}, \binits{X.}},
\oauthor{\bsnm{Zeng}, \binits{W.}}:
TextHawk: Exploring Efficient Fine-Grained Perception of Multimodal Large Language Models
(2024).
\url{https://arxiv.org/abs/2404.09204}
\end{botherref}
\endbibitem

\bibitem[\protect\citeauthoryear{Yu et~al.}{2024b}]{texthawk2}
\begin{botherref}
\oauthor{\bsnm{Yu}, \binits{Y.-Q.}},
\oauthor{\bsnm{Liao}, \binits{M.}},
\oauthor{\bsnm{Zhang}, \binits{J.}},
\oauthor{\bsnm{Wu}, \binits{J.}}:
TextHawk2: A Large Vision-Language Model Excels in Bilingual OCR and Grounding with 16x Fewer Tokens
(2024).
\url{https://arxiv.org/abs/2410.05261}
\end{botherref}
\endbibitem

\bibitem[\protect\citeauthoryear{Wei et~al.}{2024}]{vary}
\begin{bchapter}
\bauthor{\bsnm{Wei}, \binits{H.}},
\bauthor{\bsnm{Kong}, \binits{L.}},
\bauthor{\bsnm{Chen}, \binits{J.}},
\bauthor{\bsnm{Zhao}, \binits{L.}},
\bauthor{\bsnm{Ge}, \binits{Z.}},
\bauthor{\bsnm{Yang}, \binits{J.}},
\bauthor{\bsnm{Sun}, \binits{J.}},
\bauthor{\bsnm{Han}, \binits{C.}},
\bauthor{\bsnm{Zhang}, \binits{X.}}:
\bctitle{Vary: Scaling up the vision vocabulary for large vision-language model}.
In: \bbtitle{European Conference on Computer Vision},
pp. \bfpage{408}--\blpage{424}
(\byear{2024}).
\bcomment{Springer}
\end{bchapter}
\endbibitem

\bibitem[\protect\citeauthoryear{Wang et~al.}{2025}]{marten}
\begin{botherref}
\oauthor{\bsnm{Wang}, \binits{Z.}},
\oauthor{\bsnm{Guan}, \binits{T.}},
\oauthor{\bsnm{Fu}, \binits{P.}},
\oauthor{\bsnm{Duan}, \binits{C.}},
\oauthor{\bsnm{Jiang}, \binits{Q.}},
\oauthor{\bsnm{Guo}, \binits{Z.}},
\oauthor{\bsnm{Guo}, \binits{S.}},
\oauthor{\bsnm{Luo}, \binits{J.}},
\oauthor{\bsnm{Shen}, \binits{W.}},
\oauthor{\bsnm{Yang}, \binits{X.}}:
Marten: Visual Question Answering with Mask Generation for Multi-modal Document Understanding
(2025).
\url{https://arxiv.org/abs/2503.14140}
\end{botherref}
\endbibitem

\bibitem[\protect\citeauthoryear{Zhang et~al.}{2024}]{dockylin}
\begin{botherref}
\oauthor{\bsnm{Zhang}, \binits{J.}},
\oauthor{\bsnm{Yang}, \binits{W.}},
\oauthor{\bsnm{Lai}, \binits{S.}},
\oauthor{\bsnm{Xie}, \binits{Z.}},
\oauthor{\bsnm{Jin}, \binits{L.}}:
Dockylin: A large multimodal model for visual document understanding with efficient visual slimming.
arXiv preprint arXiv:2406.19101
(2024)
\end{botherref}
\endbibitem

\bibitem[\protect\citeauthoryear{Hu et~al.}{2024}]{mplugdocowl2}
\begin{botherref}
\oauthor{\bsnm{Hu}, \binits{A.}},
\oauthor{\bsnm{Xu}, \binits{H.}},
\oauthor{\bsnm{Zhang}, \binits{L.}},
\oauthor{\bsnm{Ye}, \binits{J.}},
\oauthor{\bsnm{Yan}, \binits{M.}},
\oauthor{\bsnm{Zhang}, \binits{J.}},
\oauthor{\bsnm{Jin}, \binits{Q.}},
\oauthor{\bsnm{Huang}, \binits{F.}},
\oauthor{\bsnm{Zhou}, \binits{J.}}:
mPLUG-DocOwl2: High-resolution Compressing for OCR-free Multi-page Document Understanding
(2024).
\url{https://arxiv.org/abs/2409.03420}
\end{botherref}
\endbibitem

\bibitem[\protect\citeauthoryear{Gandhi et~al.}{2024}]{gandhi2024better}
\begin{bchapter}
\bauthor{\bsnm{Gandhi}, \binits{S.}},
\bauthor{\bsnm{Gala}, \binits{R.}},
\bauthor{\bsnm{Viswanathan}, \binits{V.}},
\bauthor{\bsnm{Wu}, \binits{T.}},
\bauthor{\bsnm{Neubig}, \binits{G.}}:
\bctitle{Better synthetic data by retrieving and transforming existing datasets}.
In: \bbtitle{Findings of the Association for Computational Linguistics ACL 2024},
pp. \bfpage{6453}--\blpage{6466}
(\byear{2024})
\end{bchapter}
\endbibitem

\bibitem[\protect\citeauthoryear{Ding et~al.}{2024}]{david}
\begin{botherref}
\oauthor{\bsnm{Ding}, \binits{Y.}},
\oauthor{\bsnm{Han}, \binits{S.C.}},
\oauthor{\bsnm{Li}, \binits{Z.}},
\oauthor{\bsnm{Chung}, \binits{H.}}:
David: Domain adaptive visually-rich document understanding with synthetic insights.
arXiv preprint arXiv:2410.01609
(2024)
\end{botherref}
\endbibitem

\bibitem[\protect\citeauthoryear{Li et~al.}{2024}]{li2024survey}
\begin{botherref}
\oauthor{\bsnm{Li}, \binits{D.}},
\oauthor{\bsnm{Wang}, \binits{Z.}},
\oauthor{\bsnm{Chen}, \binits{Y.}},
\oauthor{\bsnm{Jiang}, \binits{R.}},
\oauthor{\bsnm{Ding}, \binits{W.}},
\oauthor{\bsnm{Okumura}, \binits{M.}}:
A survey on deep active learning: Recent advances and new frontiers.
IEEE Transactions on Neural Networks and Learning Systems
(2024)
\end{botherref}
\endbibitem

\end{thebibliography}

\end{document}